\documentclass[conference]{IEEEtran}
\usepackage{times}

\usepackage[numbers]{natbib}
\usepackage{multicol}
\usepackage[bookmarks=true]{hyperref}
\usepackage{amsmath}

\usepackage[utf8]{inputenc} 
\usepackage[T1]{fontenc}    
\usepackage{hyperref}       
\usepackage{url}            
\usepackage{booktabs}       
\usepackage{amsfonts}       
\usepackage{nicefrac}       
\usepackage{microtype}      
\usepackage{xcolor}         
\usepackage{tcolorbox}
\usepackage{wrapfig}
\usepackage{graphicx,subcaption}
\usepackage{amssymb,amsfonts,graphicx}
\usepackage{tabularray}
\usepackage{caption}
\usepackage{subcaption}
\usepackage{dblfloatfix}
\usepackage{tikzducks}
\usepackage{titletoc}
\usepackage{booktabs}
\usepackage[table,xcdraw]{xcolor}

\newcommand{\nocontentsline}[3]{}
\let\origcontentsline\addcontentsline
\newcommand\stoptoc{\let\addcontentsline\nocontentsline}
\newcommand\resumetoc{\let\addcontentsline\origcontentsline}
\stoptoc

\usepackage{cuted}

\begin{document}

\title{What Do You Need for Compositional Generalization in Diffusion Planning?}

\author{Quentin Clark and Florian Shkurti \\ Department of Computer Science, University of Toronto}

\twocolumn[{
\renewcommand\twocolumn[1][]{#1}
\maketitle
    \vspace{-5mm}
\begin{center}
    \includegraphics[width=1.0\textwidth]{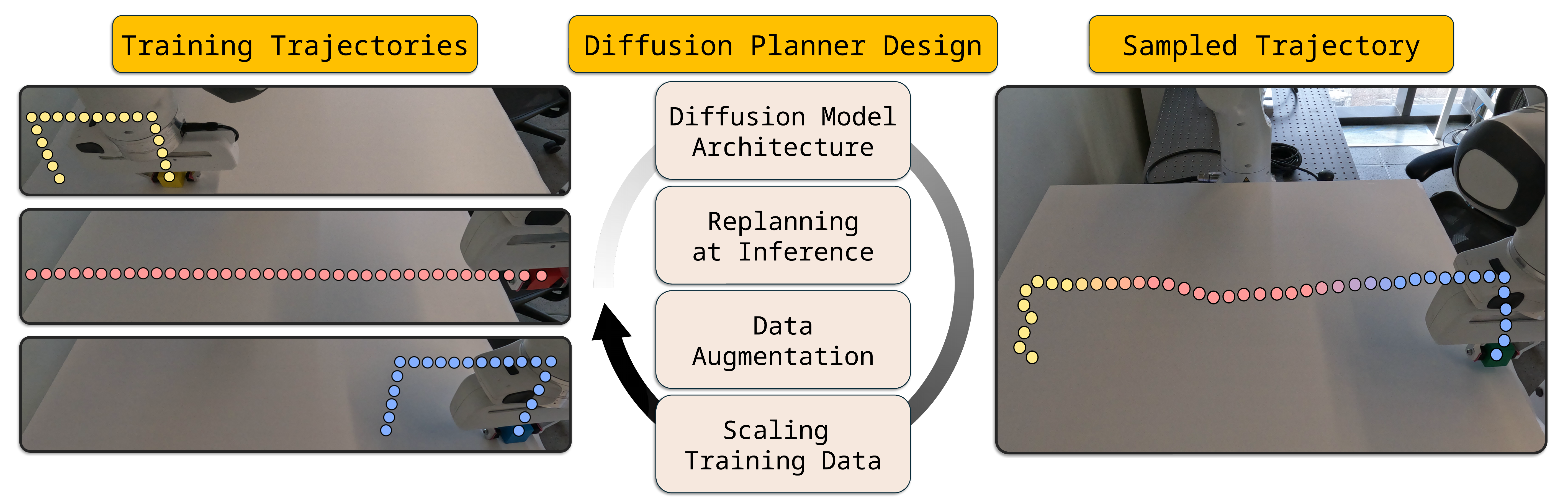}
    \captionof{figure}{ \label{fig:eyecatcher} We provide an analysis that identifies the critical design decisions that enable diffusion planners to exhibit stitching and compositional generalization via generative behaviour cloning, without resorting to dynamic programming or TD learning, as commonly done in offline RL. 
    } 
\end{center}
}]


%


\begin{abstract}
In policy learning, stitching and compositional generalization refer to the extent to which the policy is able to piece together sub-trajectories of data it is trained on to generate new and diverse behaviours. While stitching has been identified as a significant strength of offline reinforcement learning, recent generative behavioural cloning (BC) methods have also shown proficiency at stitching. However, the main factors behind  this are poorly understood, hindering the development of new algorithms that can reliably stitch by design. Focusing on diffusion planners trained via generative behavioural cloning, and without resorting to dynamic programming or TD-learning, we find three properties are key enablers for composition: shift equivariance, local receptive fields, and inference choices. We use these properties to explain architecture, data, and inference choices in existing generative BC methods based on diffusion planning including replanning frequency, data augmentation, and data scaling. Our experiments show that while local receptive fields are more important than shift equivariance in creating a diffusion planner capable of composition, both are crucial. Using findings from our experiments, we develop a new architecture for diffusion planners called Eq-Net, that is simple, produces diverse trajectories competitive with more computationally expensive methods such as replanning or scaling data, and can be guided to enable generalization in goal-conditioned settings. We show that Eq-Net exhibits significant compositional generalization in a variety of navigation and manipulation tasks designed to test planning diversity.

\end{abstract}

\IEEEpeerreviewmaketitle

\section{Introduction}
The advent of diffusion models~\cite{ho2020denoising} has inspired a wave of work using these models as planners and policies in robotic systems~\cite{janner2022planning,chi2023diffusion}. Image and video diffusion models exhibit strong \emph{compositionality}: the ability to take component elements seen in samples of their dataset and compose them in new, feasible ways~\cite{ho2021classifierfree,du2020compositional,liu2021learning}. When and why diffusion models for planning and control compose, however, is not well understood. Diffusion Policies~\cite{chi2023diffusion}, which generate short action chunks, naturally compose behaviours through frequent replanning. However, Diffusion Planners~\cite{janner2022planning}, which generate long-horizon plans, often struggle with composition~\cite{goli2025stitchope,chen2024diffusion}, instead reproducing exact training trajectories. 

Composition in diffusion planners is crucial to address, as the capability to compose behaviours into new trajectories at test time is important property for planners broadly. Reasons include allowing for better candidate set selection~\cite{zhang2020falco}, encouraging multi-modal behaviour~\cite{lee2023stamp}, and generating better data for offline reinforcement learning~\cite{kumar2022should}. 

To address this gap, we analyze a variety of approaches that induce composition in diffusion planners. We find that there are two key ingredients: the ability to use sub-skills at any point in a sequence (\emph{shift equivariance}~\cite{kamb2024analytic}), and a locality bias towards attending to nearby states when generating a state in a trajectory (\emph{local receptive fields}~\cite{niedoba2024towards}). We show that previous methods achieve composition through these two ingredients, identifying composition can be ensured through careful inference choices, data augmentation strategies, or data scaling. 

Using these two ingredients, we develop a neural network architecture for the diffusion planning denoising network called Eq-Net, which is simple, produces diverse trajectories, and can be guided to enable strong goal-conditioning. We show Eq-Net produces more diverse trajectories than other architectures in a variety of navigation and robotic manipulation settings. 

\newpage
\noindent Our contributions are as follows:
\begin{itemize}
\setlength\itemsep{0em}
\item We show that local receptive fields and shift equivariance are key enablers for composition in diffusion planners, as has been shown in the image generation domain~\cite{niedoba2024towards,kamb2024analytic}. However, unlike the image generation case, they are not the only important factors: the inference method used to sample trajectories from diffusion planners also plays a crucial role in achieving compositional generalization. 
\item We connect the presence or absence of composition in past work on diffusion planning~\cite{chen2024diffusion,luo2025generative,chen2024diffusion,song2025history,chi2023diffusion} to the key factors identified above. 
\item We show that stitching achieved through data scaling, as opposed to architectural choices, still induces local receptive fields in models.
\item We demonstrate that when induced through architecture changes, the stitching capabilities of whole-sequence diffusion can be effectively guided through inpainting to achieve strong goal-conditioned performance on unseen tasks.

\par
\end{itemize}

\noindent Our work aims to provide clarity behind the key factors that guide diffusion planners to compose trajectories at test-time, informing future algorithmic and architectural developments in compositional diffusion planners.

\section{Preliminaries and Related Work}
\label{section:related}

\textbf{Diffusion Models:} Diffusion models~\cite{sohl2015deep,ho2020denoising} are generative models trained on data points drawn from some target distribution $P$ by taking a sample $x \sim P$ and iteratively adding Gaussian noise, such that $x_{t+1} = x_t + \eta_t, \eta_t \sim \mathcal{N} (0,\sigma^2)$, called the \emph{forward diffusion process}. The diffusion model learns the \emph{reverse diffusion process}, where a neural network $f_\theta$ is trained to learn $f(x_t,t) = x_{t-1}$ by minimizing the MSE loss between the model's output and the reverse diffusion step: $ \theta^{\star} = \text{argmin}_\theta \sum_{t=1}^T\mathbb{E} || f_\theta(x_t) - x_{t-1}||^2_2$. Sampling is performed by sampling from the Gaussian and using the model to reverse-sample the diffusion process by taking repeated steps of size $\Delta t$ from $x_1\to x_0$: $x_{t-1} = f(x_t,t) + \mathcal{N} (0,\sigma^2\Delta t)$\cite{24diffusiontutorial,chan2024tutorial}. Many different formulations to diffusion model prediction exist, but most works either predict a clean sample from noise~\cite{li2026basicsletdenoisinggenerative}  or the residual noise to remove~\cite{ho2020denoising}.

\textbf{Diffusion Planning:} Introduced by \citet{janner2022planning}, diffusion planners model entire trajectories by sampling from diffusion models. A diffusion planner generates an upcoming state $s_i\in R^n$ trajectory $\tau$ consisting of a sequence of states  $\tau = [s_1,s_2,...,s_{H_F}]$ horizon $H_F$ time-steps long, sometimes conditioned on a state memory of length $H_P$ in an environment with $T$ max steps. Actions from the plan are executed for a horizon $H_A$ before generating a new trajectory. Here we mostly investigate the \emph{whole-sequence} planning domain, where $H_A = H_F = T$, and no replanning occurs. The more common Diffusion Policy~\cite{chi2023diffusion} formulation, where the diffusion model generates a short sequence of actions with high reactivity from frequent re-generation of actions, can be thought of as a special form of diffusion planning where $H_A \approx2H_F <<T$. We elaborate on this distinction further in Appendix \ref{appendix:other_diffusion_planners}.


\label{related:diffgen}

\textbf{Compositionality in Diffusion Planners:} Many approaches exist to explicitly induce compositionality in diffusion planners. Hierarchical approaches use diffusion to stitch existing dataset trajectories \cite{li2024diffstitch}, generate subsequences jointly conditioned on each-other \cite{luo2025generative}, or separately generate high and low level plans\cite{chen2024simplehierarchicalplanningdiffusion,li2023hierarchical}. By contrast, we target whole-sequence compositionality, where composition does not occur from explicit stitching but from implicit model behaviour, similar to image diffusion. Benefits of a monolithic planning approach includes reducing algorithmic complexity, sidestepping potentially harmful planning hierarchy~\cite{bacchus1991downward}, and enhanced performance in long-horizon tasks~\cite{chen2024diffusion,lu2025makes}.

\textbf{Generalization in Image Diffusion Models:} Work in Diffusion model generalization typically examines image-based models. Approaches include examining the effect of the diffusion process noise \cite{vastola2025generalization}, architectural biases \cite{kamb2024analytic,niedoba2024towards}, data statistics~\cite{lukoianov2025locality} and representation learning \cite{kadkhodaie2024generalization} among others. We extend these works by connecting their observations to robotics and showing that through introducing architectural biases, we can enable compositional models in low-data regimes.




\section{Methodology}

\subsection{Do Existing Diffusion Planners Compose by Default?}
\label{section:current_memorization}
\begin{figure}[b]
\centering
  \includegraphics[width=.98\linewidth]{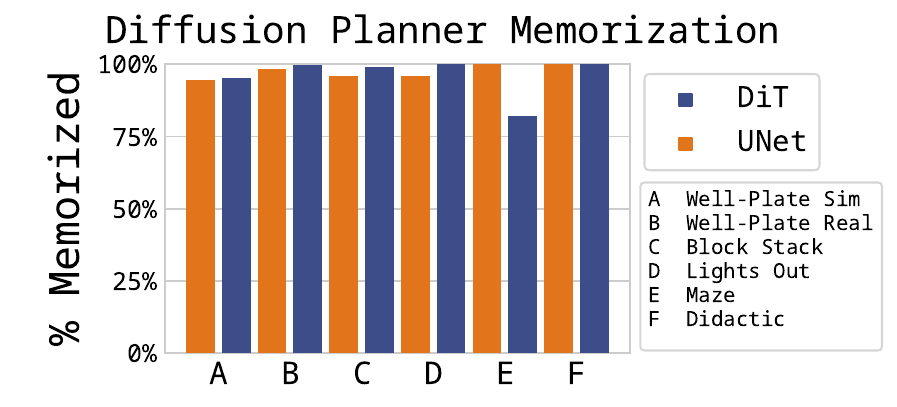}
\caption{\textbf{Common Diffusion Backbones Memorize}: We define memorization as new trajectories only consisting of sub-skills previously seen before in the same trajectory. High memorization rates for both U-Net and DiT architectures used as diffusion planners in a variety of environments shows that composition cannot be taken for granted in diffusion planning. See Appendix~\ref{appendix:why_janner_no_compose} for more discussion.}
\label{fig:memorization}
\end{figure}

\begin{figure*}[b!]
    \centering
  \includegraphics[width=.98\linewidth]{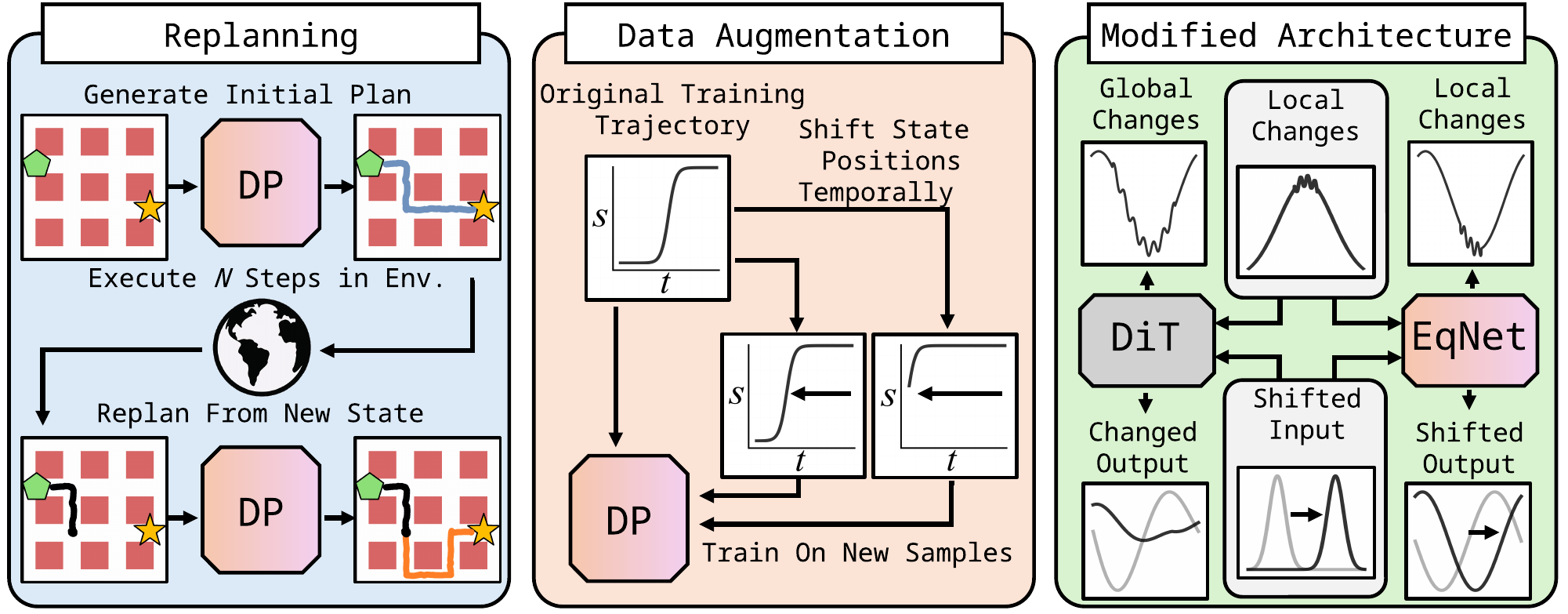}
\caption{The methods to enable trajectory composition that we examine. The left shows replanning (an inference technique), the middle positional augmentation (a dataset augmentation technique) and the right architectural modification. Each of these enables composition, showing that the broad principles of local receptive fields and shift equivariance can be applied in different contexts.}
\label{fig:method}
\end{figure*}

To motivate our problem, we observe that when standard diffusion planners are naively applied to a variety of tasks, the planners tend to exactly reproduce training trajectories. Figure~\ref{fig:memorization} shows the rate that the two most common diffusion planner backbones, U-Net~\cite{janner2022planning} and DiT~\cite{dong2023aligndiff}, produce memorized trajectories in our testing environments designed to test compositionality, as we define in Section~\ref{section:define_novelty}. This is surprising, given that diffusion models exhibit creativity in image and video domains~\cite{han2025enhancing}. 

However, diffusion planners are often trained and evaluated differently than image diffusion models. Image diffusion models are trained to produce whole images at once. Meanwhile, in diffusion planning people employ different inference and training techniques, usually introducing training augmentations or generating smaller subsequences. We outline the most common of these techniques in Figure~\ref{fig:method}.

\subsection{Local Receptive Fields and Shift Equivariance in Score-Based Diffusion Models}

To unify prior explanations for compositionality in diffusion planners, we rely on two properties previously described for diffusion models in vision~\cite{kamb2024analytic,niedoba2024towards}. The first is \emph{local receptive fields}, meaning the diffusion model's denoising prediction for step $t$ in the trajectory is conditionally independent of far-away steps in the trajectory. Local receptive fields ensure that at each step, the diffusion model only attends to a small patch around part of a sample when denoising. However, through multiple denoising steps, information can propagate across a generated sample, shown in Appendix~\ref{appendix:propogation}. 

\textbf{Definition 1: Locally receptive fields.} Assume we are interested in generating a trajectory $\tau$, and have a generative trajectory model $M_t$ whose output estimates the denoising function $M_t[\tau]$ for individual states $M_t[\tau](x_s)$ where $x_s$ indexes states $s = \tau(x_s)$. We say that $M_t[\tau]$ has a field locally receptive to a state neighbourhood $\Omega$ if for all trajectories $\tau$ and state positions within those trajectories $x_s$, $M_t[\tau](x_s) = M_t[\tau | \Omega_{x_s}](x_s)$. 

The second is \emph{shift equivariance}, which requires that the model ignore the absolute position of a state in the sequence as a condition during the denoising process. 

\textbf{Definition 2: Shift equivariance.}  A trajectory score model $M_t[\tau]$ is shift equivariant~\cite{mcgreivy2022convolutional} if for all shifts $U$ on trajectories $\tau$ which uniformly change the position of states within the trajectory, $M_t[U[\tau]] = U[M_t[\tau]]$. For diffusion planners, this can also be thought of as the diffusion model having consistent denoising predictions for similar sub-trajectory skills, regardless of their absolute position in trajectory. For instance, a diffusion planner should be able to de-noise a "pick" sub-skill for a robotic arm near the beginning or near the end of a trajectory, regardless of where that pick appeared in training trajectories.

\subsection{Additional Factors for Composition in Diffusion Planners}

\label{section:comp_methods}
Our main insight is that these notions of local receptive fields and shift equivariance are similar to other strategies used to increase composition in diffusion planners. These other strategies are a crucial component to how modern diffusion planners compose~\cite{chi2023diffusion,luo2025generative}. 

\textbf{Increasing Compositionality by Replanning at Inference:} The first strategy we address is \emph{replanning}. A trajectory planning model replans when conditioned on past observations $H_P$ it generates a trajectory $\tau$ of length $H_F$, executes some action horizon $H_A \le H_F$, and then generates a new trajectory starting at the current state. Increasing $H_F$ extends the model's effective planning horizon. Formally, given a generative trajectory model $f$ that generates trajectories of length $H_F$ conditioned on a history of length $H_P$, each state in the trajectory will only be directly attentive to a local field of size $l = \max(H_F,H_P)$, i.e, $P(s_t |s_{t-l},s_{t-l+1},...s_{t+l-1},s_{t+l})$. Short-horizon diffusion planners, such as diffusion policies~\cite{chi2023diffusion} achieve composition through this mechanism. However, this does not elucidate how longer-horizon planners compose. 

Note that fixed-memory replanning and local receptive fields in the reverse-diffusion process are not identical, as the many steps in the reverse diffusion process lets influence spread along a trajectory through many local denoising steps, shown in Appendix~\ref{appendix:propogation}. However, the similar mechanisms highlight how increasing the influence of nearby states in a given trajectory is important for composition. Replanning also ensures a form of shift equivariance, as the diffusion model must be capable of generating chunks at any time-step in the trajectory.

\textbf{Increasing Compositionality by Data Augmentation:} Another strategy to increase compositionality is \emph{data augmentation}, where data is modified during training to teach the model equivariance to a transformation~\cite{biscione2021convolutional,shorten2019survey}. The most common augmentation to encourage stitching is stitching together sub-trajectories in the training set randomly~\cite{ghugare2024closing,char2022bats,hepburn2022model,lee2025state}. However, this strategy requires tuning the distance metric that determines when segments are close, and is infeasible for datasets with a large combinatorial set of legal stitches because such sets often grow exponentially. 

We instead focus on augmenting the data to provide shift equivariance. Specifically, we randomize the position of states in a trajectory during training to prevent the model from overfitting to a state's position. We sample a random state in a trajectory, removing states in the past, and then pad the end of the trajectory with the final state, so that the trajectory is randomly "shifted" by a number of states.

Formally, given a training trajectory $\tau$ with states $\tau[1] = s_1,\tau[2] = s_2, ... \tau[T] = s_T$, we sample a position from the uniform distribution $i \sim U_{[1,T]}$. To create our new augmented trajectory $\tau^\star$, we shift the states in $\tau$ by $i$ positions to the left, such that for state indexes $x \in [1,T-i)$, $\tau^\star[x] = \tau[x+i]$. For the rest of the states $x \in [T-i,T]$, we pad the states with the final state: $\tau^\star[x] = \tau[T]$. This modified trajectory is then used in training.

\textbf{Increasing Compositionality via Architecture Design:} The next strategy we address are \emph{architecture modifications}. While equivariance can be learned through data augmentation, another strategy is to build equivariance to certain transformations through architecture changes\cite{cohen2021equivariant,cohen2016group}.

The more modern DiT backbone is Transformer based~\cite{vaswani2017attention}, which does not have an inductive bias for local receptive fields when processing input sequences and has explicit positional encoding. While CNNs are often said to have locality and shift equivariance~\cite{janner2022planning,kamb2024analytic}, we found that the CNN based U-Net~\cite{janner2022planning} has a globally receptive field and is not equivariant to shifts, shown in Appendix~\ref{appendix:unet_analysis}. In fact, down-sampling and pooling operators common in CNNs provably break shift equivariance~\cite{zhang2019making,chaman2021truly}. 

To analyze the effect that local receptive fields and shift equivariance have on composition, we propose a network called \emph{Eq-Net}, a residual CNN~\cite{he2016deep}, with two important design decisions:
\begin{itemize}
    \item Convolutional layers \textbf{without pooling or down-sampling} guarantee equivariance to shifts~\cite{zhang2019making}.
    \item Small convolutional kernels guarantee local receptive fields.
\end{itemize}
See Appendix \ref{appendix:eqnet_details} for further elaboration and an architecture diagram.

\textbf{Increasing Compositionality via Scaling Training Data:} The last strategy we address is \emph{scaling training data}, which increases diffusion model generalization~\cite{zhang2024emergence} and compositionality~\cite{okawa2023compositional}. This increase in compositionality does not come from memorizing more of the sample space, but from learning the rules governing composition in a particular area~\cite{favero2025compositional,lukoianov2025locality}.

Our work is specific to robotics, where data is often sparse. Vision datasets range from $\sim60,000$ samples~\cite{krizhevsky2014cifar} to millions~\cite{deng2009imagenet,xu2023demystifying} of images or more. By contrast, robotics datasets are smaller, varying from a couple dozen trajectories \cite{chi2023diffusion} to a few hundred~\cite{mandlekar2021matters} or thousand~\cite{park2024ogbench}. Given that most robotics datasets have relatively few samples, considering other approaches to generalization becomes more important than relying on data scaling. 

\subsection{Environment Descriptions}
\label{section:env}

\begin{figure}[b]
    \centering
    \begin{subfigure}[T]{.31\linewidth}{\includegraphics[width=\linewidth]{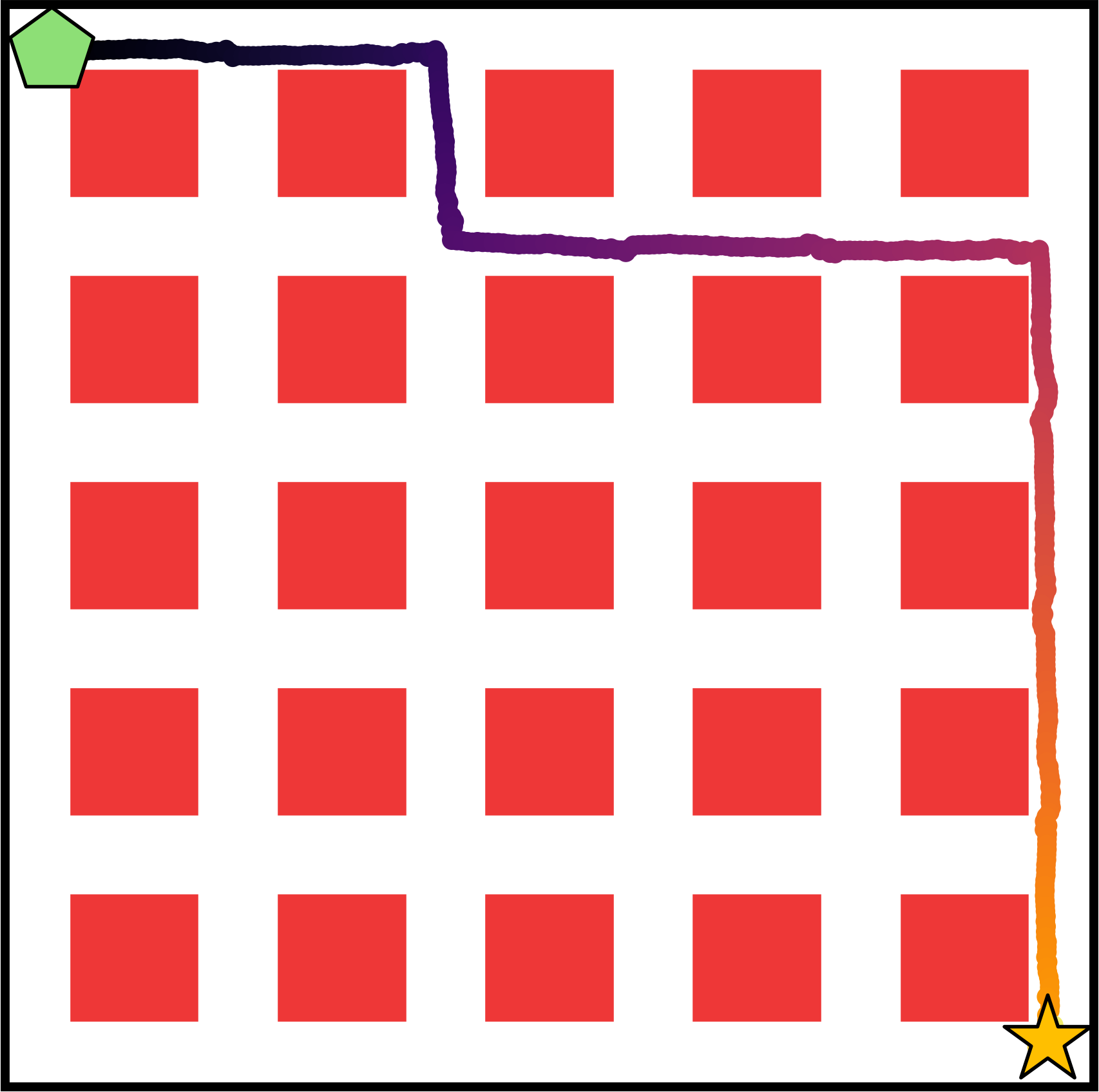}} \end{subfigure}\hspace{1mm}
    \begin{subfigure}[T]{.31\linewidth}{\includegraphics[width=\linewidth]{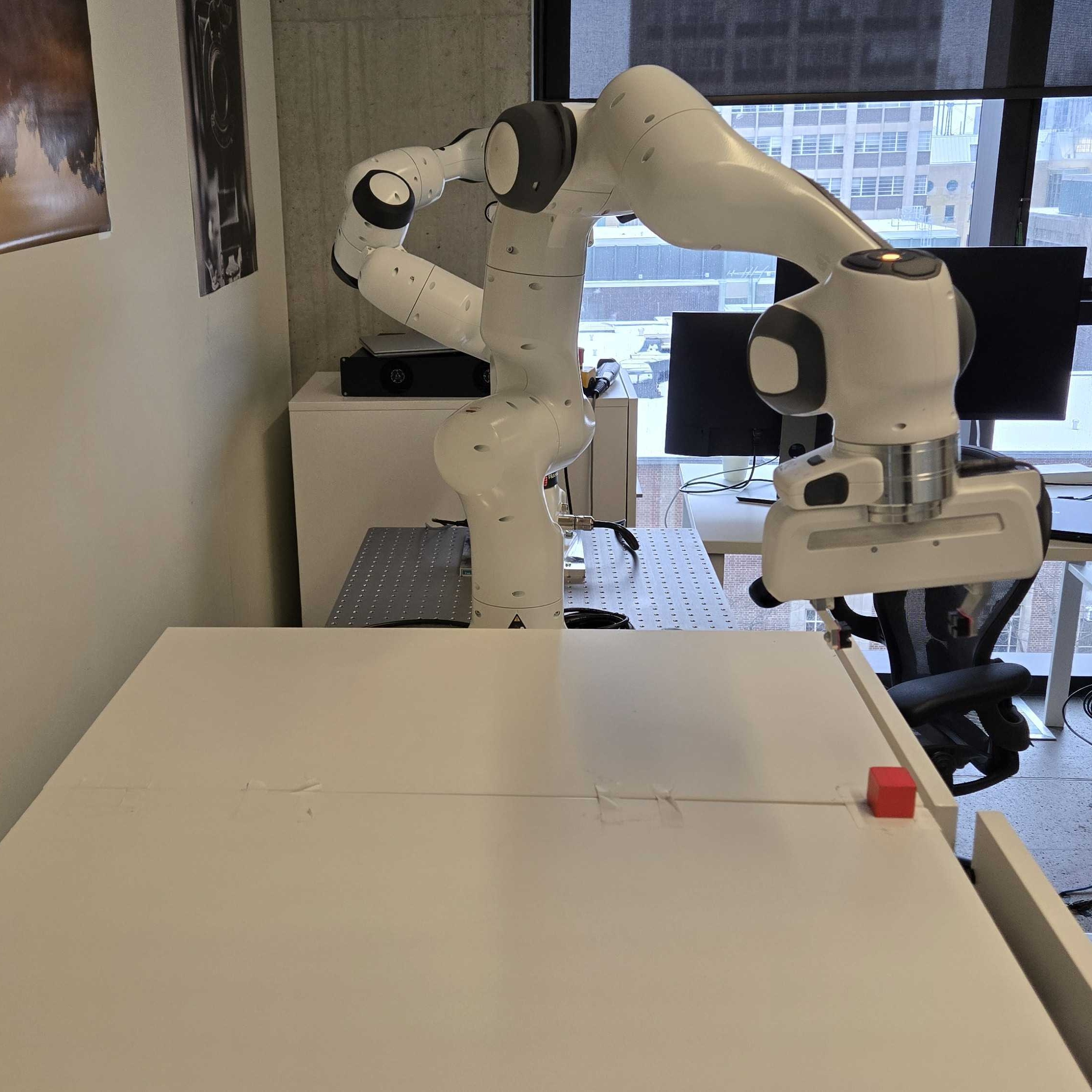}} \end{subfigure}\hspace{1mm}
    \begin{subfigure}[T]{.31\linewidth}{\includegraphics[width=\linewidth]{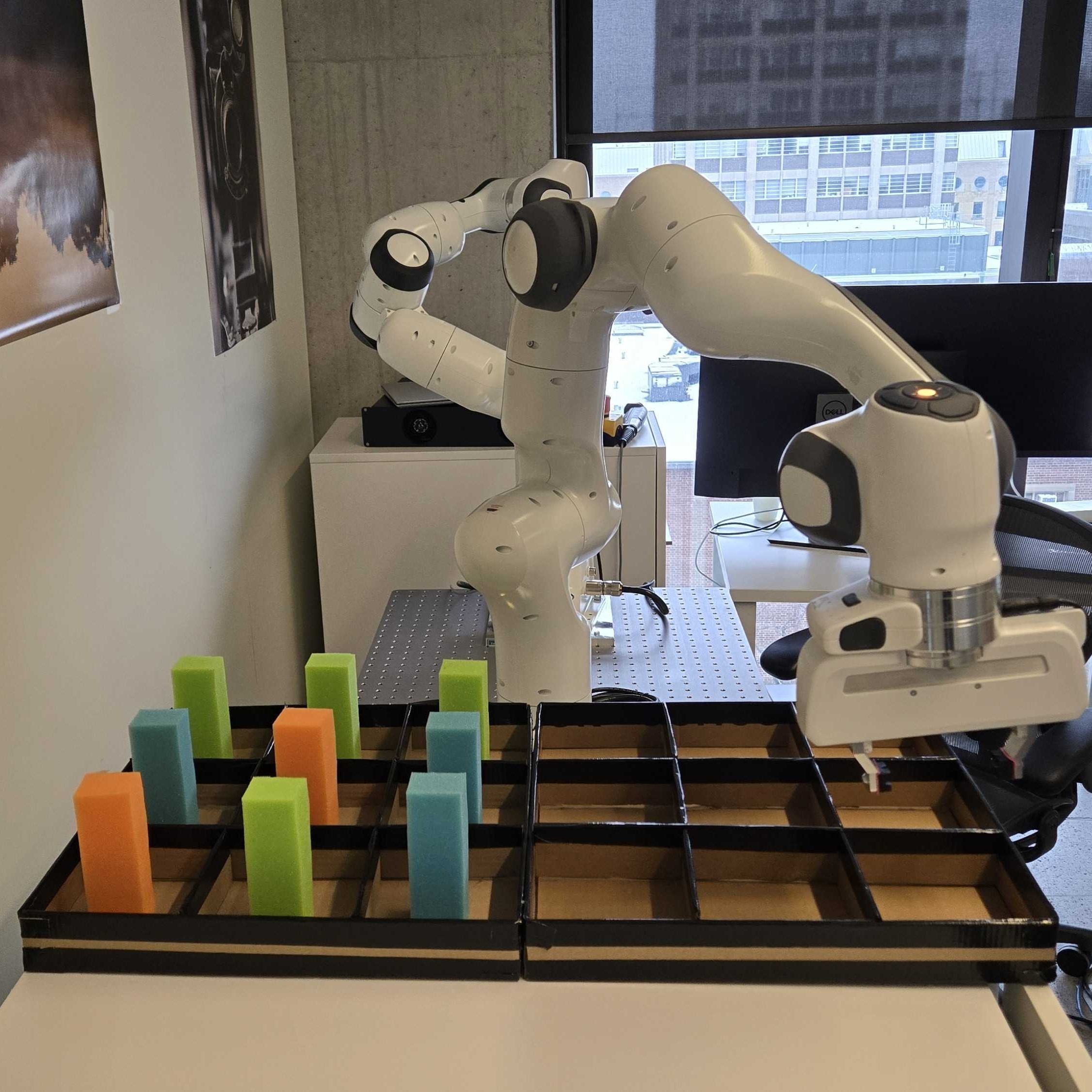}}\end{subfigure}%

\vspace*{2mm}

    \begin{subfigure}[T]{.31\linewidth}{\includegraphics[width=\linewidth]{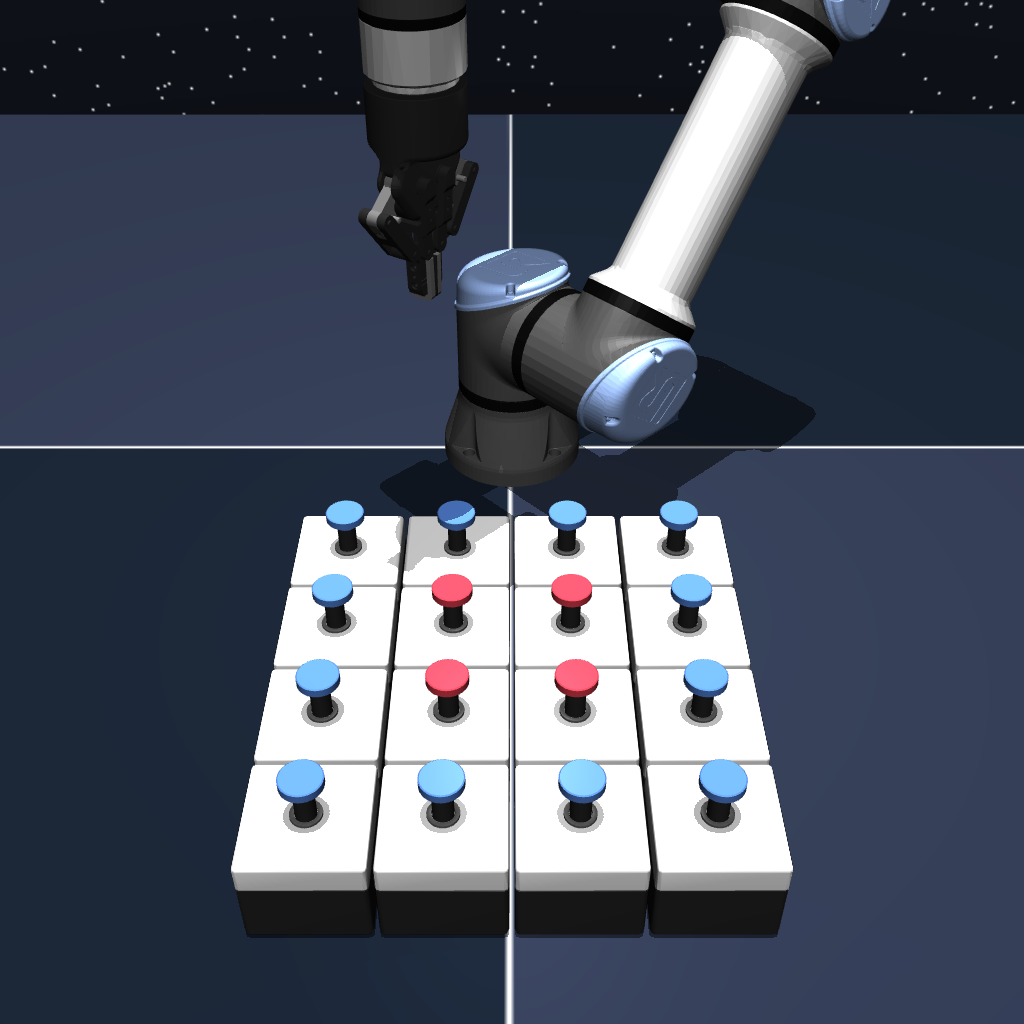}}\end{subfigure}\hspace{1mm}
    \begin{subfigure}[T]{.31\linewidth}{\includegraphics[width=\linewidth]{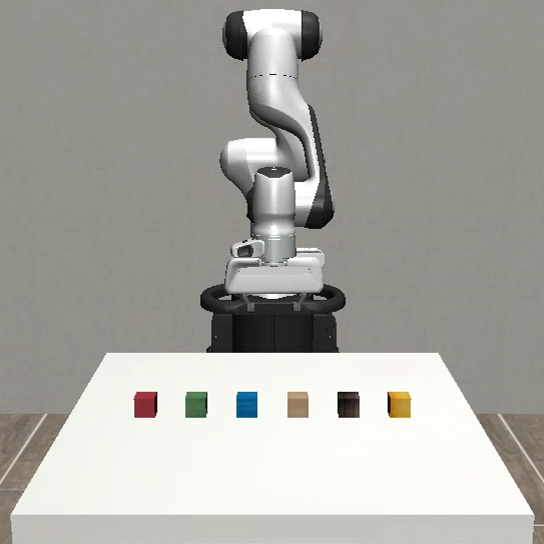}}\end{subfigure}\hspace{1mm}
    \begin{subfigure}[T]{.31\linewidth}{\includegraphics[width=\linewidth]{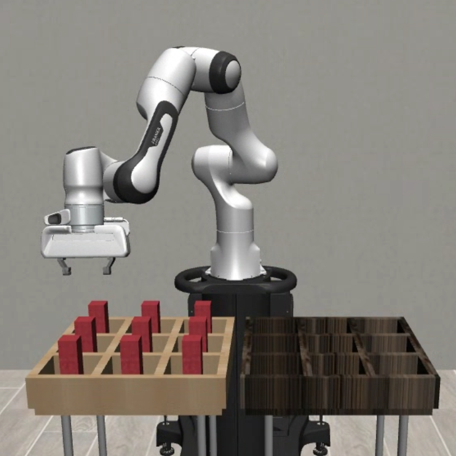}}\end{subfigure}%
    \caption{\textbf{The environments we use in our experiments.} From top left to bottom right: Maze, Didactic, Well-Plate Real, Lights, Block-Stack, Well-Plate Sim. Lights picture is taken from OGBench~\cite{park2024ogbench}.}
    \label{fig:environments}
\end{figure}

We examine diffusion composition in a maze navigation task and robotic manipulation tasks. Our environments are designed to measure \emph{test-time stitching}, with specific combinations of subskills able to be composed from the training data, but not seen in the same trajectory. Pictures of our environments are shown in Figure~\ref{fig:environments}. \textbf{Maze} is a 2D point-maze navigation task where the point agent travels from the top left to the bottom right corner, taking different routes around obstacles. Combining sub-routes into a new path requires composition. Colliding with an obstacle kills the agent, requiring precise trajectories. Figure~\ref{fig:environments} shows a $5\times5$ maze for clarity, but we used a larger $11\times11$ maze in most experiments. \textbf{Didactic} is a manipulation task where a robot arm manipulates cubes. The left side of Figure~\ref{fig:eyecatcher} shows the training three trajectories of pick and places in two separate areas, and a trajectory connecting the two. Picking the cube from one area and placing it into the other requires composition. In \textbf{Well-Plate}, an arm must relocate $3$ blocks from one $3\times3$ plate to another. Placing blocks into novel configurations requires subskill composition, both to compose pick and place sub-skills for individuals blocks together, but also the pick skill for a particular block with the place skill of another. In \textbf{Block-Stack}, an arm stacks six cubes. Not all stacking orders are seen during training, so stacking the cubes in new orders requires composition. In \textbf{Lights}, an arm presses buttons on a $4\times4$ table at random to light them up. Novel trajectories require new permutations of active lights. We implemented the Maze task in Numpy~\cite{harris2020array}, the Well-Plate Sim and Block-Stack tasks in RoboSuite~\cite{zhu2020robosuite}, and the Didactic and Well-Plate Real tasks on a real Franka Emika Robot~\cite{haddadin2022franka,elsner2023taming}. We adapt the Lights task from OGBench~\cite{park2024ogbench} with modifications.

\textbf{Data Collection for Diffusion Model Training}: For each task besides the Didactic scene we collected between $50$ and $1000$ trajectories by randomly sampling subgoal permutations and executing the corresponding trajectory through task-specific controllers. Each dataset only covers a small space of the potential legal task executions, allowing for novel generations from the diffusion planner, elaborated on in Appendix~\ref{appendix:env_details}.

\subsection{Measuring Trajectory Diversity and Novelty}
\label{section:define_novelty}
To quantify the diversity of each diffusion planner's generations, we construct a general quantitative measure of the novelty of a trajectory. Our metric uses the structure of the task and training data to determine how novel a given trajectory is. 

Assume that each environment has a state represented by a set of primitives $o_1,o_2,...,o_n \in O$. Each trajectory modifies the state of the environment such that the final state can be represented by a set of these primitives. We manually design our environments around primitives that are easy to extract from the environment post execution of the trajectory. For instance, in our Maze task the primitives correspond to individual intersections in the maze. Each unique path through the maze is represented by the set of intersections it visits. Appendix~\ref{appendix:env_diversity_details} elaborates on the primitive set for each environment.

Applying this to each training trajectory means the training set can be represented by a list of primitive sets $\tau_{\text{train}_1},\tau_{\text{train}_2},...$. When we generate a trajectory with a diffusion planner and execute it in the environment, it produces a new primitive representation $\tau_{\text{eval}}$.

If $\exists \tau_{\text{train}_i}$ s.t. $\tau_{\text{eval}} \subseteq  \tau_{\text{train}_i}$, then we say that $\tau_{\text{eval}}$ is \textbf{memorized}. This is because the generated trajectory is entirely composed of sub-trajectories seen together in a trajectory previously, and so no composition had to be performed. Otherwise, it is \textbf{novel} because producing that trajectory required composing elements from two or more training trajectories together.

If $ |\tau_{\text{eval}}|\ge |\tau_{\text{train}_i}| \forall_i $ (meaning the evaluation trajectory has the same number of primitives as the training trajectories), then we say that $\tau_{\text{eval}}$ is \textbf{complete}. This is because the generated trajectory performed a number of subtasks equal to the ones seen in each training trajectory. Otherwise, it is \textbf{partial} because it only completed a smaller number of subtasks. Because we fix the number of subtasks each training trajectory performs, this lets us easily tell if a generated trajectory performs less subtasks than desired. The ideal compositional diffusion planner should produce trajectories that are both \emph{novel} and \emph{complete}.

\subsection{Algorithm Setup} 
For our experiments we use a 1D U-Net~\cite{janner2022planning} and DiT~\cite{dong2023aligndiff} diffusion planners~\cite{janner2022planning} as the base models. We use CleanDiffuser's~\cite{dong2024cleandiffuser} implementation and hyperparameters. During inference, we predict state trajectories\cite{ajay2023is} and use a Cartesian operational space controller~\cite{khatib2003unified} to follow the trajectory. Appendix~\ref{appendix:hyperparameters} contains more details.

\section{Experiments}
\label{section:experiments}

Our hypothesis is that local receptive fields and shift equivariance are key ingredients in explaining compositionality in diffusion planning. We expect that different ways of incorporating each (training augmentations, replanning, architecture choices, etc.) should increase compositionality. However, it is not known which of the two properties is more important, whether both are needed, or which way of inducting the two properties increases compositionality the most effectively. It is also not known how scaling data is related to local receptive fields. Our experiments aim to address these questions.

\subsection{How Does Architecture Affect Composition?}

\begin{figure}[b]
\centering
  \includegraphics[width=.98\linewidth]{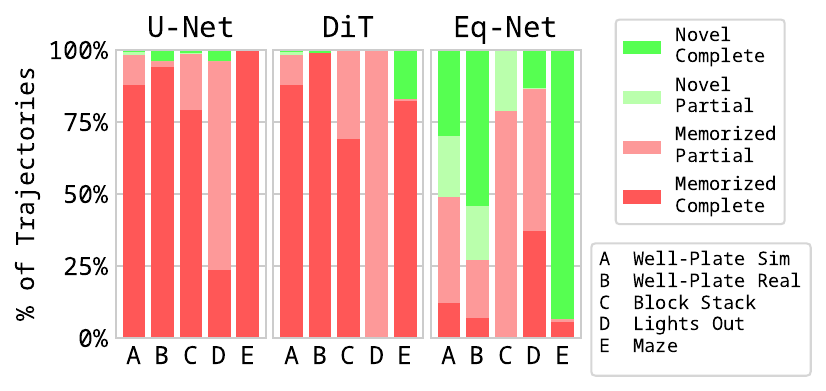}
\caption{\textbf{Architecture choices enable compositionality}. Experiments across all environments show using local and positionally equivariant architecture substantially increases compositionality, with a greater number of novel trajectories.}
\label{fig:architecture}
\end{figure}

Our primary question is if the lack of local receptive fields and shift equivariance in diffusion planner architectures hurts compositionality. We trained U-Net, DiT, and Eq-Net on all of our environments, letting us directly compare the effects of architectural biases on composition. We ran the resulting planners between $100$ and $10000$ times in each environment, and calculated the diversity and completeness of each trajectory. Figure~\ref{fig:architecture} shows the results on five of our six environments\footnote{Due to Didactic only being completion-based, we defer results to Table~\ref{table:gc_results}.}, where we report the percentage of model samples that exhibit completeness (completing the same number of subtasks as in each training sample) and novelty (whether samples must have combined elements from different training data samples). Eq-Net shows increased novelty in all environments tested, confirming our hypothesis on the importance of local receptive fields and shift equivariance.

\subsection{Locality vs Shift Equivariance: How Much Do They Matter?}

\begin{figure}[t]
\centering
  \includegraphics[width=.98\linewidth]{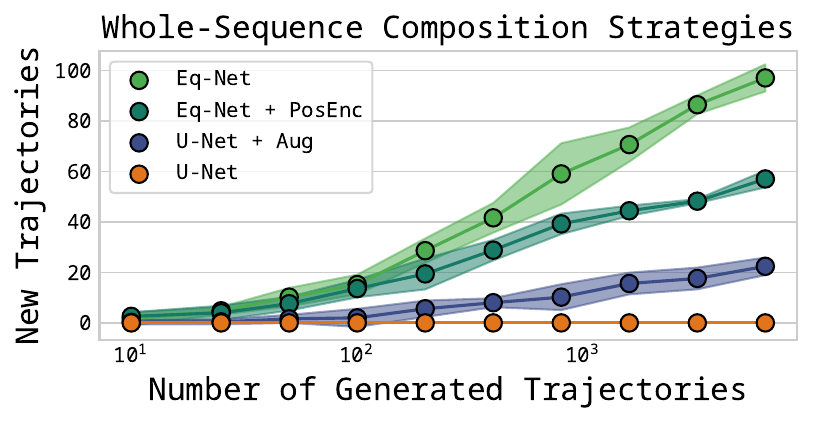}
    \caption{\textbf{Locality matters more than shift equivariance}. Our results show both local receptive fields \emph{and} shift equivariance play a role in encouraging composition. The model with both (Eq-Net) outperforms only local receptive fields (Eq-Net + PosEnc), only shift equivariance (U-Net + Aug) and neither (U-Net). However, local receptive fields have the greater marginal effect.}
    \label{fig:comp_strategies}
    
\end{figure}

Eq-Net is designed to have local receptive fields and be shift equivariant, meaning it is not clear which of the two properties is most important.  We trained additional models in a $5\times5$ version of our Maze environment, one with only local receptive fields and the other with only shift equivariance. Specifically, we train a non-local model to have approximate shift equivariance through training augmentation (U-Net+Aug, described in Fig~\ref{fig:method}) and a local model without shift equivariance by adding sinusoidal positional embeddings~\cite{vaswani2017attention} (Eq-Net+PosEnc, described in Appendix~\ref{appendix:posenc}). Figure \ref{fig:comp_strategies} shows how the diversity of each model's outputs increases as more samples are collected. Notably, Eq-Net+PosEnc performs better than U-Net+Aug, indicating that local receptive fields are a \emph{comparatively more important factor} than shift equivariance. However, the combination of both (Eq-Net) achieves the best performance, indicating that both ingredients are crucial. We show additional experiments that increasing Eq-Net's kernel size decreases compositionality in Appendix~\ref{appendix:kernel_tuning}, validating our theory that receptive local fields are important for enhanced composition.

\subsection{Replanning During Inference vs Architectural Design}



\begin{figure}[t]
\centering
  \includegraphics[width=\linewidth]{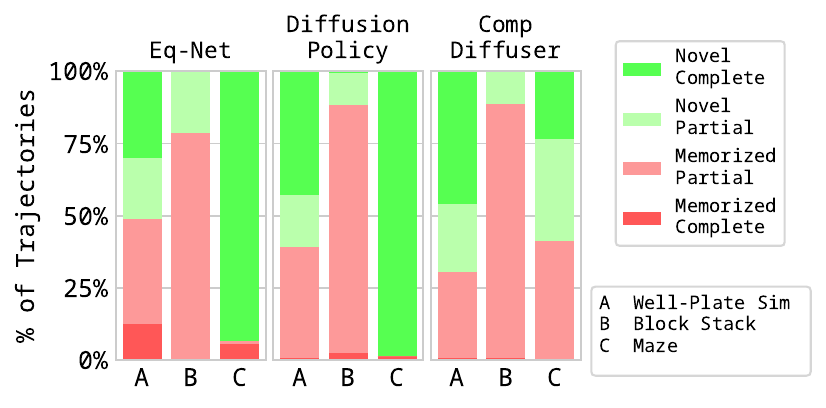}
    \caption{\textbf{Eq-Net is competitive with other compositional diffusion planners}. When comparing Eq-Net to Diffusion Policy~\cite{chi2023diffusion} and CompDiffuser~\cite{luo2025generative}, we found that performance is similar. This suggests that the underlying ingredients allowing for composition are shared between the three algorithms, but expressed in distinct ways. Eq-Net achieves local receptive fields and shift equivariance through \emph{architecture}, Diffusion Policy through \emph{replanning}, and CompDiffuser through \emph{parallel chunking}.}
    \label{fig:replanning}
\end{figure}

As discussed in Section~\ref{section:comp_methods}, compositionality from generating smaller chunks of a trajectory, either through replanning or a hierarchical planning method, is a common way to enhance compositionality in diffusion planners. Our theory predicts these algorithms will outperform monolithic planning, as their inference and training methods enforce local receptive fields and positional equivariance. We compare Eq-Net to visual Diffusion Policy~\cite{chi2023diffusion}, which generates short-horizon chunks and frequently replans, and CompDiffuser~\cite{luo2025generative}, which generates several chunks conditioned on their neighbours which are then stiched into a longer plan. We trained both algorithms with their original hyperparameters on our Robosuite and Navigation tasks. Results are shown in Figure~\ref{fig:replanning}. As expected, these chunking methods compose well, performing comparably to Eq-Net on our manipulation tasks. CompDiffuser struggles on the Navigation environment, as the chunking strategy often led to sub-par trajectory quality along the overlap points between trajectories which would guide the point agent into obstacles. This shows that monolithic planners can preserve overall trajectory fidelity that chunking methods struggle with.

\subsection{Compositionality via Data Scaling vs via Architecture}
\label{section:data_results}

\begin{figure}[b]
\centering
  \includegraphics[width=.98\linewidth]{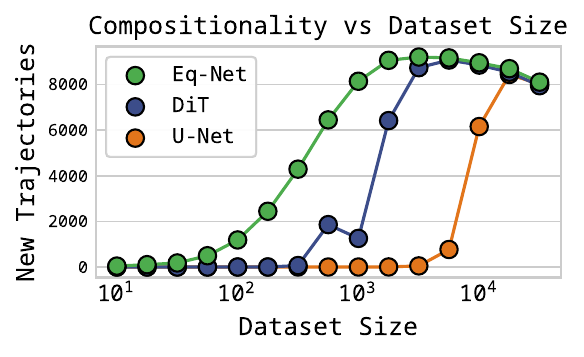}
\caption{\textbf{Scaling training data increases compositional generalization}. As training data scales, all network architectures eventually learn to produce composed trajectories, although the point at which this happens depends on the architecture. Notably, Eq-Net learns to generalize with much less data than either U-Net or DiT. Note that creativity saturates as a greater proportion of the possible trajectory space is covered by the training set.}
\label{fig:data_scaling}
\end{figure}

To confirm that data scaling can increase compositional generalization, we produced larger and smaller versions of our Maze dataset. As expected, we found that simply increasing the amount of training data was often sufficient to increase a model's ability to compose. Figure \ref{fig:data_scaling} shows that all models learn to compose with enough training data. However, Eq-Net's built-in architectural biases help it generalize faster than the models without inductive biases.
To examine how scaling training data relates to local receptive fields, we examined the receptive field of our models trained at different data scales. Following prior work~\cite{niedoba2024towards,kamb2024analytic}, we calculate the gradient of an intermediate sample point with respect to the input sequence. 

Formally, for a diffusion planner $x_{t+1} = f(x_t,t)$ where $x_{t+1}$ is the next de-noised sample  $x_t$ is the previous sample, the samples are length $H$ and state are indexed by $x_t[i]$, and $t$ is the denoising step, we calculate $|\frac{\partial x_{t+1}[i]}{\partial x_t}| = |[\frac{\partial x_{t+1}[i]}{\partial x_t[1]},\frac{\partial x_{t+1}[i]}{\partial x_t[2]},...,\frac{\partial x_{t+1}[i]}{\partial x_t[H]}]^T|$ for each diffusion time-step from $1 \to T$. We compute these gradients through the diffusion model exactly using auto-differentiation.

\begin{figure}[b!]
\centering
  \includegraphics[width=.98\linewidth]{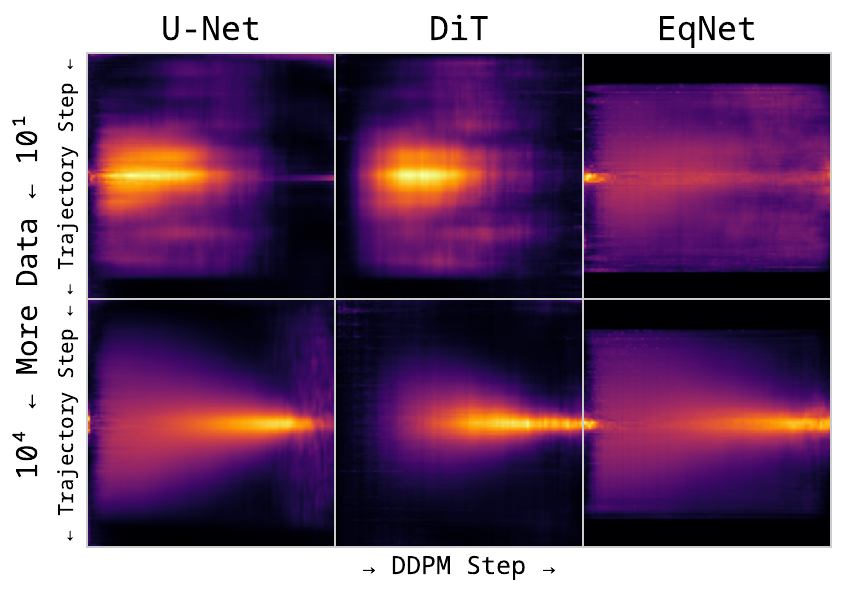}
\caption{\textbf{Scaling training data shrinks the receptive field of the diffusion model}. Models trained on different amounts of training data behave differently late in the DDPM generation process. We sample diffusion planners trained on increasing data amounts 50 times. Next, we calculate the trajectory gradient with respect to the centre state in the trajectory. Each figure shows this gradient normalized over the average gradient value for the whole sampling process. In late DDPM steps, models trained on more data have a local receptive field, while the ones trained on less data have more global receptive fields. }
\label{fig:grads}
\end{figure}

Because diffusion planners generate a 1-D sequence, we can plot these gradient maps as an image, where the x-axis is the denoising step and the y-axis represents the sequence gradient from the start to the end of the trajectory. Figure~\ref{fig:grads} shows the gradient maps for each architecture trained on a small amount of data ($10$ samples) and a large amount of data ($10000$ samples). To gain a lower-variance estimate, we average this gradient over $50$ samples.

These show that the receptive field of the models trained on less data are larger for the non-local models (U-Net and DiT) than for the local Eq-Net, but as the amount of training data is increased the receptive fields look more similar and local. In particular, the receptive field starts out large early in the reverse diffusion process, but shrinks later in generation, with more shrinkage occurring for models trained on more data. We present the gradient maps for all models in Appendix~\ref{appendix:grad_maps}.

To create a single quantitative measure of how local each model is, we follow prior works~\cite{niedoba2024towards} and estimate the size of the receptive field by measuring the size of the centred window required to capture $50\%$ of the gradient magnitude. To estimate the total non-local behaviour of each model, we take the integral over the receptive field size across the entire trajectory generation.

\begin{figure}[]
\centering
  \includegraphics[width=.98\linewidth]{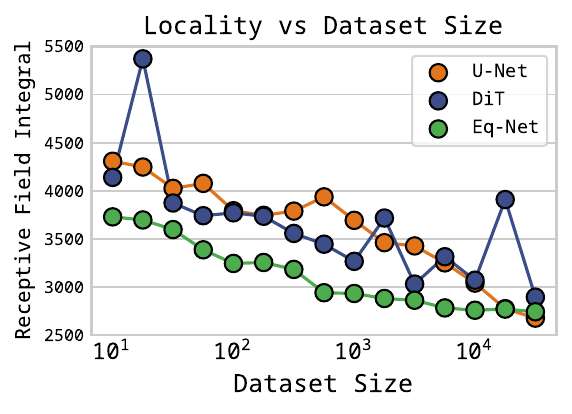}
\caption{\textbf{Scaling size of training data induces locally receptive fields}. When comparing on our final quantitative measure of a model's local receptive field strength, models trained on more data are generally more local than those trained on less. Eq-Net, which has an additional architectural bias towards local receptive fields, is far more local than models without a built-in inductive bias.}
\label{fig:total_locality}
\end{figure}

Figure~\ref{fig:total_locality} shows this measure of the strength a model's local receptive fields for the three architectures at different data scales. There is a clear trend where increasing the amount of training data increases a model's local receptive fields. Eq-Net also shows stronger local receptive fields across the board than U-Net and DiT, confirming the role that Eq-Net's smaller kernels have in enabling stronger compositionality.

Taken together, our findings show that even when a diffusion planner does not have inductive biases towards local receptive fields, scaling data will teach the model to have local receptive fields. Combined with recent findings that diffusion models learn different biases from different datasets~\cite{lukoianov2025locality}, this suggests Eq-Net takes advantage of the fact that trajectory data is statistically more locally biased than globally biased. This is coherent with the fact that the rough Markovian property of trajectory data means individual states more strongly correlate with nearby states, and that in mixed behaviour/multi-task datasets long-horizon state correlations are weak.

\subsection{Are More Compositional Models More Steerable?}

One of the motivations behind planners with compositional generalization is that creating more diverse plans may be important when planning to solve a specific goal. While many guidance approaches exist~\cite{lu2025makes}, most works in diffusion planning use some form of conditioning to steer the model towards plans that solve a particular task~\cite{janner2022planning,ajay2023is}. To understand if our more diverse models are more guidable, we examined guidance in a variant of our Maze environment (Appendix~\ref{appendix:grid_gc_env_details}) and in a didactic manipulation environment (Section~\ref{section:env}). For both tasks we guide the models with inpainting like in prior diffusion works~\cite{lugmayr2022repaint,janner2022planning,sohl2015deep}, where the starting state and desired ending or intermediate states are inpainted to the generated trajectory and fixed during denoising.

\begin{figure}[t]
\begin{subfigure}[t]{.5\linewidth}
    \centering
    \includegraphics[width=\linewidth]{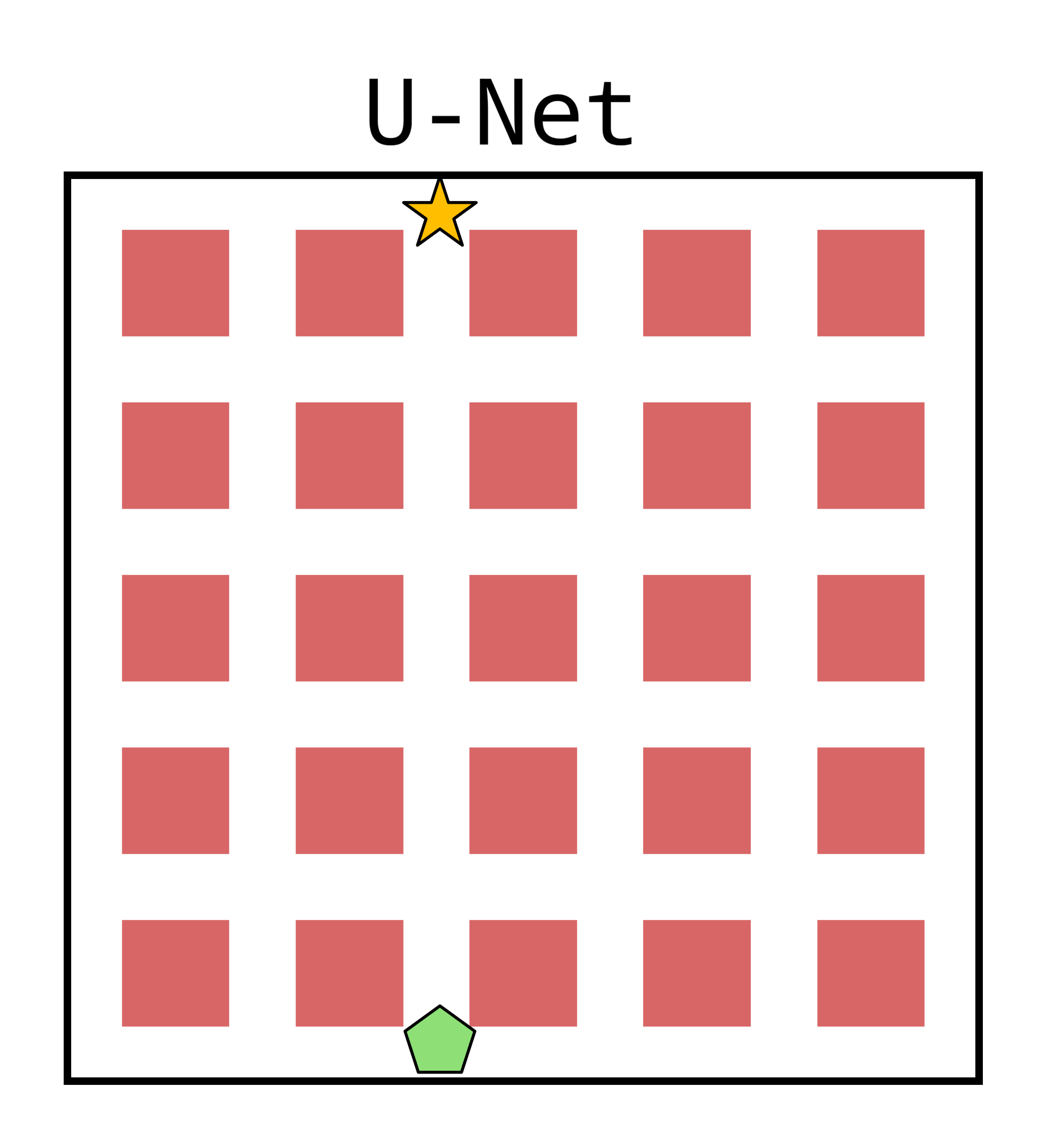}
    \end{subfigure}%
\begin{subfigure}[t]{.5\linewidth}
    \centering
    \includegraphics[width=\linewidth]{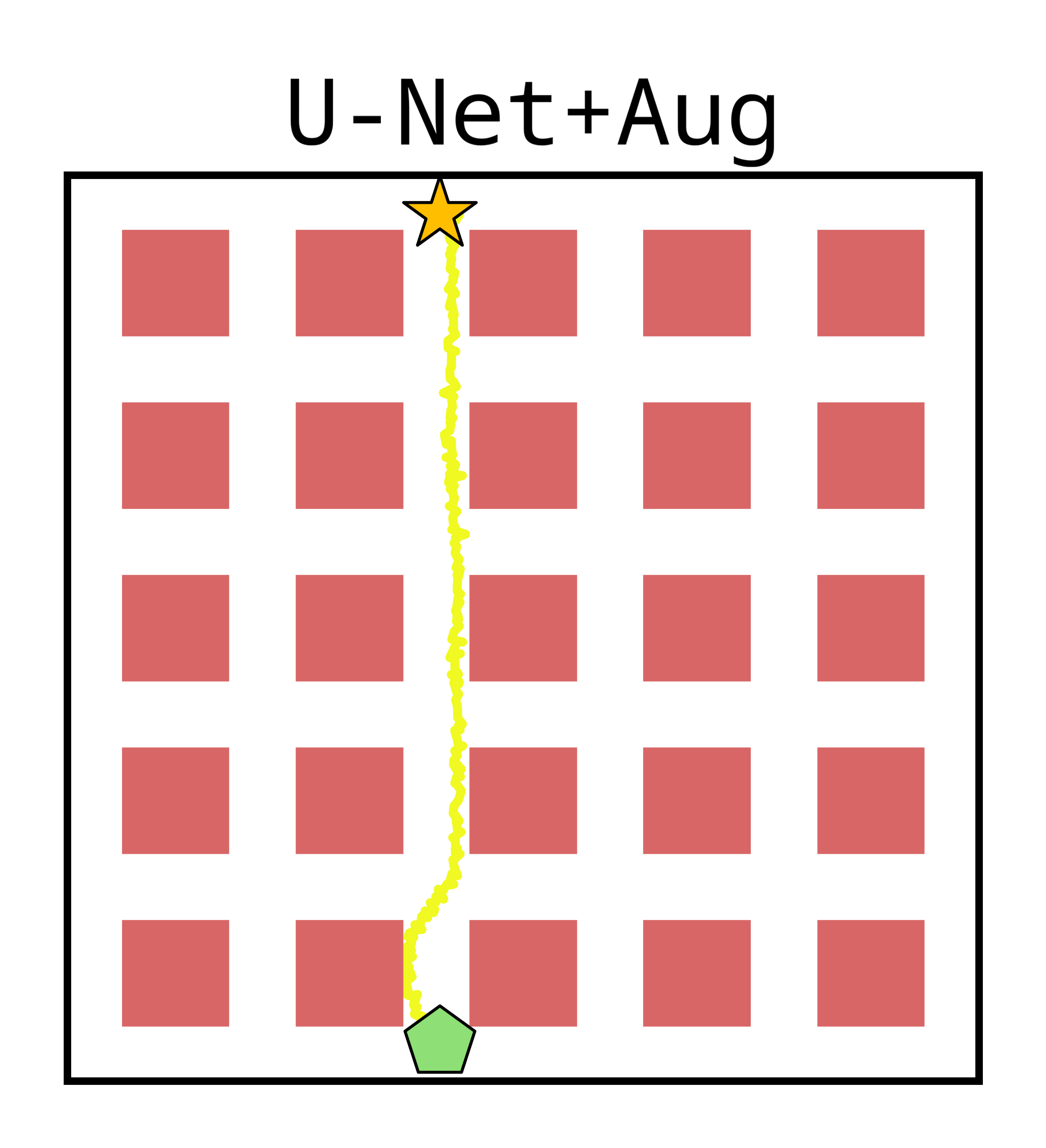}
    \end{subfigure}%
\vskip\baselineskip
    
\begin{subfigure}[t]{.5\linewidth}
    \centering
    \includegraphics[width=\linewidth]{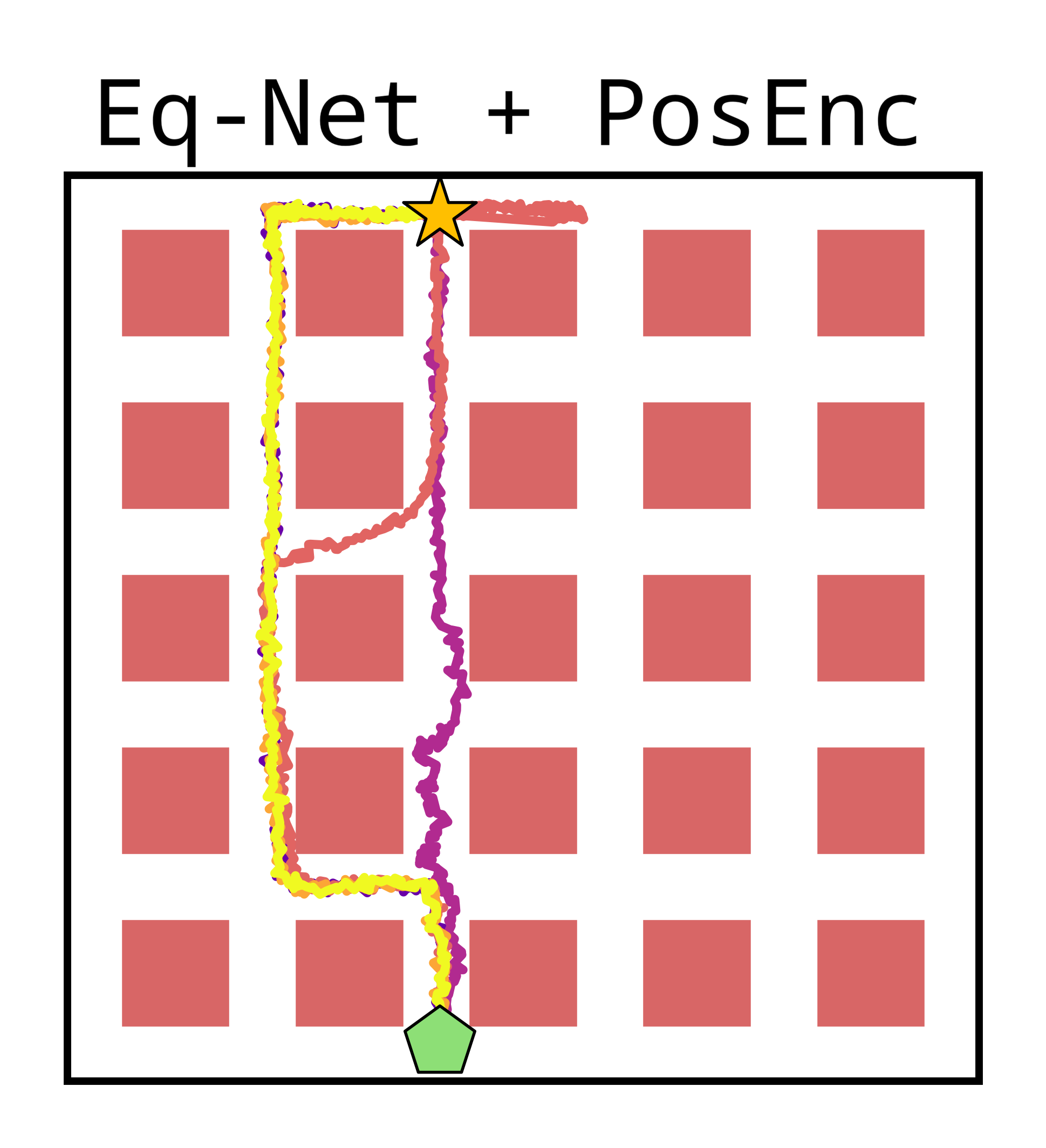}
    \end{subfigure}%
\begin{subfigure}[t]{.5\linewidth}
    \centering
    \includegraphics[width=\linewidth]{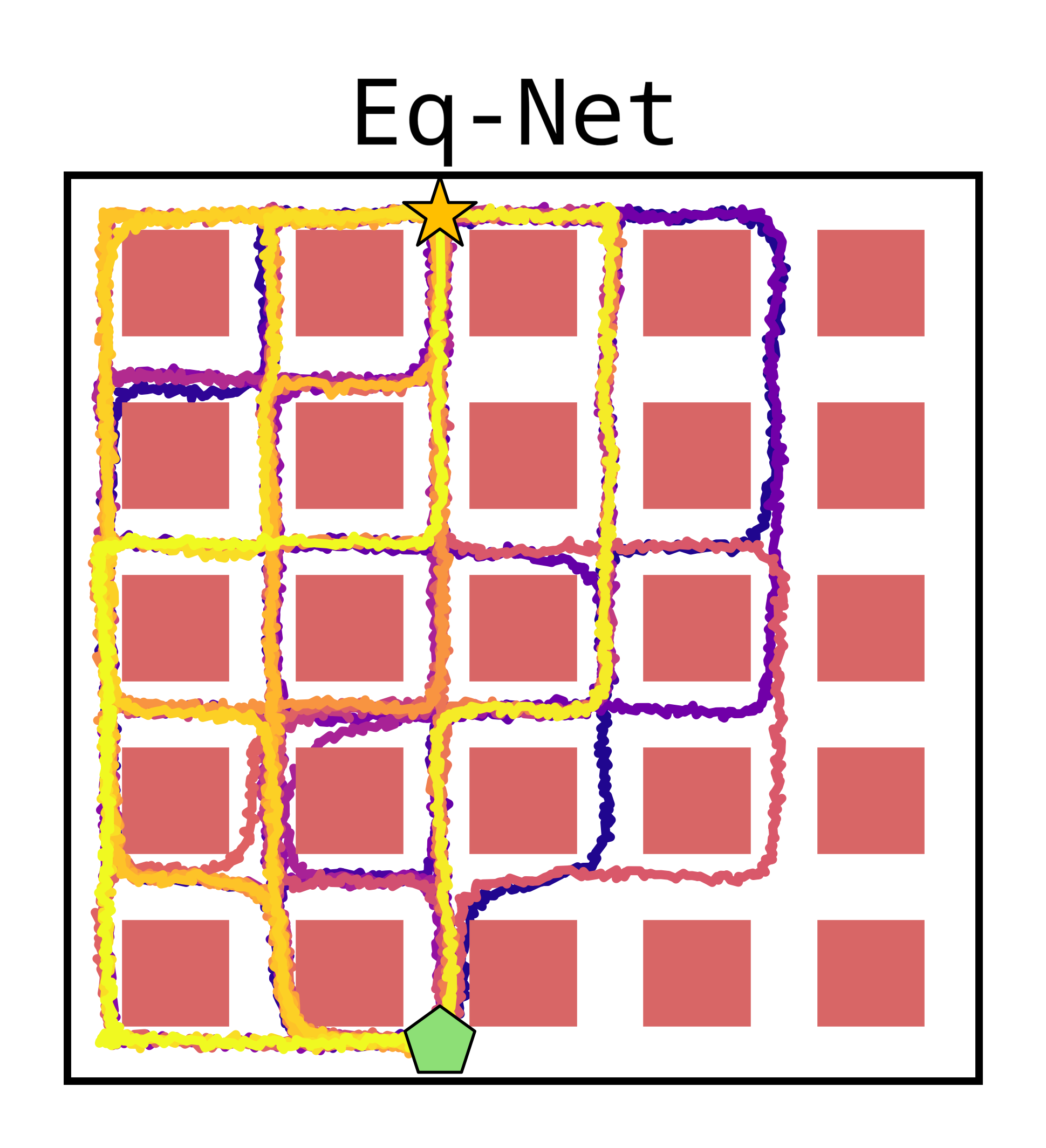}
    \end{subfigure}%
  \caption{Successful trajectories from a random sample of $50$ runs on one of the unseen start-goal pairs in our goal-conditioned environment. Eq-Net is able to produce a far greater diversity of candidate solutions than any other planner. Each colour represents a different generated trajectory.}
  \label{fig:gc_demo}
\end{figure}

Results for both goal-conditioned tasks are shown in Table~\ref{table:gc_results}. Eq-Net far outperforms all other models in Maze, achieving an average $83.3\%$ completion rate, roughly $7.5 \times$ higher than the model with only local receptive fields but not shift equivariance. To visualize the difference in composition for each model, we plotted the unique successful runs from $50$ random generations from each algorithm on one of the held-out start-goal pairs in Figure \ref{fig:gc_demo}. Similarly, Eq-Net was the only architecture capable of producing a successful trajectory in our didactic environment, one of which is shown in the right-hand side of Figure~\ref{fig:eyecatcher}.

\begin{table}[]
\centering

\begin{tabular}{>{\kern-\tabcolsep}lllll<{\kern-\tabcolsep}}
\toprule
\rowcolor[HTML]{ECF4FF} 
                           &       & \textbf{Maze-GC}  &               &                     \\ \midrule
Model                      & U-Net & U-Net/Aug         & Eq-Net/PosEnc & Eq-Net              \\ \midrule
Completion  $\uparrow$ & 0±0   & 0.44±0.0       & 13.39±5.0    & \textbf{83.8±6.4} \\
Unique Paths $\uparrow$    & 0±0   & 0.09±0.0            & 11.16±1.0    & \textbf{116.8±4.8} \\ \midrule
\rowcolor[HTML]{ECF4FF} 
                           &       & \textbf{Didactic} &               &                     \\ \midrule
Model                      & U-Net & DiT               & Eq-Net         &                     \\ \midrule
Completion  $\uparrow$ & 0     & 0                 & \textbf{77}   &                    
\end{tabular}
\caption{\textbf{Eq-Net is more steerable}: Results in our two goal-conditioned environments show Eq-Net able to reliably complete both tasks. Other augmentation and architecture strategies achieved less compositional generalization and suffered from worse completion rates. Completion is reported as a percentage and unique paths as a count.}
\label{table:gc_results}

\end{table}

In total, our goal-conditioned tasks show that models with compositional generalization are indeed more steerable, giving our findings use in guided planning methods.

\subsection{Remaining Factors for Diffusion Planner Composition}

While we cannot rule out other potential influences on compositionality, we found that some other phenomena had a small effect. In particular, we found
\begin{itemize}
\item Early stopping helps compositionality somewhat as seen elsewhere~\cite{bonnaire2025diffusion}, but model bias matters much more.
\item The number of denoising steps greatly affects trajectory completion rates, but not compositionality.
\item Other algorithmic and parameter choices, such as using the discrete or continuous stochastic differential equation formulation, the diffusion sampling temperature, predicting noise versus clean samples. etc. did not make a difference.
\end{itemize}
We defer further experimental results to Appendix~\ref{appendix:more_experiments}.

\section{Discussion}

\subsection{Explaining Composition in Other Diffusion Planners} 
\label{section:whichisbest}
Our goal in examining the factors that lead to composition is to help explain why prior works compose (or fail to). We include a thorough discussion of other works in Appendix \ref{appendix:other_diffusion_planners}. We find the majority of existing works achieve local receptive fields and shift equivariance either through frequent replanning~\cite{song2025history,chen2024diffusion,chi2023diffusion} or some form of hierarchical decomposition where sub-trajectories are generated and then composed manually \cite{chen2024simplehierarchicalplanningdiffusion,luo2025generative,li2023hierarchical}. This suggests that the form of composition seen in image diffusion models is \emph{mechanistically different} from how most diffusion planners achieve composition, as most image diffusion models do not manually stitch together patches/frames. We hope our work proves the viability of a simple, purely architectural approach. We provide additional discussion and recommendations in Appendix \ref{appendix:discussion}.



\subsection{Conclusions, Limitations, and Future Work}

Our work has several limitations. First, we only consider environments carefully designed to require stitching. While we motivate this in Section \ref{section:env}, evaluating our findings in multi-task or even multi-embodiment robotics dataset would be useful future work. Second, our guidance strategy was limited to inpainting, which is effective in environments with clear endpoint goals, but less useful when goals are abstract (such as with language-conditioned agents) or multi-modal. Incorporating a way to leverage other guidance strategies like classifier-based~\cite{ho2020denoising} or free~\cite{ho2021classifierfree} guidance into our proposed high-compositionality diffusion planners is potentially fruitful future work.

\section*{Acknowledgements}

We thank the Vector Institute for computing resources. We thank Nathan Samuel de Lara, James Ross, Hossein Goli, Qi Chen, Brandon Huang, Giseung Park, and others for helpful comments on drafts and conversations.


\bibliographystyle{plainnat}
\bibliography{references.bib}
\clearpage
\section{Appendix}
\resumetoc
\renewcommand*\contentsname{Appendix Table of Contents}

This supplementary appendix includes additional information on the experiments presented in the main text, additional experiments and ablations, more discussions of related works and findings of our paper, and some explanatory exposition for certain concepts.

\tableofcontents

\subsection{More Related Work}

\textbf{Stitching through RL}: Historically, stitching through bootstrapping has been seen as one of the primary benefits of temporal difference (TD) learning~\cite{kumar2022should} in reinforcement learning (RL). Planning with diffusion models offers some of the stitching benefits of TD learning without needing to handle RL-specific challenges such as the deadly triad~\cite{irpan_2018,engstrom2019implementation,van2018deep} and slow temporal credit assignment~\cite{chen2021decision}.

\textbf{Architecture in Diffusion Planning:} Another work that similarly analyzes architecture and inference strategies for diffusion planning is \citet{lu2025makes}, which interestingly finds that the best diffusion backbone are transformers, which have \emph{no} inductive biases towards locality and usually have positional attention added. However, their work is about general performance in offline RL and not about emphasizing compositionality, which as discussed may not be an important properly for strong performance on the D4RL benchmark due to a lack of required compositionality \cite{ghugare2024closing}. 

\textbf{Diffusion Planner Inference Choices}: When analyzing replanning, we analyze the most common form where replanning happens after a fixed number of execution steps. However, more sophisticated approaches to replanning exist which may have different effects on composition. For instance,~\citet{zhou2023adaptive} adaptively re-plans when the estimated likelihood of the current partially executed trajectory falls too low. 

\textbf{Other Forms of Compositionality:} In our analysis of data scaling, we specifically analyze stitching as a property of data, but other works analyze other forms of generalization including concept composition~\cite{okawa2023compositional}, consistency~\cite{zhang2024emergence} and grammar~\cite{favero2025compositional}. 

\textbf{Stitching in Non-Diffusion Imitation Learning:}  Prior to diffusion planners, other generative models were common in imitation learning~\cite{ho2016generative}. Most modern approaches are either based around autoregressive decoder-only transformers \cite{radford2018improving}, such as the Decision Transformer \cite{chen2021decision}, ACT \cite{george2023one}. Other works analyzes stitching in imitation learning include these works~\cite{char2022bats,hepburn2022model}. We were inspired by~\citet{ghugare2024closing}, which analyzes stitching through a compositional generalization perspective. They revealed that applying supervised-learning as a form of offline RL~\cite{emmons2022rvs} fails to stitch, requiring training data augmentation which manually stitches together trajectories. However, their result does not directly translate to ours because the properties we analyze (local receptiveness and shift equivariance) both enforce a form of regularization that creates composition, even when such behaviour is not explicitly built into the data.

\textbf{Other Works Examining Diffusion Planner Composition:} Several other works have observed that Diffusion models often produce stitched trajectories when frequently replanning during inference~\cite{chen2024diffusion} and limiting the history the next generation is conditioned on ~\cite{song2025history}. Our work attempts to formalize these qualitative observations with more robust experiments and explain it under a single conceptual framework (locality for short history and shift equivariance for the replanning process) that also extends to architectural or training choices. 

\subsection{More Method Details}
\label{appendix:more_method}

\subsubsection{Information Propagation Across a Sequence in Diffusion}
\label{appendix:propogation}

\begin{figure*}
\centering
  \includegraphics[width=.90\linewidth]{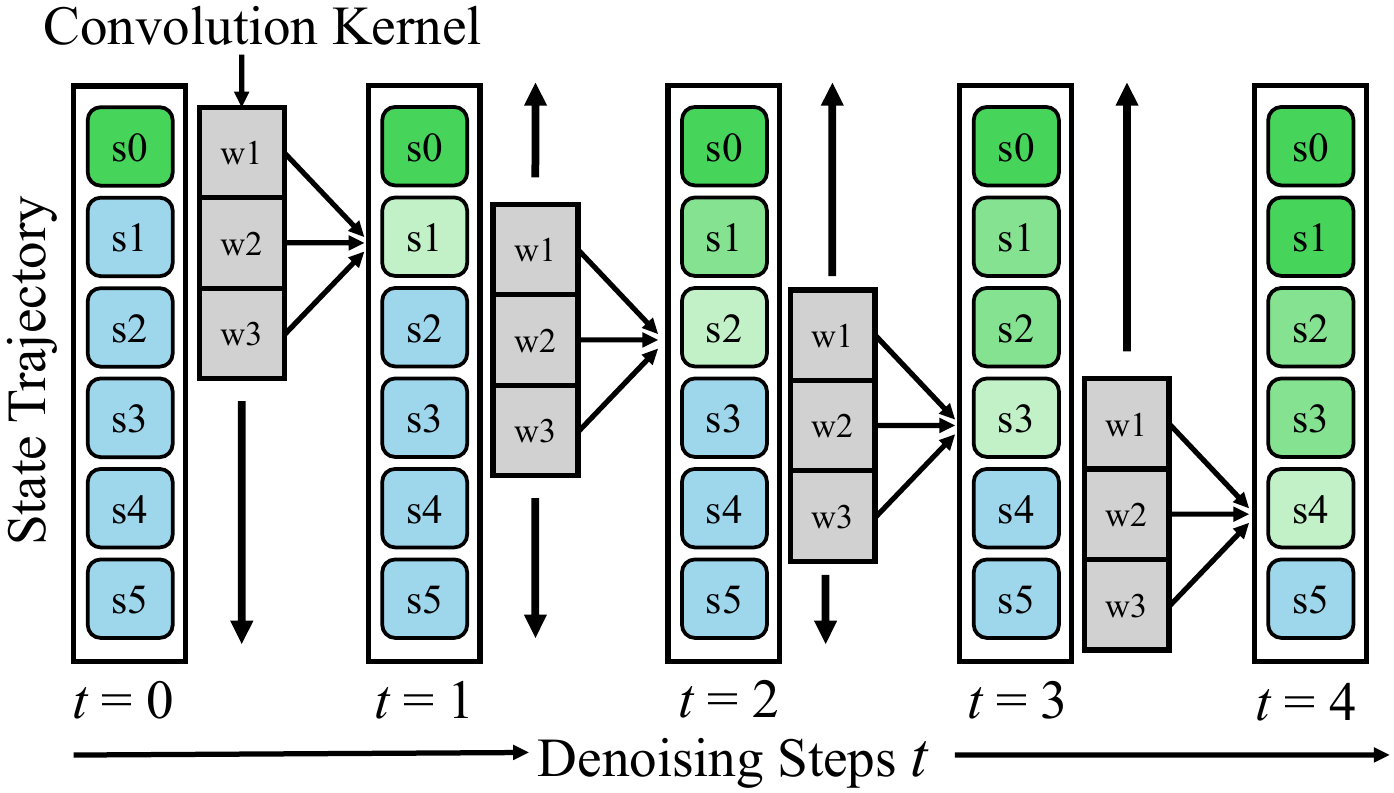}
\caption{A visualization of how information during diffusion denoising can be propagated by locally receptive models. Here, we imagine a toy setting with a sequence of length $6$, being denoised by a model consisting of just a single convolutional kernel with a receptive field of size $3$ and stride $1$. State s0 has some information that must be considered when denoising the rest of the sequence. Even though the later states in the sequence cannot receive information from s0 during a single denoising step, over several diffusion steps information propagates across the state trajectory. Note that this does not just happen in one direction down the trajectory, but can also happen up it as each convolution happens in parallel.}
\label{fig:propogation}
\end{figure*}

To visually illustrate how information can be propagated across a sequence in diffusion, we provide an illustration in Figure~\ref{fig:propogation}.

\subsubsection{How does shift equivariance work for diffusion policies?}
\label{appendix:pos_enc_dp}
Diffusion planners can condition their plans on past state observations in two ways: either by including it as an explicit condition to the diffusion network, usually combined with classifier-free guidance (CFG)~\cite{ho2021classifierfree}, or by inpainting $H_P$ states in the new generated trajectory. Full memory with inpainting does \emph{not} guarantee shift equivariance, as each state in the trajectory can end up only being drawn to the same position in the trajectory that they were seen in during training. However, replanning with a smaller history shifts the viewed positions of a particular segment of the trajectory over by $H_P$ steps. Paradoxically, this means models with strong replanning capability must already have some form of shift equivariance. In practice however, because most diffusion policies are trained on a generation length much shorter than the horizon length (i.e, $H_F <<< T$) most of this equivariance comes from randomly sampling intermediate chunks of the trajectory during training.

\subsection{Experiment Details}

\subsubsection{Baseline Descriptions}
\label{appendix:baseline}

For our baselines, we used the same core Diffusion Planning algorithm~\cite{janner2022planning} with U-Net~\cite{janner2022planning} and DiT~\cite{dong2023aligndiff} backbones, Diffusion Policy~\cite{chi2023diffusion}, and CompDiffuser~\cite{luo2025generative}. We used the implementations of Diffuser with U-Net and DiT and Diffusion Policy from CleanDiffuser~\cite{dong2024cleandiffuser} with minimal changes, and used our own implementation of CompDiffuser. Changes to each are elaborated on below:

For Diffusion Policy, we condition on a stack of past proprioceptive observations but only on the current image. We did not perform any image augmentation to the images. 

For CompDiffuser, when conditioning on the starting or ending state, we repeated the state to match the overlap size between trajectories. We slightly modify their setup in training by first noising the prediction output, and then using the same randomly sampled diffusion step parameter $t$ to seperately noise the condition. We use exponential trajectory blending with $\beta = 2.0$ on only the overlapping trajectory segments, like in the original paper.

\subsubsection{Algorithm Hyperparameters}
\label{appendix:hyperparameters}
Here we report the common hyperparameters we ran our experiments with. We tuned each algorithm differently for the navigation environment (Maze) manipulation environments (the other five environments described in~\ref{section:env}) as we found the Maze's more precise demands required different parameters to work well. 

For all environments, we down-sample the trajectory given by the diffusion model to a shorter horizon trajectory, sometimes called skip-planning~\cite{lu2025makes}. During inference we then linearly interpolate these down-sampled trajectories into a full-length one before they are given to the trajectory-following controller. We did this because we found that DiT on sequences longer than $128$ struggled with producing coherent trajectories and required unacceptable compute to train. For the navigation environments we use the same horizon across all algorithms; for the manipulation environments where fine-grained detail around picks was important we use less skip planning for the CNN-based models. We generally found the CNN models capable of handling planning horizons up to $1024$ without issue. 

We performed $[-1,1]$ min-max normalization to the observation and generation space for all algorithms, except the video inputs to visual Diffusion Policy, which we performed $[0,1]$ min-max normalization on. 

For Diffusion Planner and CompDiffuser, instead of jointly generating states and actions as in~\citet{janner2022planning}, we use a separate inverse dynamics model (as in~\citet{ajay2023is}). This removes the need to tune action versus state weighing in the diffusion objective, and generally performs better than joint prediction \cite{lu2025makes}. We execute this inverse dynamics model with feedback, executing the action $a_t$ at each time step $t$ by giving the model the current state in the plan $p_t$ and the current state from the environment $s_t$, such that $a_t = \text{invdyn}(p_t,s_t)$. 

Full details for our Diffusion Planners are given in Table~\ref{table:diffusion_planner_params}, Diffusion Policy in Table~\ref{table:diffusion_policy_params}, and CompDiffuser in Table~\ref{table:compdiffuser_params}. 

\begin{table}[]
\begin{tabular}{lll}
\hline
\rowcolor[HTML]{ECF4FF} & Common Parameters                 &                                   \\ \hline
Hyperparameter          & Navigation                        & Manipulation                      \\ \hline
Optimizer               & AdamW~\cite{loshchilov2017fixing} & AdamW~\cite{loshchilov2017fixing} \\
Base LR                 & 2e-4                              & 2e-4                              \\
Weight Decay            & 1e-5                              & 1e-5                              \\
LR Schedule             & Cosine                            & Cosine                            \\
Diffusion Prediction    & Noise                             & Clean                             \\
Sampling Temperature    & 0.5                               & 0.5                               \\
Batch Size              & 256                               & 64                                \\
Convolution Padding     & Repeat                            & Repeat                            \\
Solver                  & DDPM                              & DDPM                              \\
SDE Form                & Continuous                        & Continuous                        \\
Diffusion Steps Train   & 100                               & 100                               \\
Diffusion Steps Eval    & 500                               & 100                               \\ \hline
\rowcolor[HTML]{ECF4FF} & U-Net Parameters                  &                                   \\ \hline
Hyperparameter          & Navigation                        & Manipulation                      \\ \hline
Embedding Dimension     & 64                                & 64                                \\
Layer Sizes             & (1,2,2,2)                         & (1,2,2,2)                         \\
Planning Horizon        & 128                               & 512                               \\
Model Size              & 120 MB                            & 120 MB                            \\ \hline
\rowcolor[HTML]{ECF4FF} & DiT Parameters                    &                                   \\ \hline
Hyperparameter          & Navigation                        & Manipulation                      \\ \hline
Embedding Dimension     & 64                                & 64                                \\
Model Dimension         & 256                               & 256                               \\
N Heads                 & 64                                & 64                                \\
Depth                   & 8                                 & 8                                 \\
Planning Horizon        & 128                               & 128                               \\
Model Size              & 74.2 MB                           & 74.2 MB                           \\ \hline
\rowcolor[HTML]{ECF4FF} & Eq-Net Parameters                 &                                   \\ \hline
Hyperparameter          & Navigation                        & Manipulation                      \\ \hline
Kernel Expansion Rate   & 10                                & 5                                 \\
Channel Dimension       & 256                               & 128                               \\
Depth                   & 15                                & 25                                \\
Planning Horizon        & 128                               & 512                               \\
Model Size              & 61.1 MB                           & 48.9 MB                          
\end{tabular}

\caption{Diffusion Planner~\cite{janner2022planning} hyperparameters.}
\label{table:diffusion_planner_params}
\end{table}

\begin{table}[]
\begin{tabular}{llll}
\hline
\rowcolor[HTML]{ECF4FF} 
                & Diffusion Policy & Parameters &             \\ \hline
Hyperparameter  & Maze             & Well-Plate & Block-Stack \\ \hline
To              & 1                & 2          & 6           \\
Tp              & 16               & 16         & 16          \\
Ta              & 15               & 16         & 16          \\
Img Res         & N/A              & (160,160)  & (160,160)   \\
Video Encoder   & N/A              & ResNet18   & ResNet18    \\
Gradient Steps  & 1000000          & 1000000    & 1000000     \\
Pos/Vel Control & Positional       & Positional & Positional  \\
Architecture    & Chi U-Net        & Chi U-Net  & Chi U-Net   \\
Learning Rate & 2e-4 & 1e-4 & 1e-4 \\
Diffusion Prediction & Noise & Noise & Noise \\
Diffusion Steps Train   & 100       & 100                & 100                               \\
Diffusion Steps Eval    & 100          & 100            & 100                               \\ 
Batch Size & 256 & 256 & 256
\end{tabular}
\caption{Diffusion Policy~\cite{chi2023diffusion} hyperparameters. To is the observation horizon (how many past proprioceptive observations we give to the model as conditioning). Tp is the prediction horizon (how many future actions we predict). Ta is the action execution horizon (how many actions we execute from the action chunk). For the observation conditions, we used a separate encoder for the manipulation tasks and inpainting for the navigation task. Other parameters are the same as in Table~\ref{table:diffusion_planner_params}.}
\label{table:diffusion_policy_params}
\end{table}

\begin{table}[]
\begin{tabular}{lll}
\hline
\rowcolor[HTML]{ECF4FF} 
                          & CompDiffuser Parameters &              \\ \hline
Hyperparameter            & Navigation              & Manipulation \\ \hline
High-Level Traj. Horizion & 128                     & 512          \\
Single Model Horizon      & 32                      & 128          \\
\# of Trajectories         & 5                       & 5            \\
Gradient Steps            & 1250000                 & 1250000      \\
Diffusion Prediction      & Clean                   & Clean        \\
Architecture              & U-Net                   & U-Net        \\
Batch Size                & 250                     & 250          \\
Diffusion Steps Train   & 100                               & 100                               \\
Diffusion Steps Eval    & 100                               & 100                               \\ 
Inference Mode            & Parallel                & Parallel    
\end{tabular}
\caption{CompDiffuser~\cite{luo2025generative} hyperparameters. Generally, we tried to use parameters similar to the original paper's when possible. Other parameters are the same as in Table~\ref{table:diffusion_planner_params}.}
\label{table:compdiffuser_params}
\end{table}

\subsubsection{Additional Architecture Experiment Details}
\label{appendix:arch_exp_details}

\begin{figure}
\centering
  \includegraphics[width=.90\linewidth]{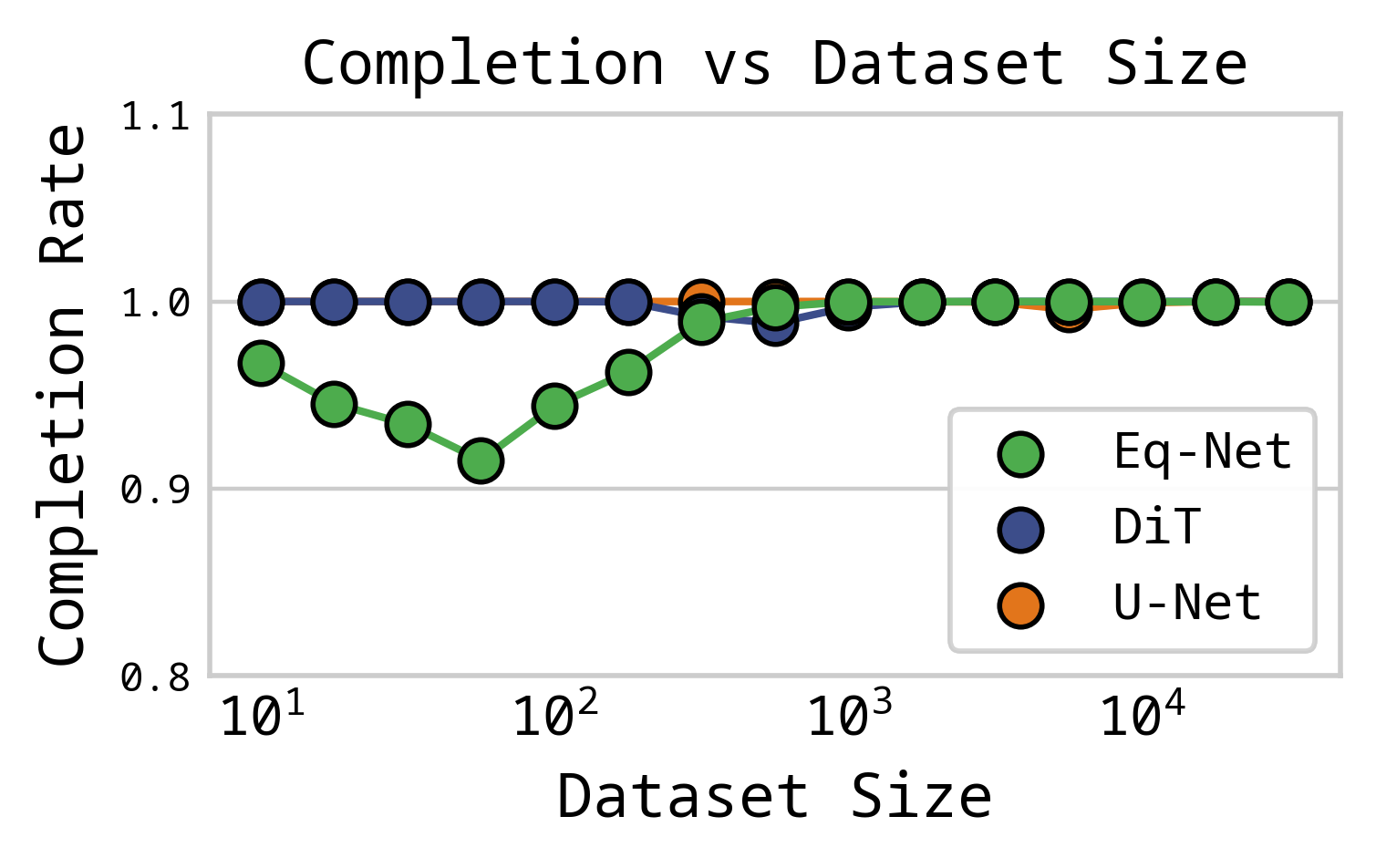}
\caption{Completion rates for each architecture in Maze, as a function of training data scale. Completion rate is a percentage out of 1.}
\label{fig:appendix_complete}
\end{figure}

Here we show the completion rate of each architecture in our Maze environment. We found that completion dips slightly for Eq-Net near the boundry where it achieves compositionality, and a smaller dip for DiT where it achieves compositionality. See Figure~\ref{fig:appendix_complete}. We consistently found that models experienced a dip in completion around the threshold they began to generalize (for Eq-Net around $\sim 10^{1.75}$ training samples, for DiT around $\sim 10^{2.75}$ training samples and for U-Net around $\sim 10^{3.75}$ training samples. Similar behaviour was seen in~\citet{pham2025memorization} where diffusion models undergo a three-phase transition as training data scales, starting with memorization, then undergoing a spurious generalization phase where sample quality drops, and finally achieving true generalization where quality is restored along with the achievement of creativity. 

\subsubsection{Experiment Computing Resources}

Experiments used roughly 168 GPU-days of compute cluster use on GPUs equivalent to NVIDIA L40s.

\subsection{More Environment Details}
\label{appendix:env_details}

Here we describe our environments in more detail. 

\textbf{Lights}: We adapt this environment from the Lights Out/Puzzle environment from OgBench~\cite{park2024ogbench} with modifications. Specifically, in our environment pressing a button only changes the state of the button itself and not the surrounding buttons, and all trajectories start from the same state where no lights are pressed. We did both of these to make it easier to isolate when a generated trajectory has combined subskills from training trajectories. 

\subsubsection{Dataset Details}

Table~\ref{appendix:env_detail_table} describes details of the environment parameters and dataset description.

\begin{table*}[]
\centering
\begin{tabular}{lllllllll}
\hline
\rowcolor[HTML]{ECF4FF} 
                &           &            & Environment        & Descriptions      &           &            & \cellcolor[HTML]{ECF4FF}    &                         \\ \hline
Environment     & Type      & Embodiment & \# of Trajectories & Trajectory Len & \# of Obs & \# of Acts & \# of Possible Traj & \# of Eval Traj \\ \hline
Maze            & Numpy     & Point      & 316                & 1000              & 2         & 2          & 705,421                     & 10000                   \\
Didactic        & Real      & Franka     & 3                  & $\sim$12000       & 4         & 4          & 1                           & 100                     \\
Cube-Stack      & RoboSuite & Franka     & 100                & $\sim$1000        & 4         & 4          & 720                         & 500                     \\
Lights          & OGBench   & UR5e       & 1000               & 500               & 4         & 4          & 65,536                      & 2500                    \\
Well-Plate Sim  & RoboSuite & Franka     & 50                 & $\sim$1000        & 4         & 4          & 7056                        & 500                     \\
Well-Plate Real & Real      & Franka     & 50                 & $\sim$51000       & 4         & 4          & 7056                        & 100                    
\end{tabular}
\caption{More details about our environments.}
\label{appendix:env_detail_table}
\end{table*}

\subsubsection{Data Collection}
\label{appendix:data_collection}

Here we elaborate on our data collection process for each environment. 

\textbf{Maze}: In Maze, we collect data by starting with the point agent in the top left corner. At each intersection, it randomly selects whether to travel down or to the right at that intersection, and repeats for future intersections. If the point agent is along the bottom or right-hand face, it will only go to the right or down respectively. 

\textbf{Well-Plate}: In both the simulated and real Well-Plate environments, we enumerated all possibilities of start-plate cells to goal-plate cells (81, as there are 9 cells in each plate). Out of these 81 start-goal pairs, we randomly take 76 of them, and then randomly sample from the remaining pairs in groups of 3, excluding cases where there are collisions in either the start or goal plate. We did this until we had 50 trajectory descriptions consisting of three pick and places. To generate trajectories for these descriptions we generate a set of keypoints for the arm to travel between, and them follow them with a positional impedance controller. We fixed the RNG when generating the trajectory descriptions between the simulated and real well-plate environments, but the actual trajectories vary because of differences between the well plate placements between environments. 

\textbf{Block-Stack}: In Block-Stack, to generate a trajectory description we index the coloured blocks from 0 to 5, generate all ordered permutations of those indexes, and then randomly select 100 from the list. To execute description we use a similar system to Well-Plate where we generate keypoints guiding the arm to pick up each block and place it at the stack location, advancing the height the place occurs at in accordance with the growing height of the stack. 

\textbf{Lights}: We use the same trajectories in the puzzle-4x6-play-v0 dataset from ~\citet{park2024ogbench}. This environment has $2^{16}$ possible trajectories, as each button can either be turned on or off. 

\textbf{Didactic}: For Didactic, we manually wrote the set of keypoints for the three training trajectories. 

\subsubsection{Environment Diversity Primitive Descriptions}
\label{appendix:env_diversity_details}

Here we describe the environment observation primitives we use to describe diversity, as defined in Section~\ref{section:define_novelty}. We also describe how many possible unique trajectories exist under our primitive formulation. We do not define primitives for Didactic, because the task is designed to only have $1$ possible trajectory that completes the task. 

\textbf{Maze}: In maze, the diversity primitives are the set of visited intersections in the maze. Intersections are the crossroads between the corners of four obstacles (or two obstacles and the wall). We calculate these visitations by taking the executed trajectory and seeing if the difference between each state in the trajectory and each intersection is smaller than some threshold. This means that each trajectory is defined by the set of intersections it passes through, and because each training trajectory moves from the top-left to bottom-right, each trajectory visits the same number of intersections. See Section~\ref{appendix:grid_possible_traj} for the number of possible trajectories.

\textbf{Well-Plate}: In well-plate, the diversity primitives are the set of \emph{occupied cells} in the goal plate, and the \emph{empty cells} in the starting plate. Each training trajectory moves three blocks from the starting plate to the goal place, so each training trajectory is defined by a set of $6$ primitives. This makes the diversity measure blind to the order the blocks are moved in, but the end result is still a large space of potential trajectories. The total number of possible complete trajectories is equal to $\binom{9}{3}^2 = 7056$, as you choose 3 blocks from the starting plate to move to 3 cells in the goal plate.

\textbf{Block-Stack}: In block-stack, the diversity primitives are tuples of a cube's colour and it's position along the stack. For instance, the set {(Red,1),(Blue,2),(Yellow,3)} is a possible trajectory description. The number of possible complete trajectories is $6!$, as you pick $1$ block from $6$, then $1$ from the remaining $5$, et cetera.

\textbf{Lights}: In Lights, the diversity primitives are binary variables associated with each light being turned on (via being pushed). The number of possible trajectories is $2^{16}$, as each light can either be turned on or off. 

\subsubsection{Goal-Conditioned Environment Details}
\label{appendix:grid_gc_env_details}
For the Goal-conditioned Maze experiment, we generated a new dataset in the $5\times5$ Maze consisting of random walks from points along the edge of the maze to other edge points with $\sim 200$ trajectories. We then found the set of all edge point start-goal pairs unseen in the same training trajectory to use as evaluation pairs. This ensure that to complete the unseen goals, the diffusion model must stitch sub-trajectories from the training set without simply replicating existing trajectories.

\subsubsection{Grid Environment Possible Trajectories}
\label{appendix:grid_possible_traj}
\textbf{Lemma 1}: For a grid of size $N$ where $N$ is the number of blocks along the grid's height and width, the number of possible topologically distinct trajectories that are optimal in length is upper-bounded (inclusive) by $\frac{N!}{(N/2)!(N/2)!}$ .

\textbf{Proof}: First, observe that the number of optimal topologically distinct trajectories for any start-goal pair that is \emph{not} from a corner to the opposite corner will be upper bounded by those that are. This is because they will have a smaller number of segments to traverse, meaning all of the analysis below applies with a smaller asymptotic quantity. This means finding the maximum number of distinct optimal trajectories in the corner to opposite corner case will upper bound the same quantity for any other start-goal pair.

Assume we are navigating from the top left to the bottom right corner for notational purposes. Observe that for any optimal path, the agent will move right $N/2$ times and down $N/2$ times. This means all possible trajectories can be represented by a binary string of length $\frac{N}{2} + \frac{N}{2} = N$ where there are an equal number of $1$s and $0$s. The size of this set is $(\frac{N}{N/2}) = \frac{N!}{(N/2)!(N/2)!}$, because from $N$ positions we choose $N/2$ to be filled with either $0$s or $1$s.

\subsection{Further Architecture Details and Experiments}

\subsubsection{U-Net is Not Shift Equivariant}
\label{appendix:unet_analysis}

Strictly speaking, empirical evidence that U-Net is not shift equivariant or that Eq-Net is should not be required, as it directly follows from known properties of their architecture: the U-Net has down-sampling strides which break shift equivariance~\cite{zhang2019making}, and Eq-Net has only 1-length strides with no pooling layers which preserves convolutions' shift equivariance. 

For the sake of visual demonstration, we show this by taking our trained U-Net and Eq-Net models in the unconditional environment and giving the model an input sequence of all zeros, besides a patch of Gaussian noise in the middle of the trajectory. We then shift this Gaussian patch by $40$ pixels and analyze the output of each model. Formally, we take two initially empty trajectories $\tau_1,\tau_2  = [0,...0]^T$ and add a patch of Gaussian noise $50$ pixels wide to the middle of the first trajectory: $n\sim \mathcal{N}(0,10)$,$\tau_1[225:275] = n$ and add the same Gaussian noise shifted by $40$ pixels from the middle to the second trajectory: $\tau_2[265:315] = n$. Finally, we run each through a given diffusion model at $t=0$: $x_1 = f(\tau_1,0),x_2 = f(\tau_2,0)$. A model with shift equivariance should find that the pixels of the output de-noised trajectory corresponding to the same patch as the Gaussian noise should be equal, that is, $x_1[225:275] = x_2[265:315]$.

\begin{figure}
  \begin{subfigure}{0.25\textwidth}
    \centering
    \includegraphics[width=.9\linewidth]{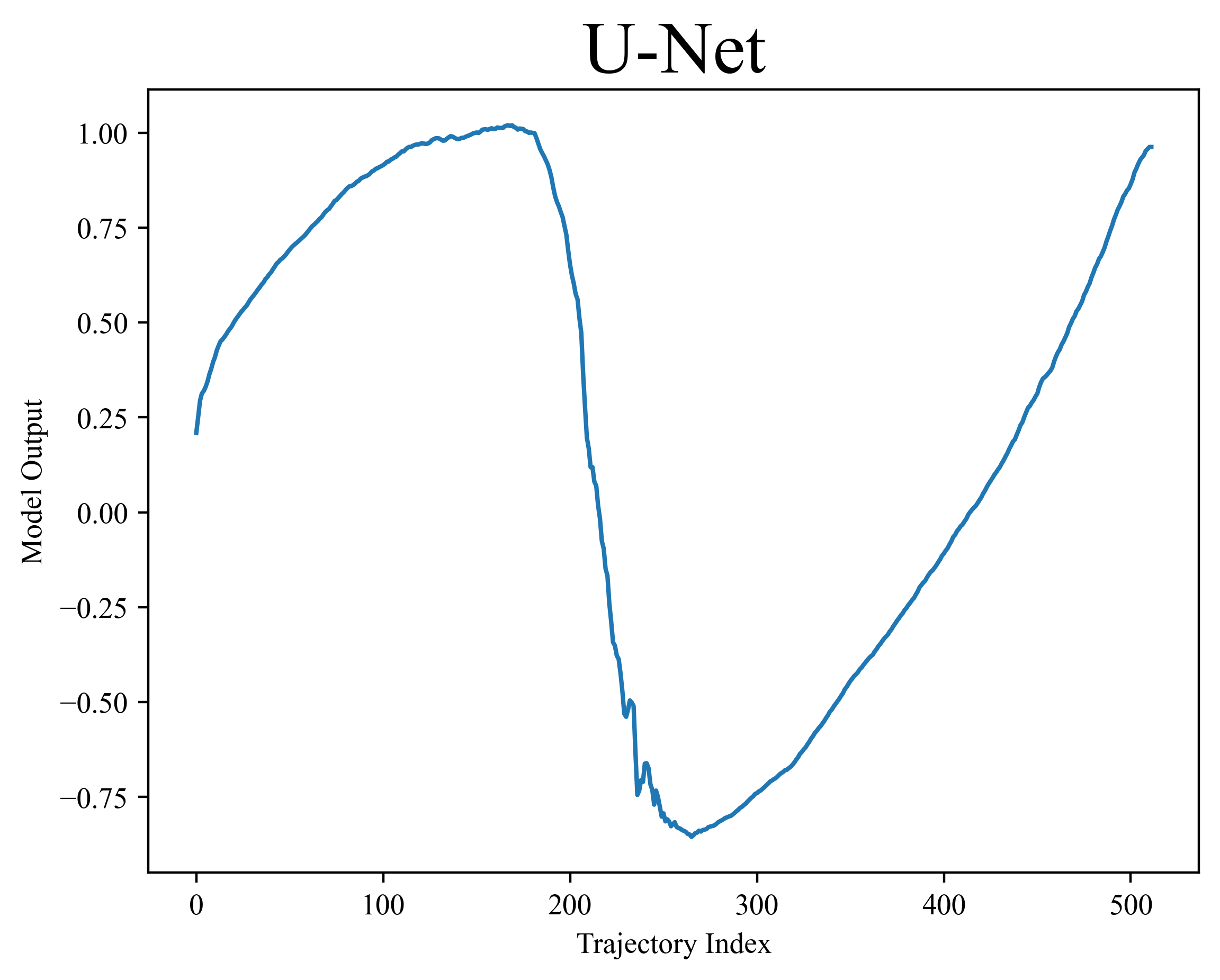}
  \end{subfigure}%
  \begin{subfigure}{0.25\textwidth}
    \centering
    \includegraphics[width=.9\linewidth]{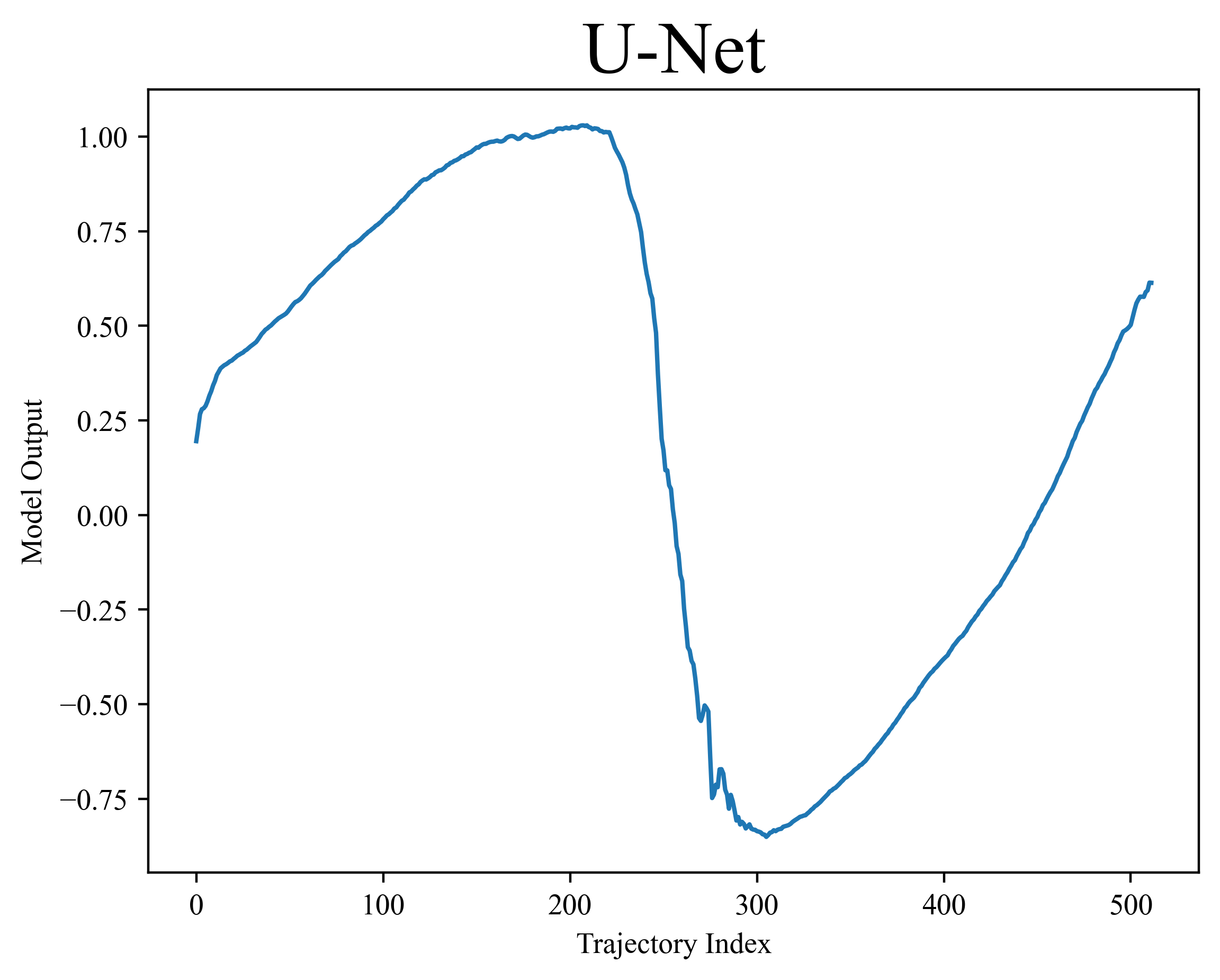}
  \end{subfigure}
  \begin{subfigure}{0.25\textwidth}\quad
    \centering
    \includegraphics[width=.9\linewidth]{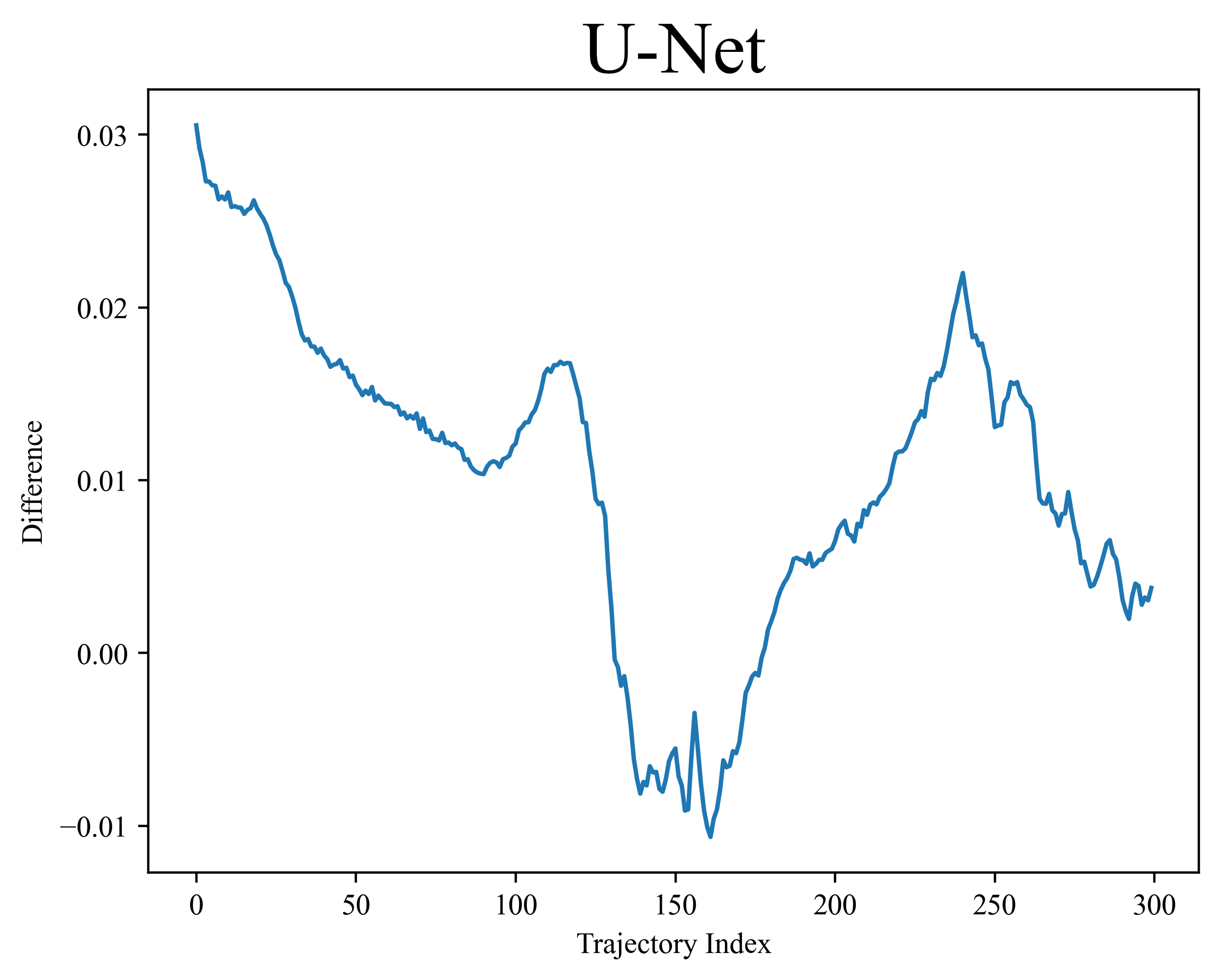}
  \end{subfigure}
  \medskip

  \begin{subfigure}{0.25\textwidth}
    \centering
    \includegraphics[width=.9\linewidth]{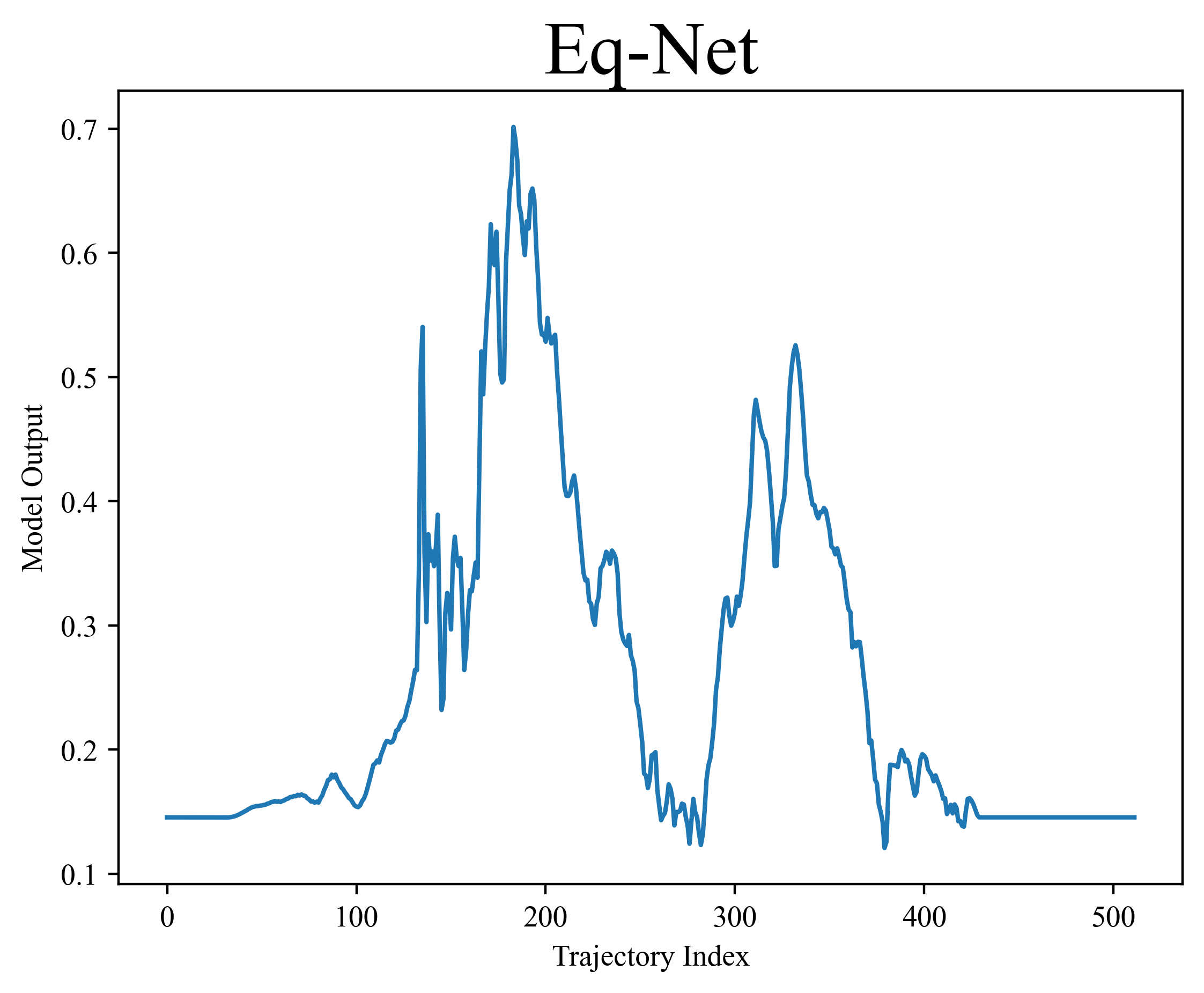}
  \end{subfigure}
  \begin{subfigure}{0.25\textwidth}
    \centering
    \includegraphics[width=.9\linewidth]{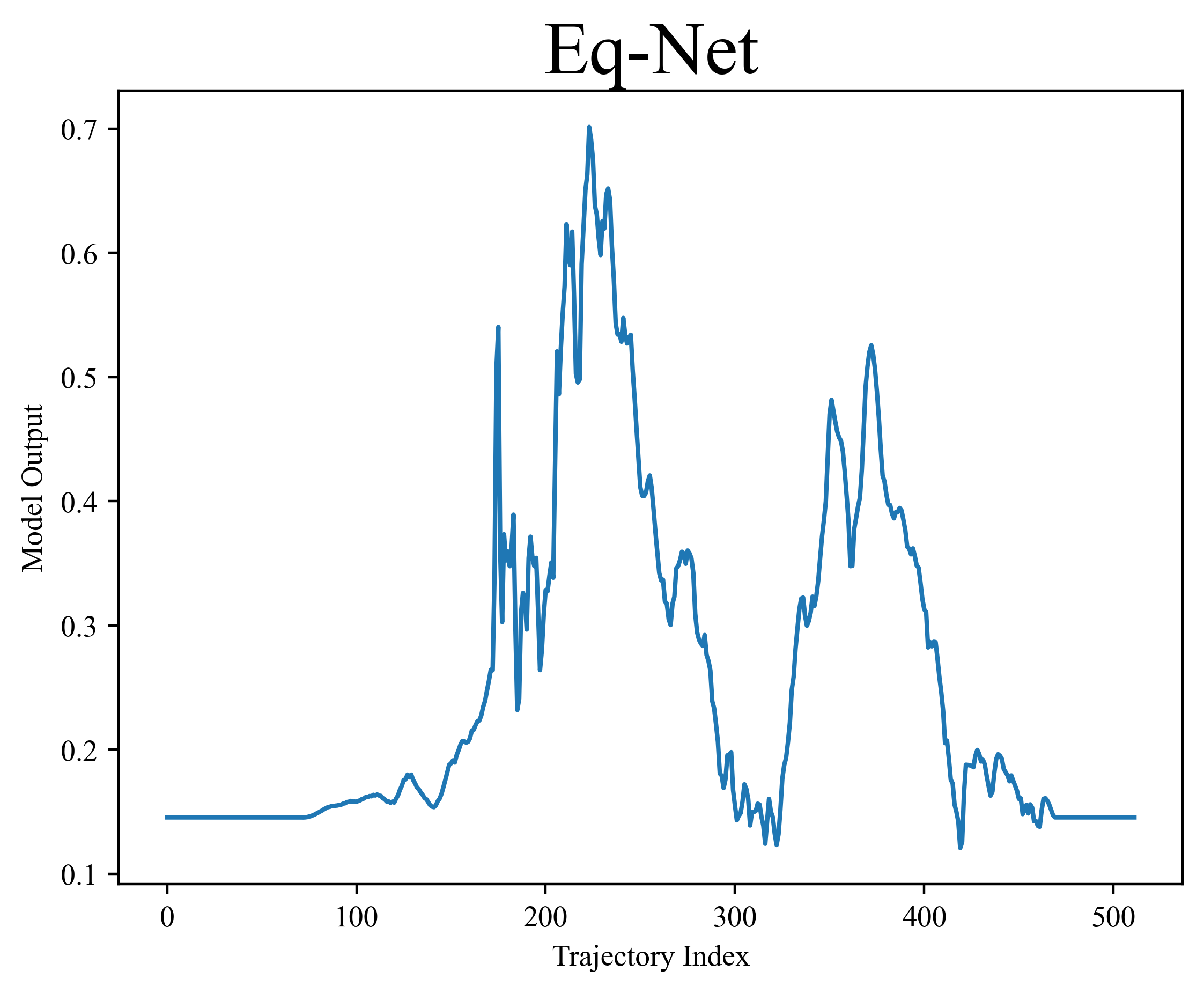}
  \end{subfigure}
  \begin{subfigure}{0.25\textwidth}
    \centering
    \includegraphics[width=.9\linewidth]{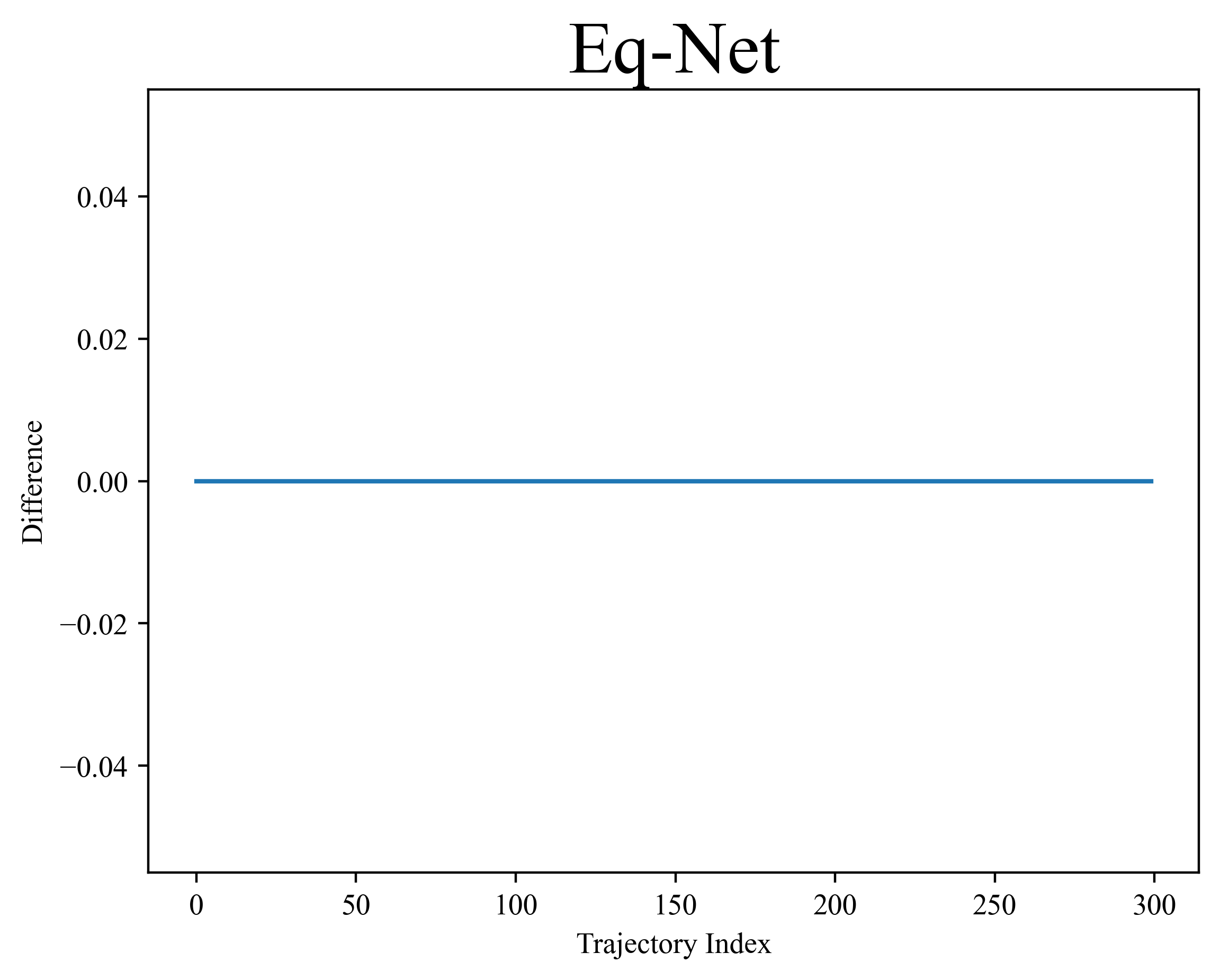}
  \end{subfigure}
  \caption{A small experiment showing that the U-Net architecture is not shift equivariant, while our Eq-Net is. The U-Net architecture has substantial differences in the output when shifted, shown as a different in the rightmost plot, while Eq-Net retains the same output but translated.}
  \label{fig:pos_eq}
\end{figure}

Figure \ref{fig:pos_eq} shows the output of each model at the original input position and the input position shifted to the right, as well as the difference between the outputs when the shift is corrected for. The U-Net architecture has substantial differences in the output when shifted, shown as a difference in the rightmost plot, while Eq-Net retains the same output but translated.

\subsubsection{Engineering Details for Eq-Net}
\label{appendix:eqnet_details}
\begin{figure}[b]
  \begin{center}
    \includegraphics[width=0.48\textwidth]{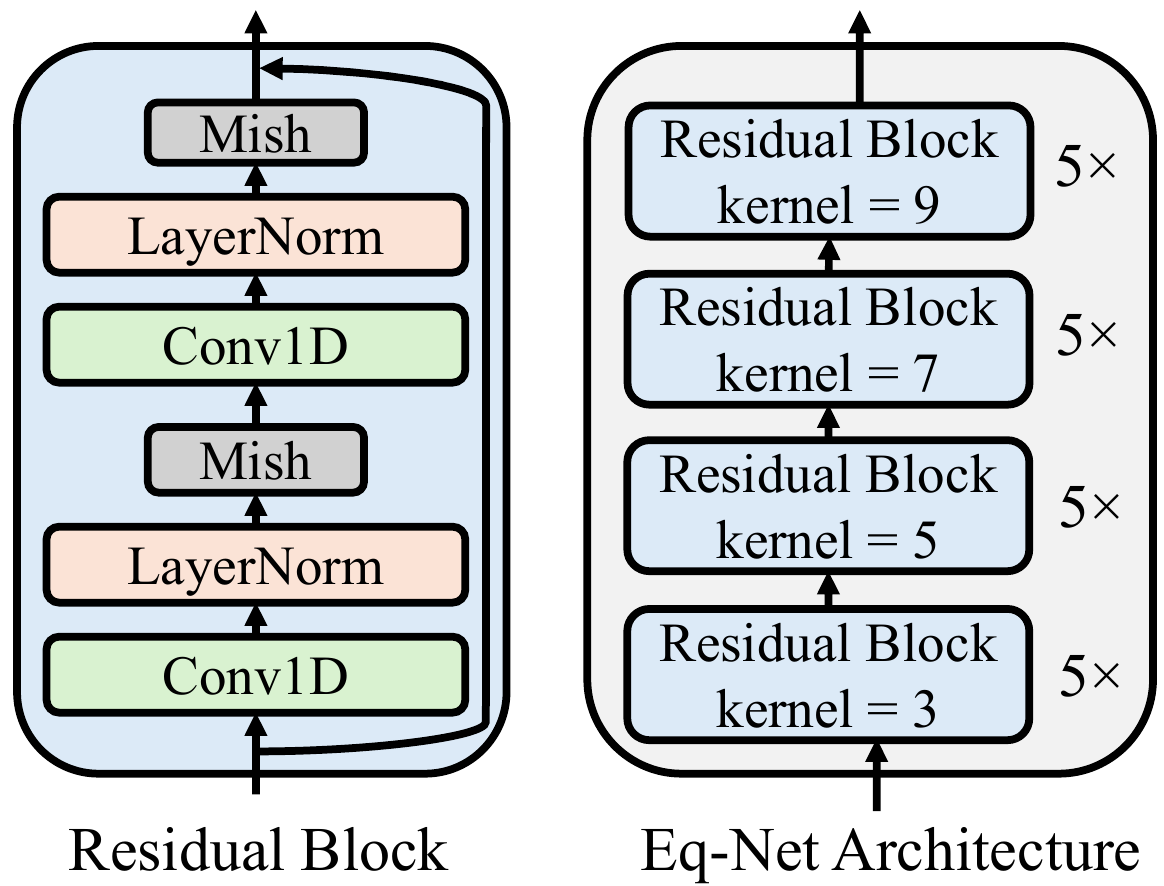}
  \end{center}
  \caption{Our fully local, shift equivariant architecture. By stacking stride 1 convolutions with a small kernel length and removing pooling and down-striding, we ensure the CNN cannot attend to position within the trajectory or far-away states during denoising.}
  \label{fig:arch_diagram}
\end{figure}

For an architecture diagram of Eq-Net, see Figure~\ref{fig:arch_diagram}.

As noted in ~\cite{kamb2024analytic}, in theory, circular padding is required for a CNN to have full shift equivariance, but in practice this is uncommon.

Future work adapting these findings into a more practical architecture might investigate using adaptive polyphase sampling \cite{chaman2021truly}, anti-aliasing layers \cite{zhang2019making}, or other modern tricks for incorporating shift equivariance into CNNs.  Additionally, the depth of the network, way the layers of differing kernel sizes are layered, and channel size can likely be greatly optimized to increase computational efficiency and model capacity.

\subsubsection{Kernel Size Tuning}
\label{appendix:kernel_tuning}

The main parameter that requires tuning in Eq-Net is the \emph{kernel expansion rate} (KER) which determines how fast the kernels grow deeper into the network. In Eq-Net, the kernels in each layer starts as small as possible (3) and grows 2 larger every KER layers, meaning for a fixed depth of $25$, the maximum kernel size the network has is determined by $\sim 25/\text{KER}$. Thus, a small KER corresponds to a network with more larger kernels, and a larger KER with many small kernels.

Broadly, we found that tuning KER trades off between \emph{overall plan consistency} (how likely the trajectories were to be locally consistent and stay outside the obstacles) and \emph{compositionality}. To show this, we trained Eq-Net models in our unconditional setting with a variety of KER values. Our results are shown in Table \ref{appendix:ker_results}. We found that the intermediate value of 10 generated both a strong diversity of plans as well as reliably well-composed plans, while making the model too non-local made the model too local and reduced completion rates.

\begin{table}[]
\begin{tabular}{llllll}
\hline
\rowcolor[HTML]{ECF4FF} 
Kernel Expansion Rate  & 6      & 8     & 10    & 12    & 16     \\ \hline
Task Completion \%      & 0.991  & 0.997 & 0.995 & 0.984 & 0.965  \\
\% of Generations Novel & 0.5056 & 0.752 & 0.802 & 0.869 & 0.9701
\end{tabular}
\caption{Results of changing the kernel expansion rate in the unconditional environment, taken over 1000 generations. Percentages are all out of 1.}
\label{appendix:ker_results}
\end{table}

Our experiments with the default U-Net architecture and our Eq-Net architectures in total confirm our earlier hypothesis that while CNNs are meant to have strong inductive biases to encourage composition, the particular set of decisions made by \citet{janner2022planning} means at small data amounts they can still overfit heavily and attend to non-local information.

\subsubsection{Kernel Diversity}
\begin{figure}

  \begin{center}
    \includegraphics[width=0.48\textwidth]{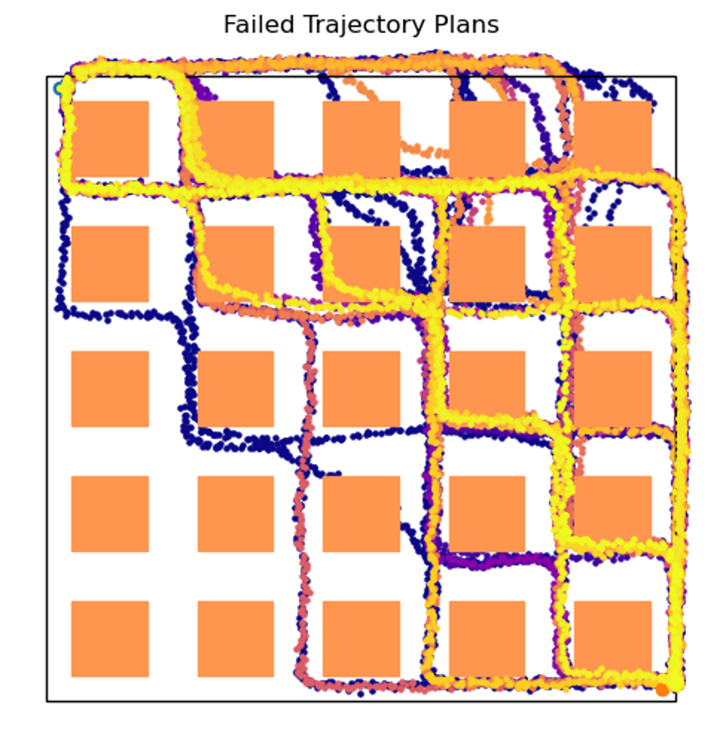}
  \end{center}
  \caption{Generated plans when only using medium-sized kernels. Trajectories end up "ungrounded" to the grid, resulting in failed executions. Each colour represents a different generated trajectory.}
  \label{fig:small_kernel_appendix}
\end{figure}

One thing that we found was important when designing Eq-Net was to include both very small (such as k=3) along with the larger kernel layers. When we used only medium sized kernels, it would lead to trajectories that were oddly "ungrounded" to the navigation plane, and would look correct in shape but would be shifted such that the resulting trajectory would nearly always hit obstacles. Figure \ref{fig:small_kernel_appendix} shows the visual results of this in early experiments in a $5 \times 5$ maze. We found that increasing the kernel size from small to large during training helped, somewhat mirroring the process in normal CNNs where fixed-size kernels stride over the whole image initially (capturing fine detail) and then smaller subsets after each downstride (capturing coarse detail). However, we did not explore whether the exact ordering of the kernel layers in Eq-Net matters or not. 

\subsubsection{Positional Encoding}
\label{appendix:posenc}

\begin{figure}

  \begin{center}
    \includegraphics[width=0.48\textwidth]{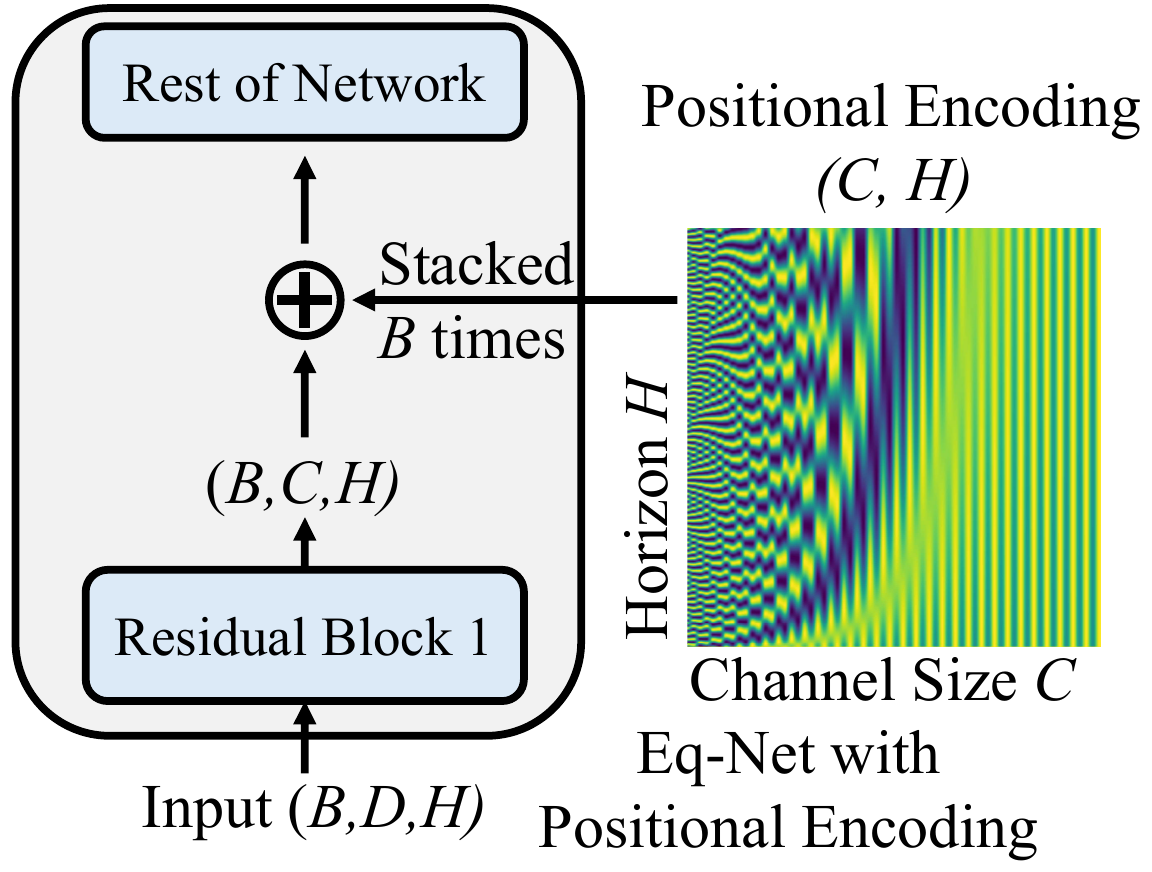}
  \end{center}
  \caption{Our positional encoding approach. Sinusoidal embeddings which encode position channel-wise are generated, with a unique embedding for each position in the sequnce. These embeddings are then added after the first convolutional layer.}
  \label{fig:appendix_pos_enc}
\end{figure}

When testing how our small receptive field model performs when it has access to positional information for each state, we use the same positional encodings as in~\citet{vaswani2017attention}, with slightly varied parameters. Figure~\ref{fig:appendix_pos_enc} shows a visualization of how we add positional encoding to the Eq-Net architecture. These encodings turn the position of each state into waves forming a geometric progression from which the position of a particular state can be inferred. The encodings are added after the first residual block. Specifically, the positional encodings are defined as

$\text{PE}_(\text{pos},2i) = sin(\text{pos}/{1000}^{2i/H})$

$\text{PE}_(\text{pos},2i+1) = cos(\text{pos}/{1000}^{2i/H})$

We used a maximum length of $1000$ instead of $10000$ used by~\citet{vaswani2017attention} because of the shorter length of our sequences.

It is worth noting that this strength of positional encoding (where the network is given enough information to precisely attend to the location of a state in the sequence) is likely quite a bit stronger than the positional attention from CNN aliasing. We included this as a representation of the worst-case scenario, to show the upper bound of models that are designed to attend to position, such as diffusion transformer architectures~\cite{peebles2023scalable}. 

\subsection{More Experiments}
\label{appendix:more_experiments}
\subsubsection{Do Common Diffusion Model Parameters Impact Compositionality?}
\label{appendix:diffusion_params}
In the following sections we report the findings of exploratory experiments we performed on a wide host of other diffusion model parameters we examined. Ultimately, we found that none of the following factors have significant effects on composition, with the exception of overfitting which has a moderate effect. All these experiments were done in our Maze environment, with reported statistics taken over $1000$ evaluation samples.

\subsubsection{Does Using the Continuous or Discrete SDE Formulation Impact Compositionality?}
No. In diffusion models, the discrete and continuous formulations refer to different ways to parameterize the underlying stochastic differential equation governing the diffusion process~\cite{24diffusiontutorial,chan2024tutorial}. Table~\ref{table:sde_appendix} shows that the continuous and discrete SDEs perform functionally identically.

\begin{table}[]
\begin{tabular}{lllll}
\hline
\rowcolor[HTML]{ECF4FF} 
                       & EqNet C & EqNet D & UNet C & UNet D \\ \hline
Completion \%           & 0.989   & 0.933   & 1      & 1      \\
\% of Generations Novel & 0.815   & 0.7894  & 0      & 0     
\end{tabular}
\caption{\textbf{Discrete vs Continuous SDEs Don't Matter for Compositionality:} Our findings show that EqNet and UNet trained with the continuous SDE (C) and discrete SDE (D) perform similarly.}
\label{table:sde_appendix}
\end{table}

\subsubsection{Does High-Temperature Sampling Impact Compositionality}

Surprisingly, not substantially. We expected the diffusion sampling temperature to have some effect, but instead found that models behaved similarly regardless of temperature, with some enhanced compositionality only occurring at temperatures high enough to greatly degrade sample quality. Table~\ref{table:high_temp_sampling} shows that high-temperature sampling does not make any model more creative.

\begin{table}[]
\begin{tabular}{llllll}
\hline
Temperature                      & 0.1                        & 0.5    & 1      & 2      & 4     \\ \hline
\rowcolor[HTML]{ECF4FF} 
\multicolumn{2}{l}{\cellcolor[HTML]{ECF4FF}Task   Completion} &        &        &        &       \\ \hline
U-Net                            & 1                          & 1      & 1      & 1      & 0     \\
DiT                              & 0.991                      & 0.991  & 0.995  & 0.952  & 0     \\
Eq-Net                           & 0.989                      & 0.987  & 0.989  & 0.993  & 0.26  \\ \hline
\rowcolor[HTML]{ECF4FF} 
\multicolumn{3}{l}{\cellcolor[HTML]{ECF4FF}\% of   Generations Novel}   &        &        &       \\ \hline
U-Net                            & 0                          & 0      & 0      & 0      & NaN   \\
DiT                              & 0.0226                     & 0.0221 & 0.0189 & 0.0297 & NaN   \\
Eq-Net                           & 0.827                      & 0.801  & 0.806  & 0.815  & 0.796
\end{tabular}
\caption{\textbf{High-Temperature Sampling Does Not Impact Compositionality}: Results in our Maze environments across a range of sampling temperatures show similar compositionality, with the only effect being on reducing sample quality at very high temperature values. NaN occurs because of a divide by zero over the number of generated trajectories. All percentages are out of 1.}
\label{table:high_temp_sampling}
\end{table}
\subsubsection{Does Predicting Noise vs Clean Impact Compositionality?}

No. In early experiments we found that predicting noise versus clean did not impact compositionality; however, it did impact sample quality (measured via completion rates). We found for the high-precision maze task that predicting noise worked better; when trained with the same parameters predicting a clean sample instead of noise, both U-Net and Eq-Net never successfully completed the maze. That our findings include environments where we train to predict noise (Maze) and where we train to predict clean (the manipulation ones) shows that the prediction target is likely irrelevant to compositionality.

\subsubsection{Does the Number of Diffusion Sampling Steps Impact Compositionality?}
Not past a certain, fairly low threshold. We found that while more sampling steps was very important to generate high-quality samples (measured by completion rate in our Maze environment, where slight perturbations from the maze causes the point agent to run into obstacles and die), we did not see a strong effect on compositionality, especially after roughly $50$ samples. Table~\ref{table:sampling_steps} shows the full results.

\begin{table*}[]
\centering
\begin{tabular}{llllll}
\hline
\# of Diffusion   Sampling Steps            & 10               & 50      & 100    & 500   & 1000  \\ \hline
\rowcolor[HTML]{ECF4FF} 
\multicolumn{2}{l}{\cellcolor[HTML]{ECF4FF}Task   Completion} &         &        &       &       \\ \hline
U-Net                                      & 0.839            & 1       & 1      & 1     & 1     \\
DiT                                        & 0.409            & 0.914   & 0.973  & 0.994 & 0.995 \\
Eq-Net                                     & 0.105            & 0.807   & 0.927  & 0.989 & 0.992 \\ \hline
\rowcolor[HTML]{ECF4FF} 
\multicolumn{3}{l}{\cellcolor[HTML]{ECF4FF}\% of   Generations Novel}    &        &       &       \\ \hline
U-Net                                      & 0                & 0       & 0      & 0     & 0     \\
DiT                                        & 0.009            & 0.00662 & 0.0258 & 0.009 & 0.032 \\
Eq-Net                                     & 0.688            & 0.796   & 0.813  & 0.804 & 0.81 
\end{tabular}
\caption{\textbf{The \# of Diffusion Steps Does Not Impact Compositionality}: After around $50$ sampling steps, while the quality of trajectories continues to climb, compositionality is not effected. Percentages are all out of 1.}
\label{table:sampling_steps}
\end{table*}

\subsubsection{Does Overfitting Impact Compositionality? }
\label{appendix:overfitting}

\begin{figure}
\begin{subfigure}[t]{.5\linewidth}
    \centering
    \includegraphics[width=\linewidth]{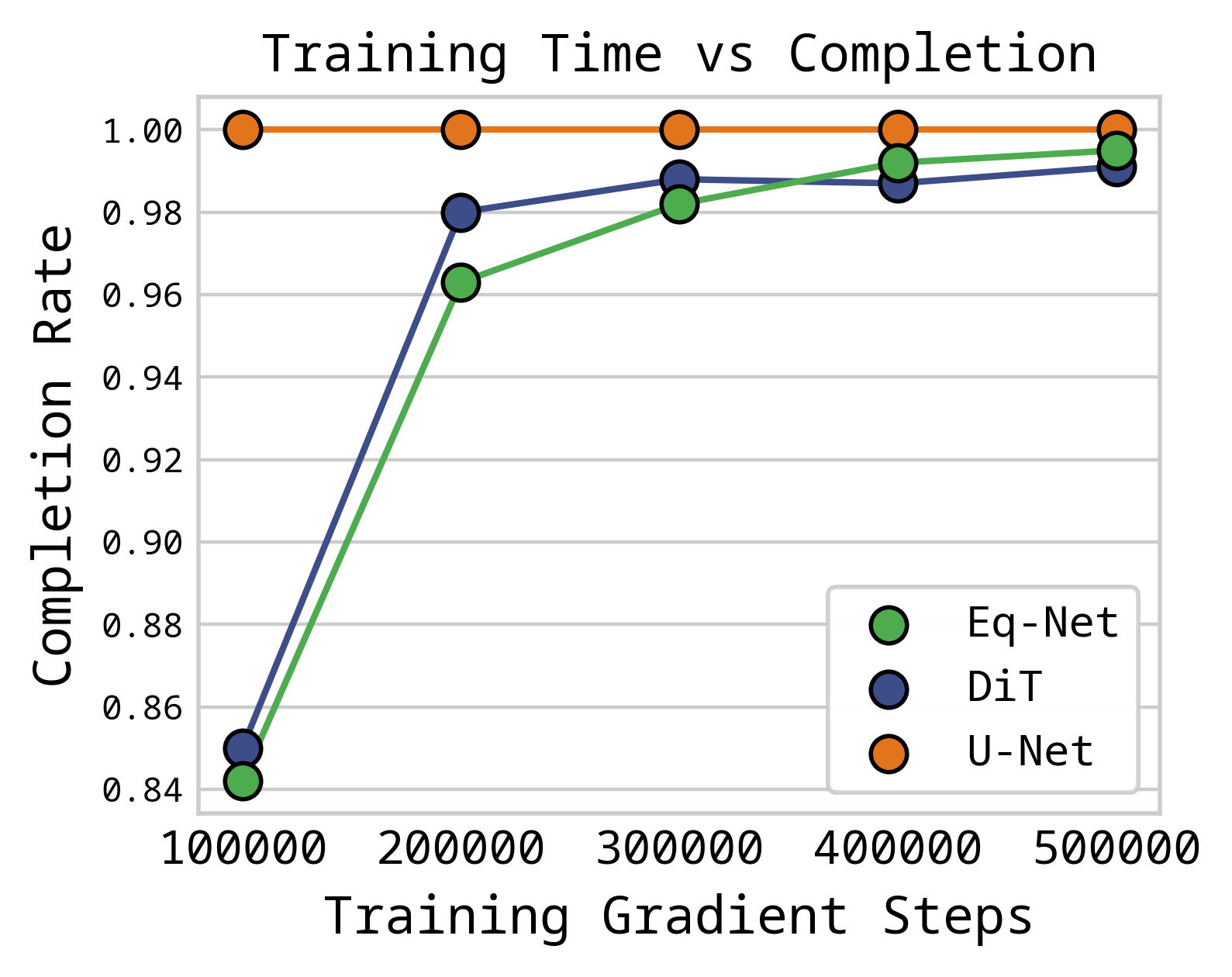}
    \end{subfigure}%
\begin{subfigure}[t]{.5\linewidth}
    \centering
    \includegraphics[width=\linewidth]{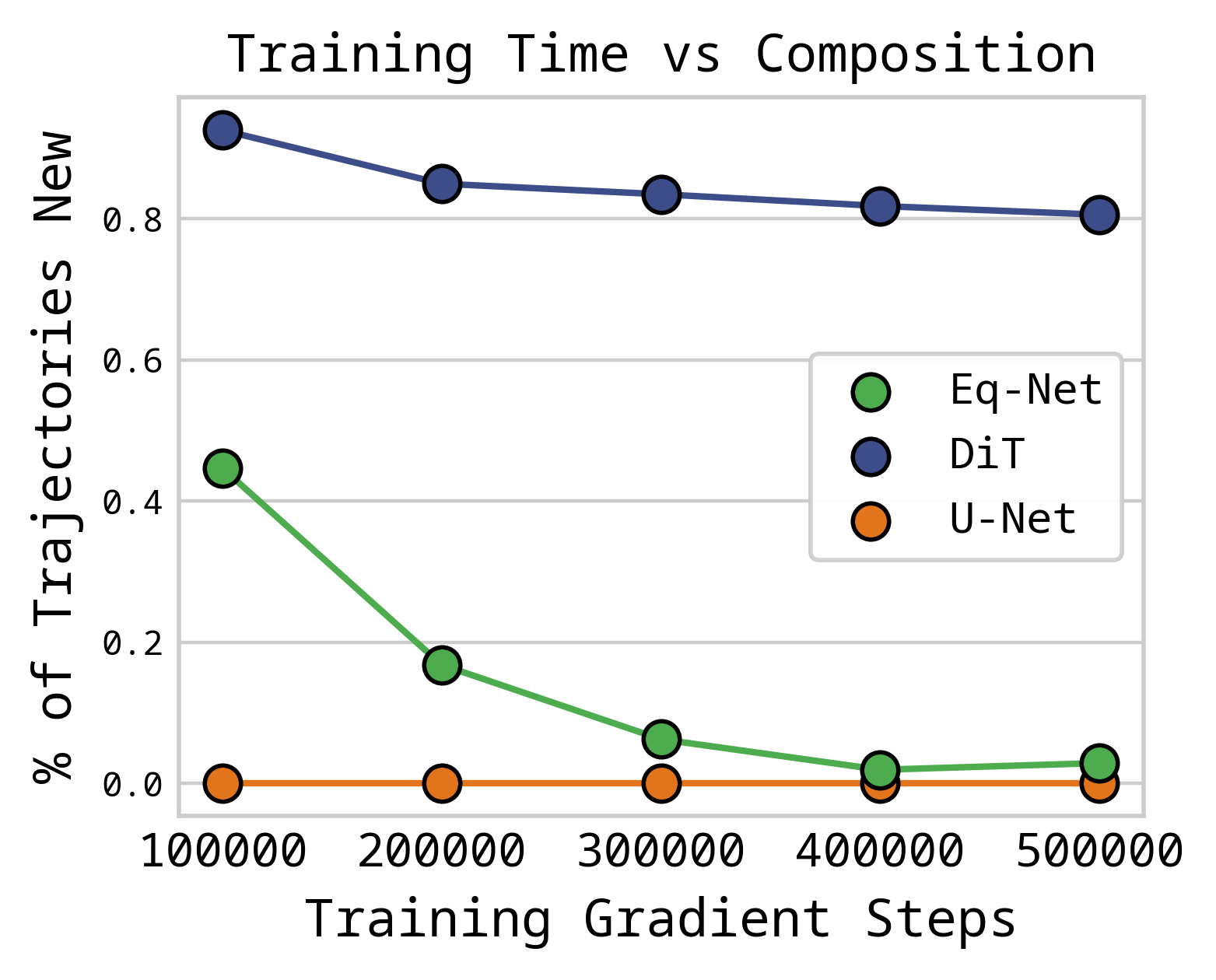}
    \end{subfigure}%
  \caption{\textbf{Overfitting Affects Model Compositionality}:Our results show that models lose some compositional ability when over-trained, but that the more compositional models retain compositionality in the training step limit. Percentages are out of 1. }
  \label{fig:overfitting}
\end{figure}

Yes. A common strategy in all machine learning to increase generalization is to perform early stopping, where training is ended before loss on the training set plateaus. Figure \ref{fig:overfitting} shows that overfitting indeed reduces compositionality in the small-data setting, as models early in training achieve both high completion rates (indicating the generated plans are kinematically feasible) as well as more novel trajectories than later in training. However, the models that show stronger compositionality (such as Eq-Net) retain this property late into training (i.e, the losses in compositionality from overfitting plateaus). This hopefully indicates that the proper inductive biases from architecture choices can avoid the overfitting problem.

\subsection{Expanded Main Text Experiment Results}
\subsubsection{Data Scaling Gradient Maps}
\label{appendix:grad_maps}

Here we include the full gradient maps for the models described in Section~\ref{section:data_results}. For spacing reasons, we include them at the very end of the appendix.

\subsection{More Discussion}
\label{appendix:discussion}

\subsubsection{Broader Impacts}

Our work does not have any broader societal impacts beyond the general impacts that sequential-decision making research may have on automation or enabling more intelligent agents. While our work is limited to planning, some of our finding might generalize to diffusion models more broadly, which have implications for copyright/artist's rights, deepfake generation, et cetera. 

\subsubsection{How Should We Select a Method for Composition?}
\label{appendix:whywholesequence}

While we focus on the whole-sequence generation case, it is not clear what the strongest method is to generate heavily temporally correlated, sequential data. There is evidence that models trained only or mostly on auto-regressive prediction can still learn to do sophisticated forms of forwards and backwards planning \cite{lindsey2025biology}. Complex sequential generation models in video sometimes incorporate a form of auto-regressive prediction \cite{song2025history} but often generate the whole sequence in parallel with diffusion \cite{singer2022make,chen2023videocrafter1}. 

Hierarchical diffusion appears to be a strong way to incorporate both the benefits of long-horizon planning and short-horizon consistency. Besides the empirical benefits shown in the hierarchical diffusion planning papers discussed in \ref{related:diffgen}, more evidence exists that hierarchical models can learn to minimize dynamics error in fewer training steps \cite{du2024compositional}. One downside of these methods is that how the task is hierarchically decomposed either needs to be set manually, requiring more tuning, or must rely on a more complex algorithm \cite{li2023hierarchical} which increases implementation cost and other tuning. Image models have generally shown that forgoing manually set hierarchy and instead letting models jointly learn both high level structure and low level detail can work well given proper data and architecture decisions.

One consideration we do not explicitly examine is that using locality to achieve composition necessarily hurts the model's ability to use important information from prior states. Long memory can be both a blessing and a curse, as it has found to increase a policy's ability to handle very distant observations \cite{ni2023transformers} but can also replicate undesired correlations (i.e, correlations that are circumstantial and which the model should not replicate) between past observations and future actions \cite{de2019causal,wang2019monocular}. The lack of compositionality of non-locally attentive models can be thought of as a form of causal confusion, as they "confuse" specific transitions as being associated with state visitations in the distant pass that may not hold relevance for the future plan. However, recent work has found that taking a linear combination of the conditioned diffusion steps on a variety of past history lengths may preserve the benefits of both long-horizon planning and short-horizon composition, so it may be possible to combine the benefits of both~\cite{song2025history}. 

\subsubsection{Explaining Composition in Other Diffusion Planning Works}
\label{appendix:other_diffusion_planners}

Here, we apply our framework to a couple of other works in the diffusion planning literature.

\textbf{Hierarchical Diffusion Planning}: As discussed in Section \ref{related:diffgen}, several works approach the composition problem by separately generating sub-sequences with a diffusion model, and then composing them. Exactly how these low-level subsequences are "orchestrated" into a single coherent plan varies, but each exhibits aspects of locality and shift equivariance. 

\citet{chen2024simplehierarchicalplanningdiffusion} propose Hierarchical Diffuser, which uses two separate diffusion models: a high level one which generates a "plan sketch", and a low level one which generates a sequence connecting the states in the plan sketch to "fill-in" the high-level plan. Functionally similar was the prior HDI paper from \citet{li2023hierarchical}, which has a different form of generating sub-goals unrelated to diffusion. These approaches enable a form of shift equivariance by data augmentation: because the low-level subsequences generated by the low-level diffusion model are randomly cut out of the longer trajectories in the training data, they have their relative position within longer sequences removed. This also enables locality by disentangling the joint prediction of states within a particular subsequence to any other states outside of it, so each subsequence functions as a small local window without connectivity to previous or future states, preventing confounding from preventing stitching. \citet{chen2024simplehierarchicalplanningdiffusion} also confirms some of our findings that Diffuser with a larger kernel size often produces more coherent long-term plans (4.3) but that this trades-off with worse ODD compositional generalization (Appendix E). 

\citet{luo2025generative} proposes a hierarchical method called CompDiffuser which, like the Hierarchical Diffuser paper, generates sub-sequences. Instead of having a separate high-level planner, they have each sub-sequence condition its generation on its immediate surrounding neighbour subsequence (3.2). This preserves shift equivariance in the same manner as Hierarchical Diffuser, but maintains locality through this process of having each sub-sequence only attend to its neighbours at each step, which stops causal confusion. 

\textbf{Diffusion Forcing}: Some recent works extend diffusion into very long-horizon generation through using a combination of auto-regressive sampling and diffusion sampling called \emph{diffusion forcing}. Diffusion forcing denoises over a whole sequence, but iteratively denoises near-future states more aggressively, leading to the sequence being progressively denoised from start to finish instead of the entire sequence being denoised consecutively. As discussed in the main text, short-memory and future prediction enforces locality by reducing the ability for long-horizon confounding effects to prevent composition. 

Results from \citet{chen2024diffusion} shows that this approach can enable compositionality, but only when keeping short or no memory and only generating short plans, essentially using the diffusion forcing model to approximate short-horizon Model Predictive Control (E.2). When extending the generation length or memory, they find that it instead reproduces existing trajectories.

Follow up work from \citet{song2025history} extends Diffusion Forcing by replacing their RNN-based backbone with a transformer, and introducing the idea of combining the scores of the same model conditioned with several history lengths to try to achieve both long-sequence compositionality and short-term reactivity. They tested this approach on a task where a robot must take a fruit in a slot and move it to another slot. This task requires both memory (of which fruit the model should move) and short term reactivity (a human perturbs the robot mid-trajectory). However, the training data does not have both memory of the original fruit location and examples of recovery behaviour from the human perturbance, requiring composition of training sub-sequences. They found that the model that keeps the full trajectory history overfits to the training datasets showing the original fruit, while only using short-horizon models would compose recovery behaviour but forget the original task. Ultimately they found combining the score produced trajectories that both remembered important fast information and were capable of composing sub-trajectories for reactive control. This supports our argument highlighting the importance of locality for composition, but highlights that locality alone can run into issues in long-horizon tasks, as discussed in Section \ref{section:whichisbest}. For details, see Sections 6.4: Task 3, C.8, and D.4. 

\textbf{Diffusion Policies/VLAs}

Perhaps the most popular diffusion-based sequential decision making algorithms today are various forms of Diffusion Policies \cite{chi2023diffusion}. These are essentially similar to the prior Diffusion Behaviour Cloning framework \cite{pearce2023imitating} where the diffusion model acts as an explicit policy and generates an action, except that Diffusion Policy predicts a short horizon of actions that are executed. This framework of predicting a short horizon of actions with diffusion is also used by most modern foundation-model Vision Language Action models like Pi0.5 \cite{intelligence2025pi05visionlanguageactionmodelopenworld} or Gr00T \cite{bjorck2025gr00t}. 

This differs from the diffusion planning framework in that a long-horizon plan is not explicitly generated, so any long-horizon task guidance is obtained from conditioning the policy (like in goal-conditioned BC) rather than from explicit planning. Thus, in these models compositionality essentially comes from the same place that it does in TD learning: because no planning is done, the model composes through replanning. This replanning, like any explicit one-step or few-step policy, assumes the environment is Markovian and so does not include temporal information or preserve long history as conditioning, thus preserving shift equivariance and locality as we discussed in \ref{section:comp_methods}. 

There is some limited early evidence that when trained from scratch on very small datasets, Diffusion Policies may perform learning more similar to memorization with nearest-neighbour interpolation than "generalization" in any meaningful sense, and thus obtain strong apparent trajectory composition just through being very reactive \cite{he2025demystifyingdiffusionpoliciesaction}. However, these results have only been found in very small data settings, so it is likely that they do not apply to larger VLAs or to multi-task Diffusion Policies trained on a heterogeneous mix of tasks.

\subsection{What explain Composition in Prior Monolithic Diffusion Planners?}
\label{appendix:why_janner_no_compose}
Our results shown in Section~\ref{section:current_memorization} and throughout the paper indicate that diffusion planners, when used to generate long-horizon plans, tend to memorize exact sequences. This seems to contradict the results of~\citet{janner2022planning} and~\citet{ajay2023is} which both also use monolithic planners. We suspect the following reasons are at play:

\begin{itemize}
\item Most of their tasks are trained on the order of thousands of trajectories, which we find is usually sufficent to generalize.
\item The D4RL Maze datasets have been found not to require stitching \cite{ghugare2024closing}, potentially letting the diffusion planner "cheat" on the task by simply memorizing appropriate trajectories.
\item For many of their experiments they use a shorter planning horizon of $32$, allowing some stitching from replanning like with diffusion policies~\cite{chi2023diffusion}.

\end{itemize}
\newpage

\begin{figure}
\begin{subfigure}[T]{.33\linewidth}
    \centering
    \includegraphics[width=\linewidth]{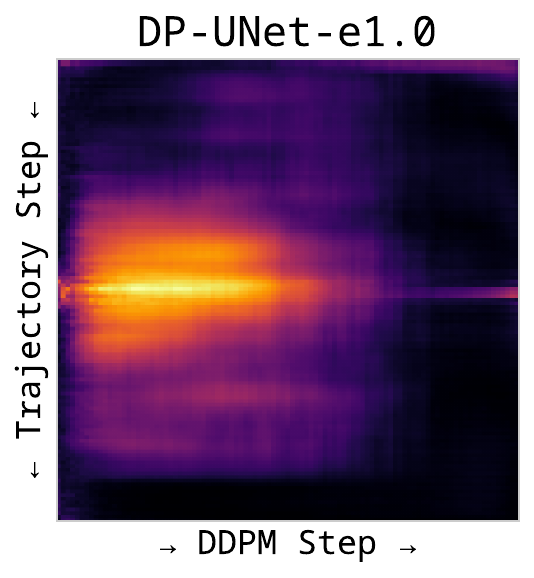}
    \end{subfigure}%
\begin{subfigure}[T]{.33\linewidth}
    \centering
    \includegraphics[width=\linewidth]{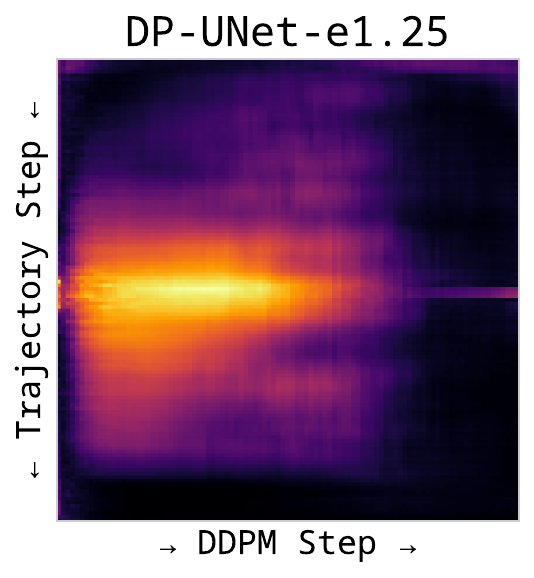}
    \end{subfigure}%
\begin{subfigure}[T]{.33\linewidth}
    \centering
    \includegraphics[width=\linewidth]{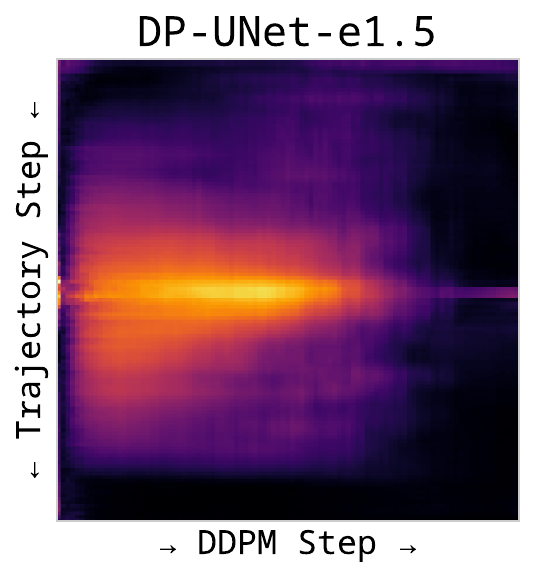}
    \end{subfigure}%
\end{figure}

\begin{figure}
\begin{subfigure}[T]{.33\linewidth}
    \centering
    \includegraphics[width=\linewidth]{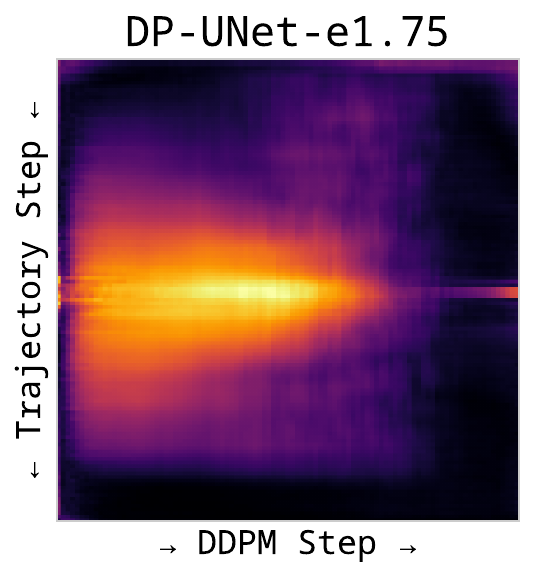}
    \end{subfigure}%
\begin{subfigure}[T]{.33\linewidth}
    \centering
    \includegraphics[width=\linewidth]{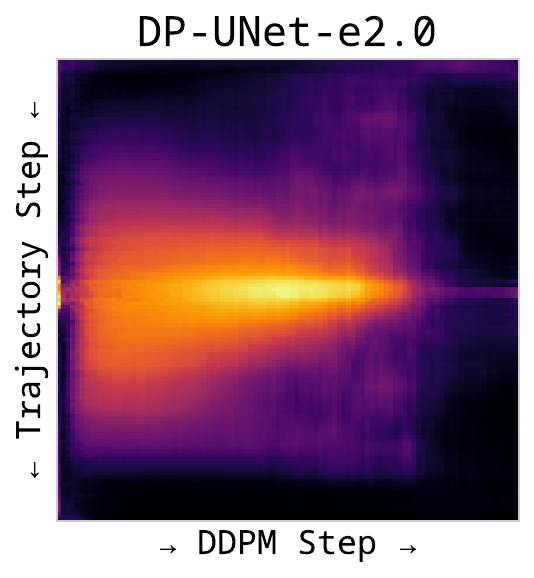}
    \end{subfigure}%
\begin{subfigure}[T]{.33\linewidth}
    \centering
    \includegraphics[width=\linewidth]{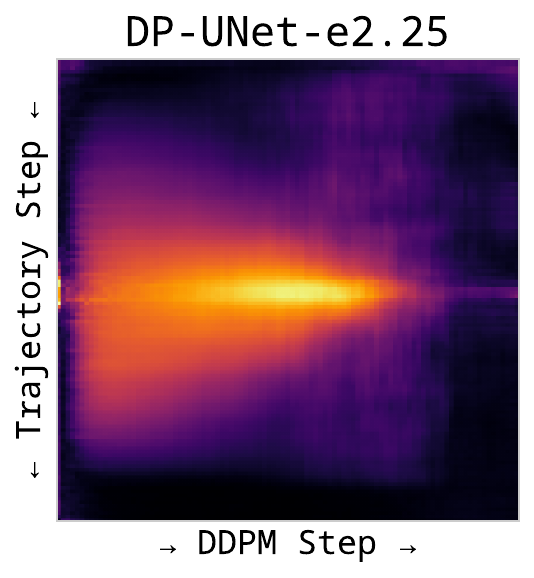}
    \end{subfigure}%
\end{figure}

\begin{figure}
\begin{subfigure}[T]{.33\linewidth}
    \centering
    \includegraphics[width=\linewidth]{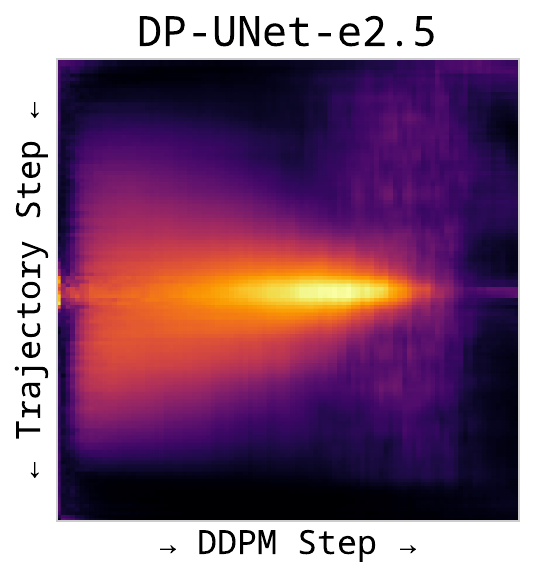}
    \end{subfigure}%
\begin{subfigure}[T]{.33\linewidth}
    \centering
    \includegraphics[width=\linewidth]{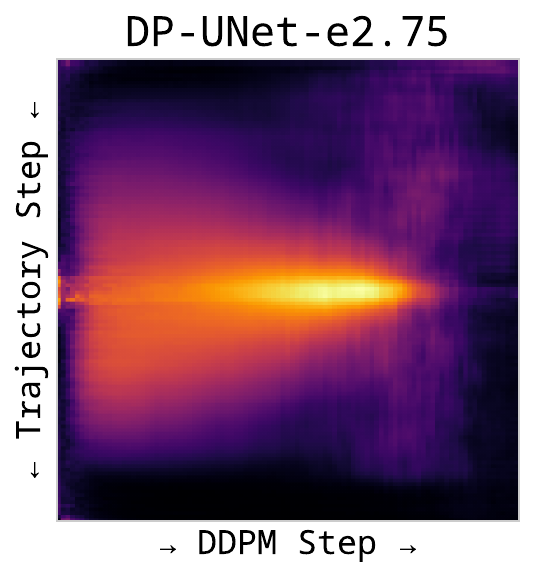}
    \end{subfigure}%
\begin{subfigure}[T]{.33\linewidth}
    \centering
    \includegraphics[width=\linewidth]{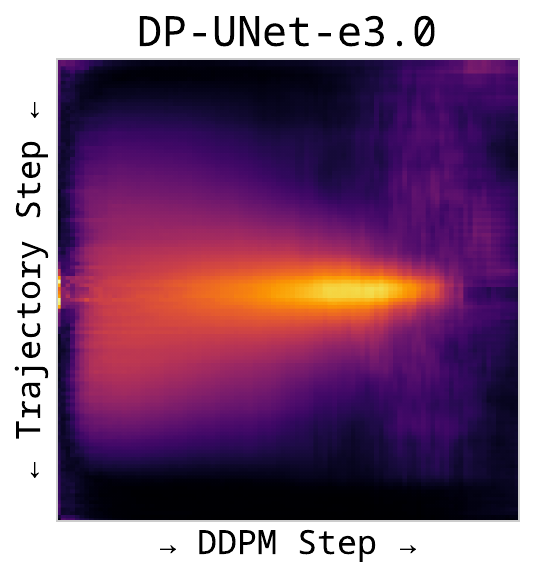}
    \end{subfigure}%
\end{figure}

\begin{figure}
\begin{subfigure}[T]{.33\linewidth}
    \centering
    \includegraphics[width=\linewidth]{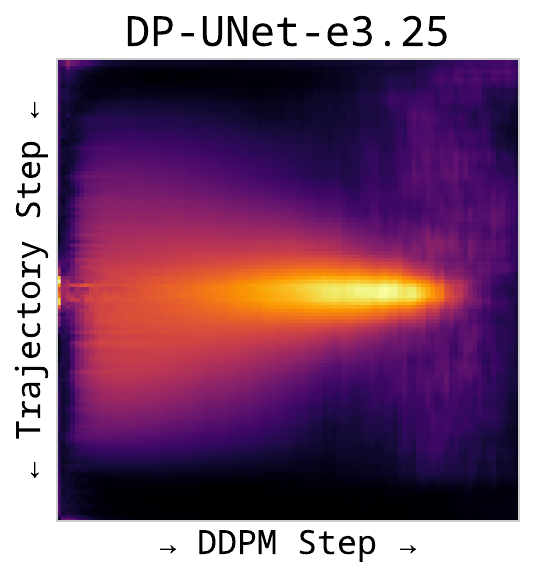}
    \end{subfigure}%
\begin{subfigure}[T]{.33\linewidth}
    \centering
    \includegraphics[width=\linewidth]{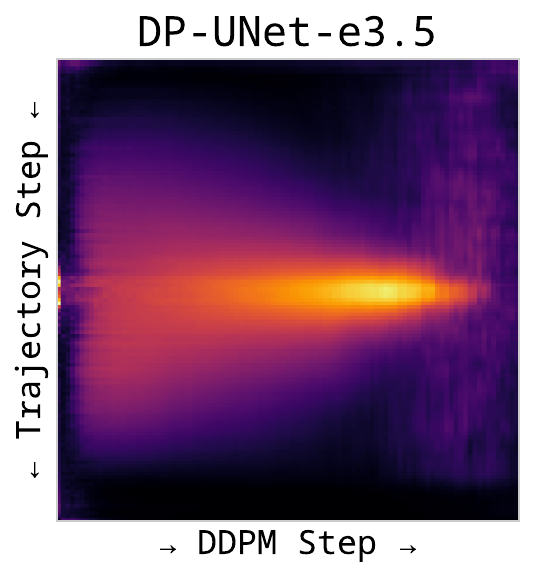}
    \end{subfigure}%
\begin{subfigure}[T]{.33\linewidth}
    \centering
    \includegraphics[width=\linewidth]{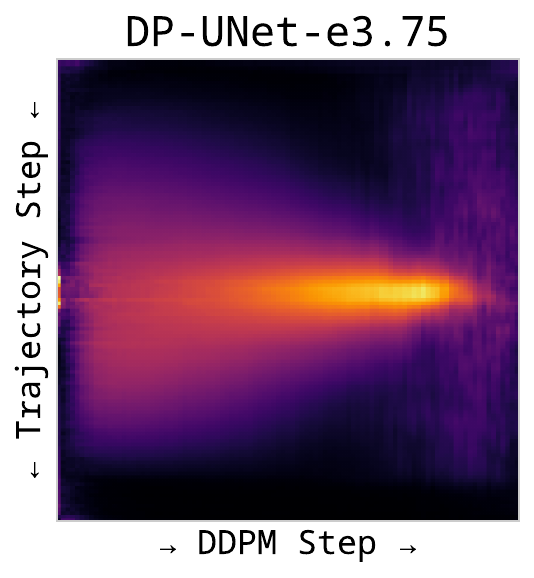}
    \end{subfigure}%
\end{figure}

\begin{figure}
\begin{subfigure}[T]{.33\linewidth}
    \centering
    \includegraphics[width=\linewidth]{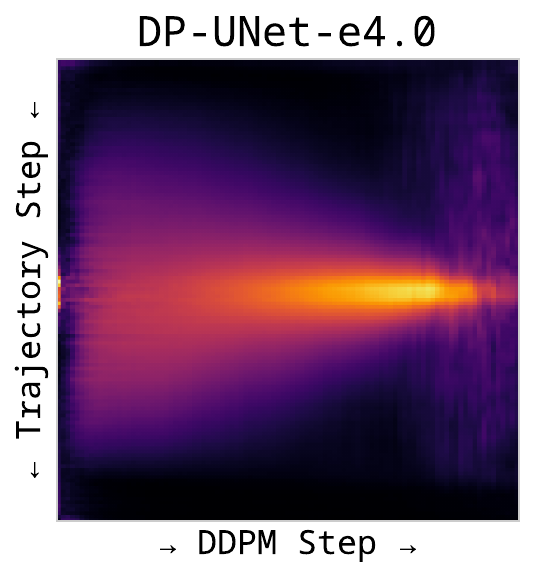}
    \end{subfigure}%
\begin{subfigure}[T]{.33\linewidth}
    \centering
    \includegraphics[width=\linewidth]{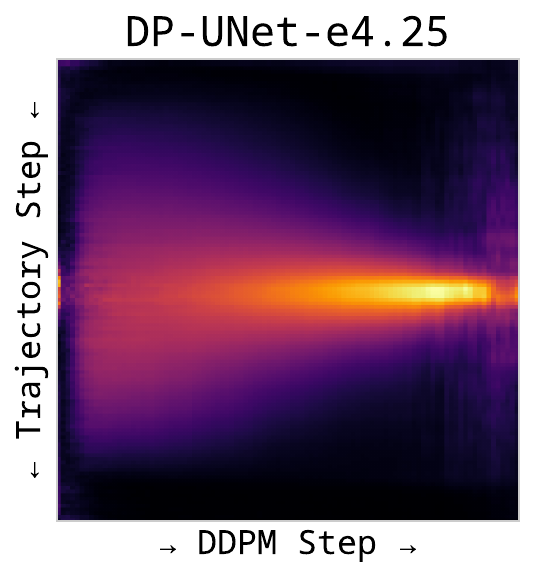}
    \end{subfigure}%
\begin{subfigure}[T]{.33\linewidth}
    \centering
    \includegraphics[width=\linewidth]{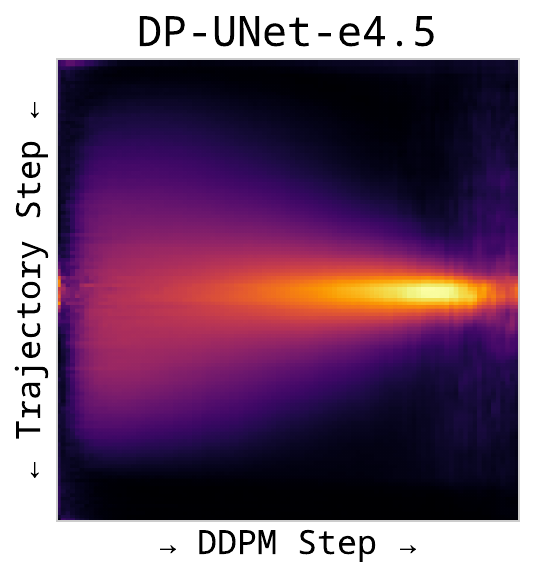}
    \end{subfigure}%
\end{figure}

\begin{figure}
\begin{subfigure}[T]{.33\linewidth}
    \centering
    \includegraphics[width=\linewidth]{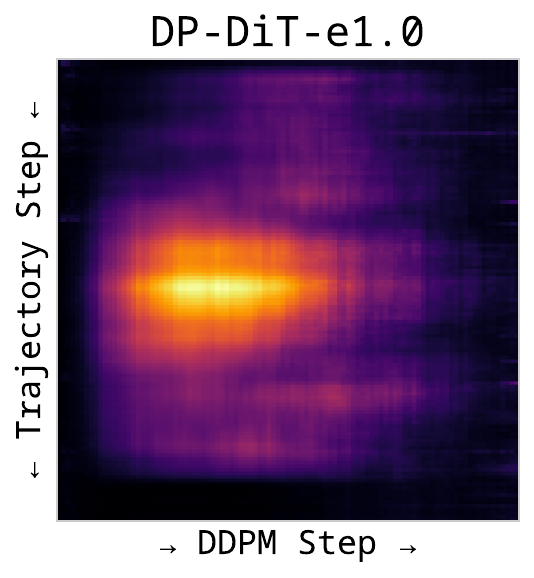}
    \end{subfigure}%
\begin{subfigure}[T]{.33\linewidth}
    \centering
    \includegraphics[width=\linewidth]{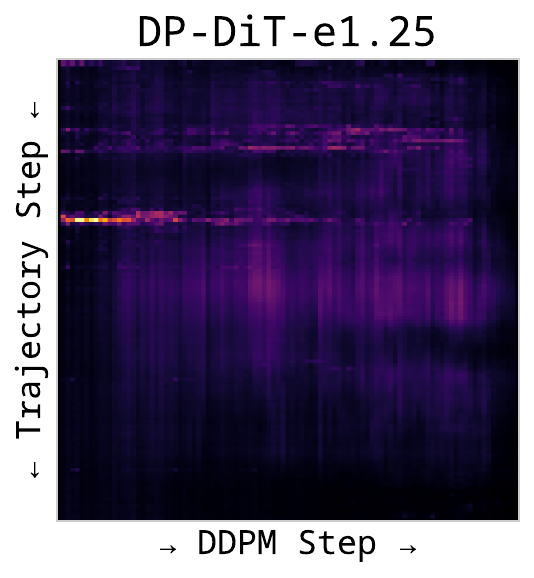}
    \end{subfigure}%
\begin{subfigure}[T]{.33\linewidth}
    \centering
    \includegraphics[width=\linewidth]{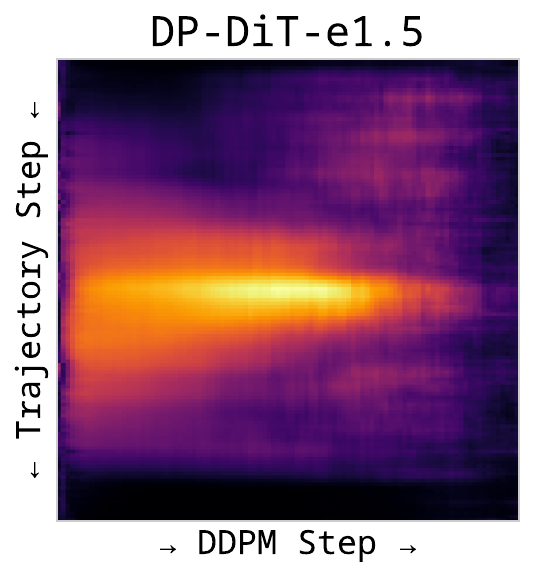}
    \end{subfigure}%
\end{figure}

\begin{figure}
\begin{subfigure}[T]{.33\linewidth}
    \centering
    \includegraphics[width=\linewidth]{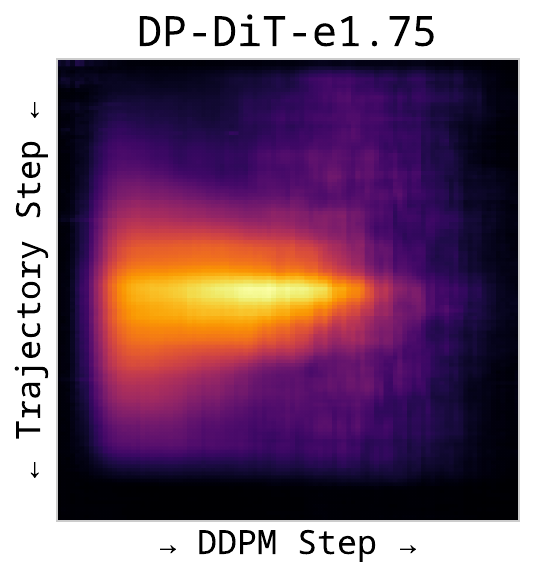}
    \end{subfigure}%
\begin{subfigure}[T]{.33\linewidth}
    \centering
    \includegraphics[width=\linewidth]{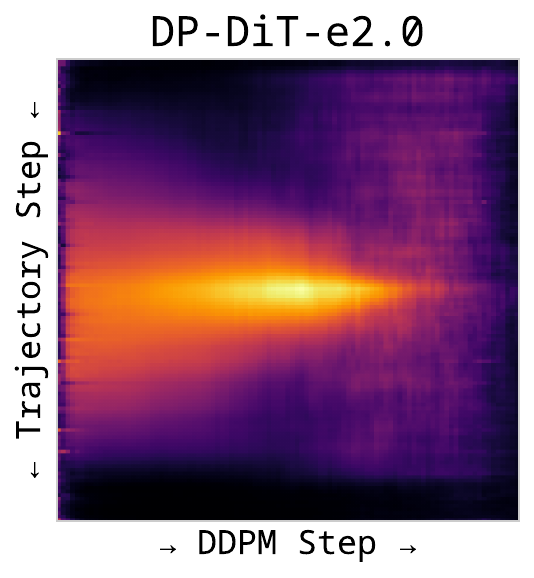}
    \end{subfigure}%
\begin{subfigure}[T]{.33\linewidth}
    \centering
    \includegraphics[width=\linewidth]{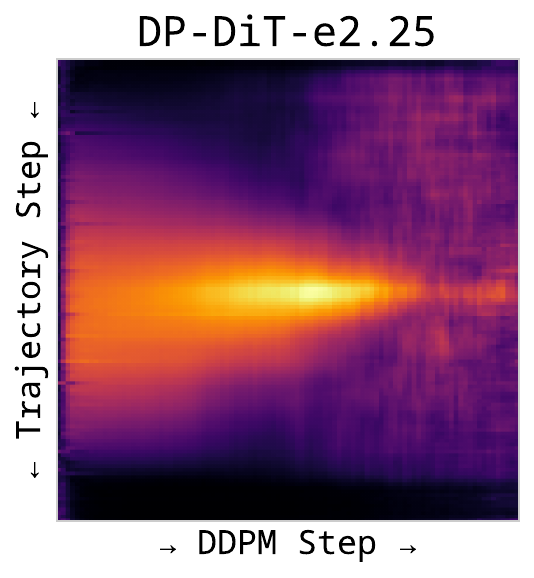}
    \end{subfigure}%
\end{figure}

\begin{figure}
\begin{subfigure}[T]{.33\linewidth}
    \centering
    \includegraphics[width=\linewidth]{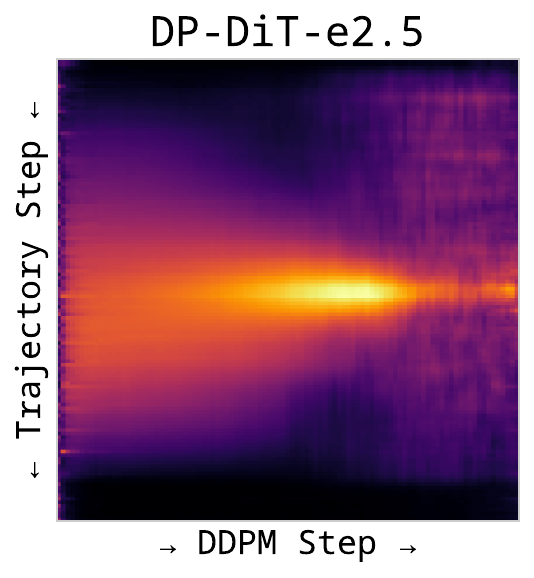}
    \end{subfigure}%
\begin{subfigure}[T]{.33\linewidth}
    \centering
    \includegraphics[width=\linewidth]{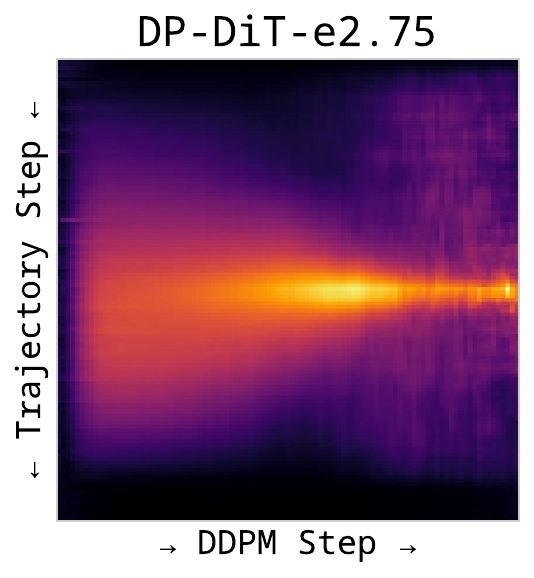}
    \end{subfigure}%
\begin{subfigure}[T]{.33\linewidth}
    \centering
    \includegraphics[width=\linewidth]{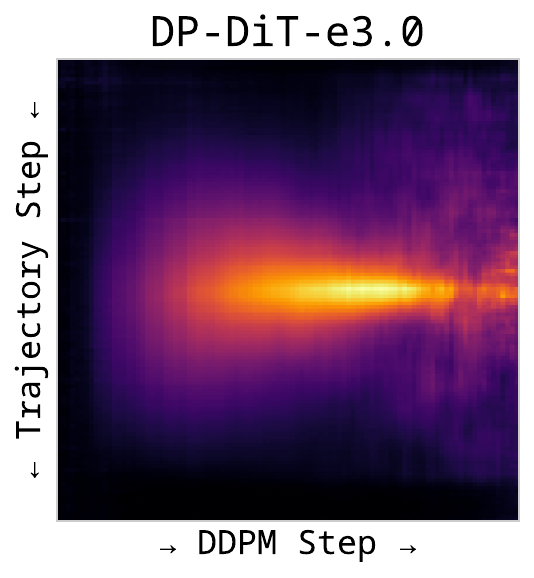}
    \end{subfigure}%
\end{figure}

\begin{figure}
\begin{subfigure}[T]{.33\linewidth}
    \centering
    \includegraphics[width=\linewidth]{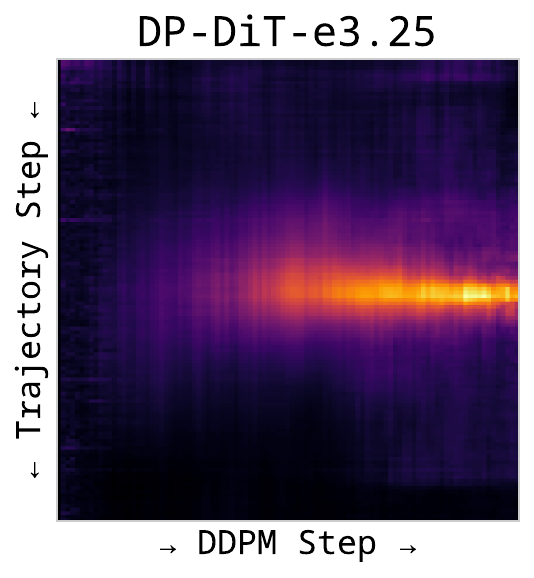}
    \end{subfigure}%
\begin{subfigure}[T]{.33\linewidth}
    \centering
    \includegraphics[width=\linewidth]{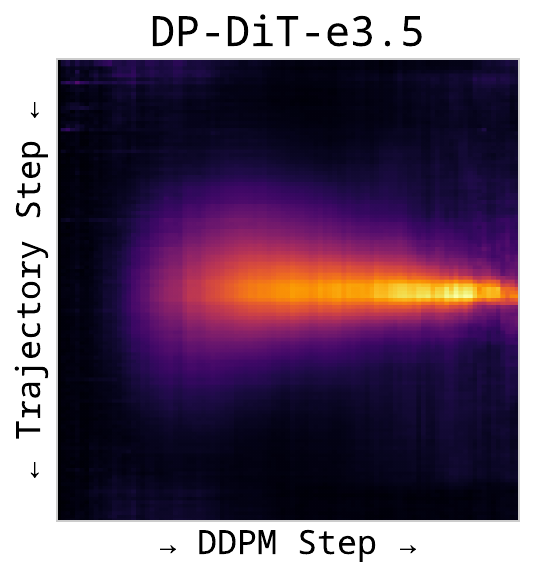}
    \end{subfigure}%
\begin{subfigure}[T]{.33\linewidth}
    \centering
    \includegraphics[width=\linewidth]{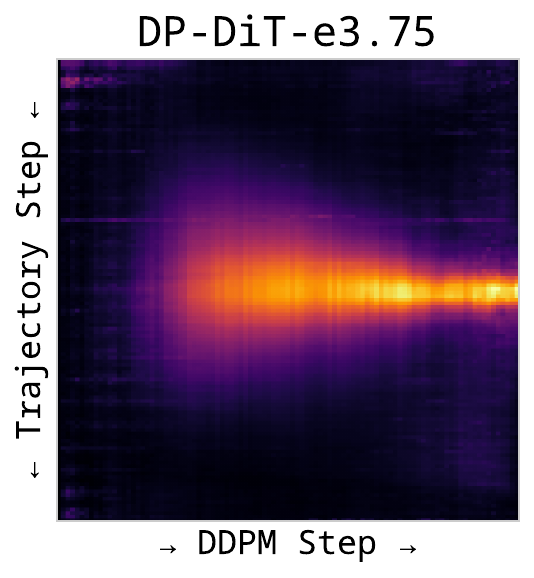}
    \end{subfigure}%
\end{figure}

\begin{figure}
\begin{subfigure}[T]{.33\linewidth}
    \centering
    \includegraphics[width=\linewidth]{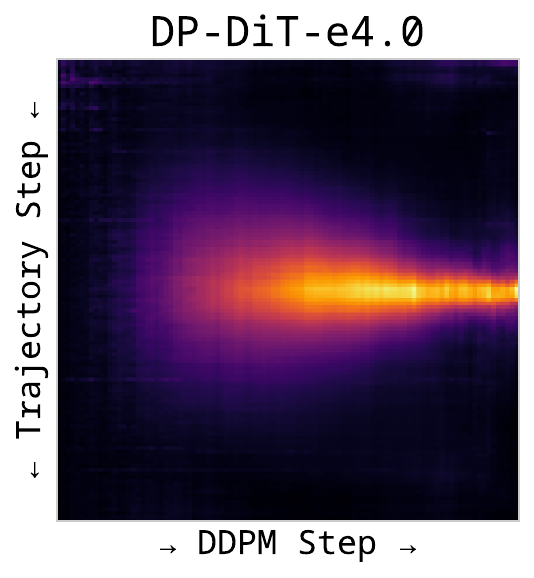}
    \end{subfigure}%
\begin{subfigure}[T]{.33\linewidth}
    \centering
    \includegraphics[width=\linewidth]{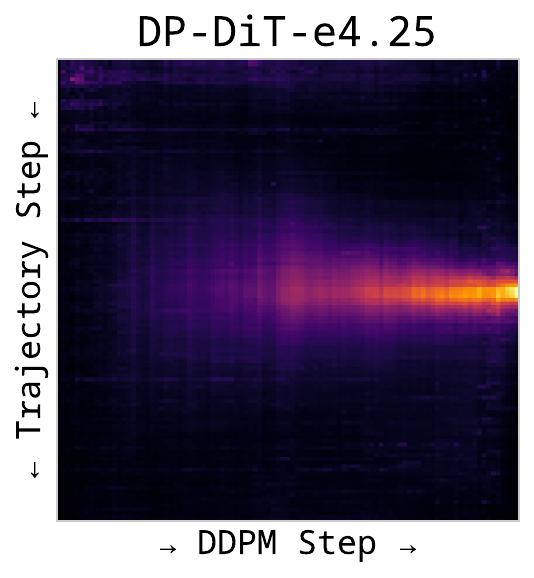}
    \end{subfigure}%
\begin{subfigure}[T]{.33\linewidth}
    \centering
    \includegraphics[width=\linewidth]{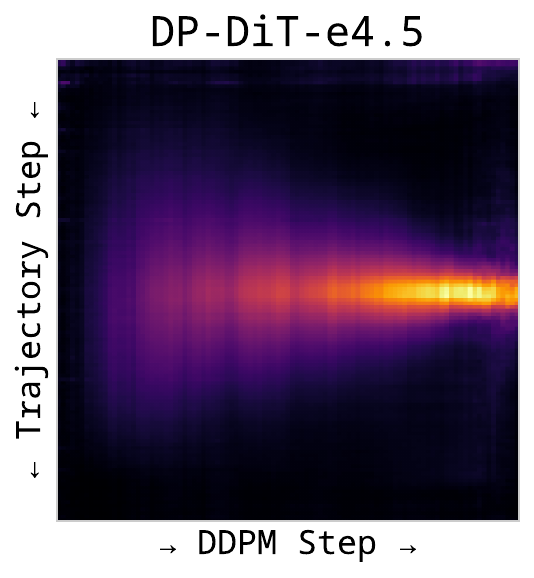}
    \end{subfigure}%
\end{figure}

\begin{figure}
\begin{subfigure}[T]{.33\linewidth}
    \centering
    \includegraphics[width=\linewidth]{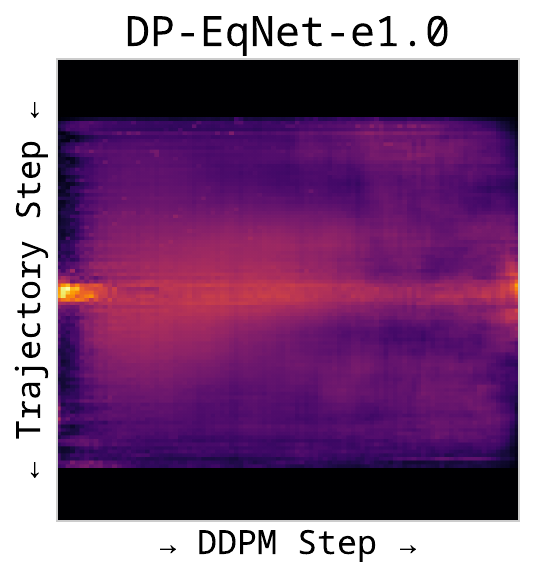}
    \end{subfigure}%
\begin{subfigure}[T]{.33\linewidth}
    \centering
    \includegraphics[width=\linewidth]{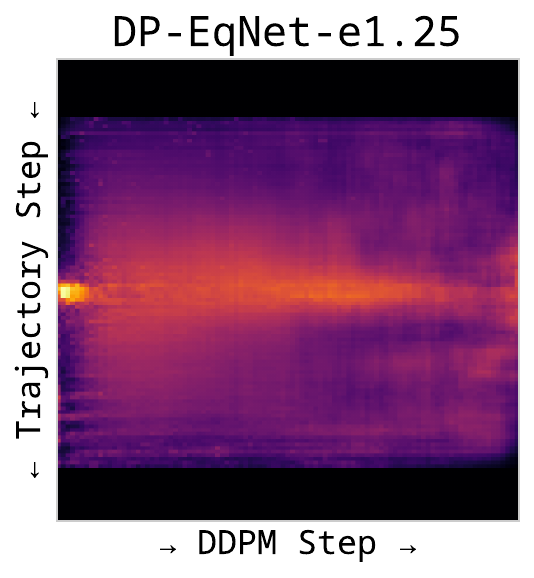}
    \end{subfigure}%
\begin{subfigure}[T]{.33\linewidth}
    \centering
    \includegraphics[width=\linewidth]{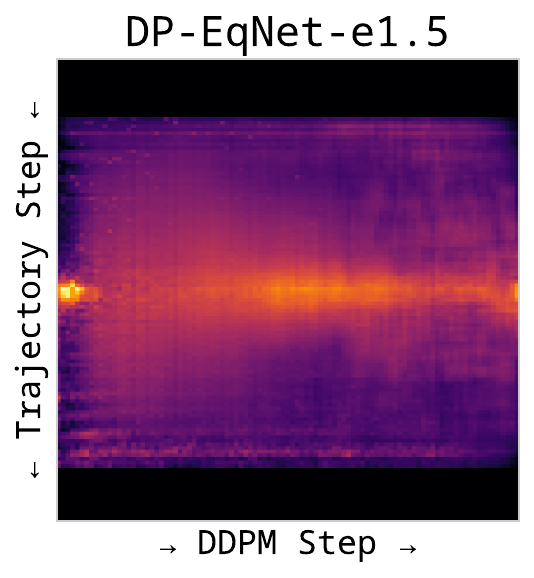}
    \end{subfigure}%
\end{figure}

\begin{figure}
\begin{subfigure}[T]{.33\linewidth}
    \centering
    \includegraphics[width=\linewidth]{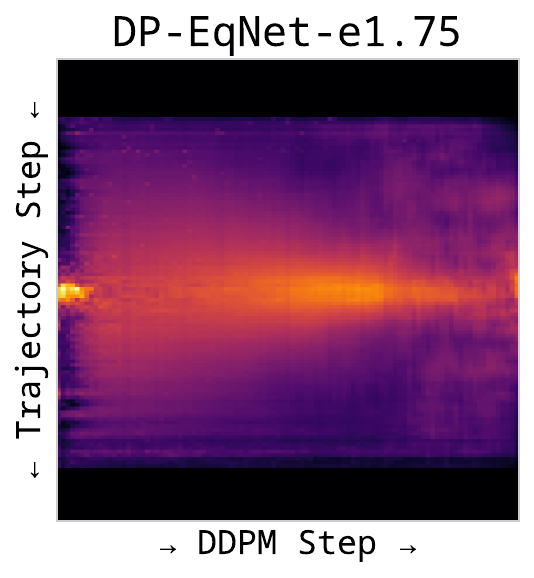}
    \end{subfigure}%
\begin{subfigure}[T]{.33\linewidth}
    \centering
    \includegraphics[width=\linewidth]{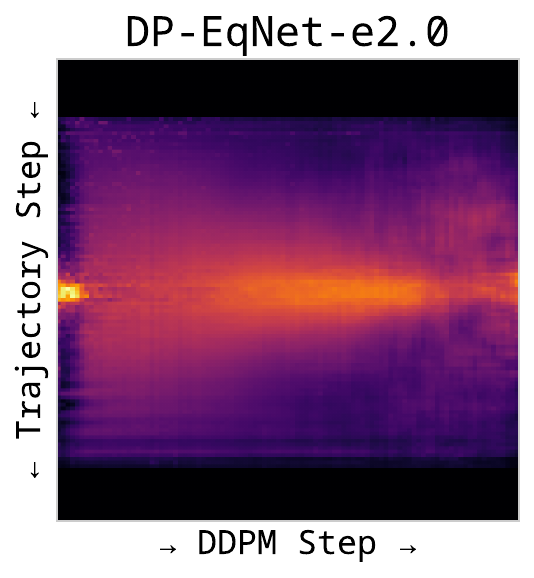}
    \end{subfigure}%
\begin{subfigure}[T]{.33\linewidth}
    \centering
    \includegraphics[width=\linewidth]{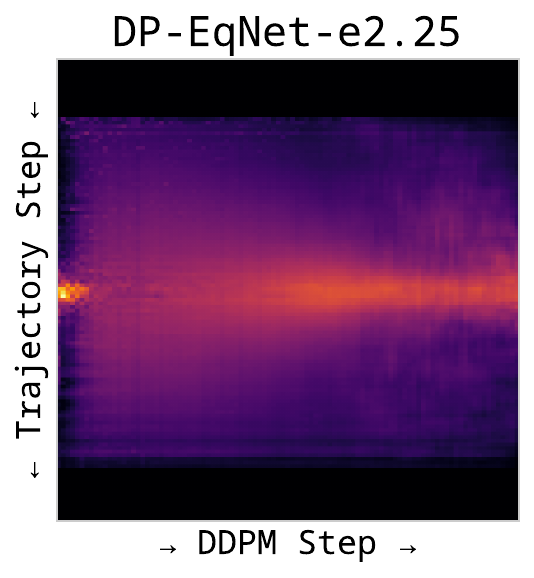}
    \end{subfigure}%
\end{figure}

\begin{figure}
\begin{subfigure}[T]{.33\linewidth}
    \centering
    \includegraphics[width=\linewidth]{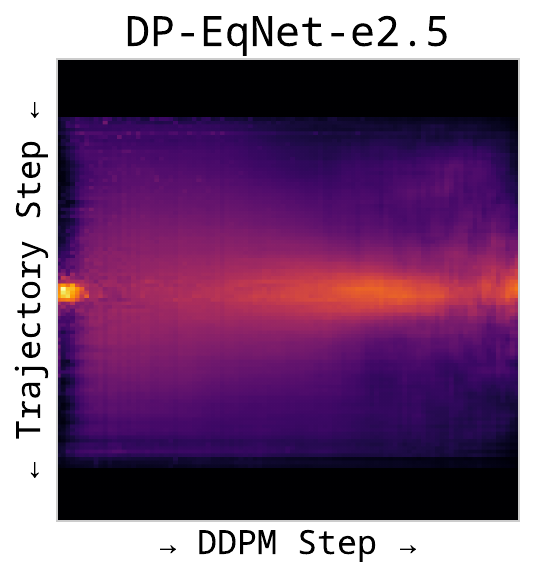}
    \end{subfigure}%
\begin{subfigure}[T]{.33\linewidth}
    \centering
    \includegraphics[width=\linewidth]{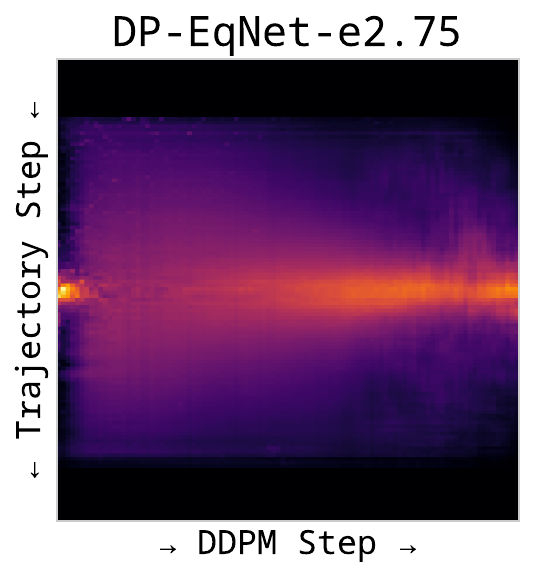}
    \end{subfigure}%
\begin{subfigure}[T]{.33\linewidth}
    \centering
    \includegraphics[width=\linewidth]{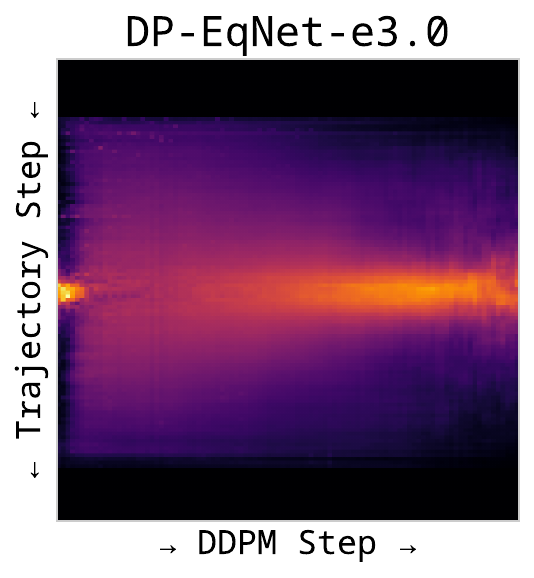}
    \end{subfigure}%
\end{figure}

\begin{figure}
\begin{subfigure}[T]{.33\linewidth}
    \centering
    \includegraphics[width=\linewidth]{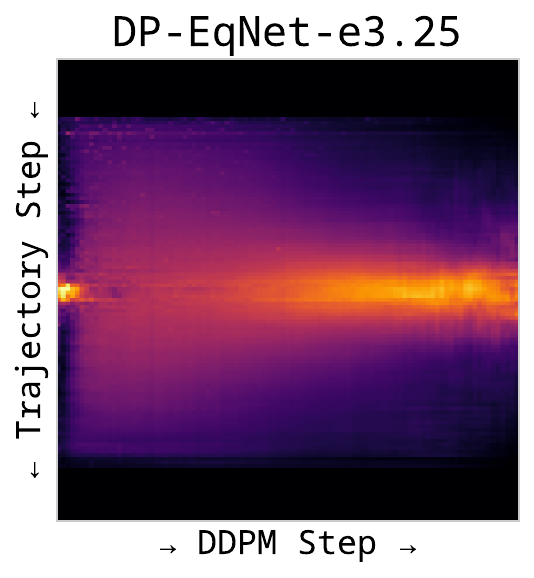}
    \end{subfigure}%
\begin{subfigure}[T]{.33\linewidth}
    \centering
    \includegraphics[width=\linewidth]{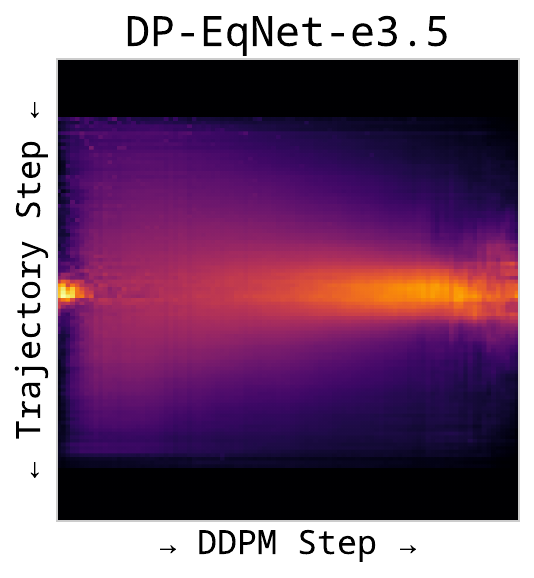}
    \end{subfigure}%
\begin{subfigure}[T]{.33\linewidth}
    \centering
    \includegraphics[width=\linewidth]{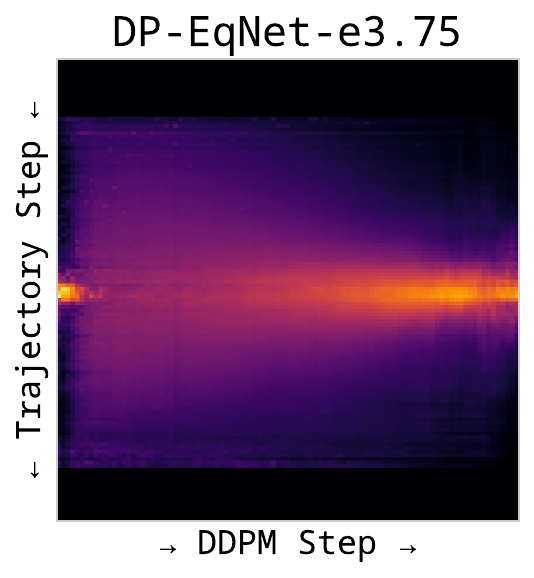}
    \end{subfigure}%
\end{figure}

\begin{figure}
\begin{subfigure}[T]{.33\linewidth}
    \centering
    \includegraphics[width=\linewidth]{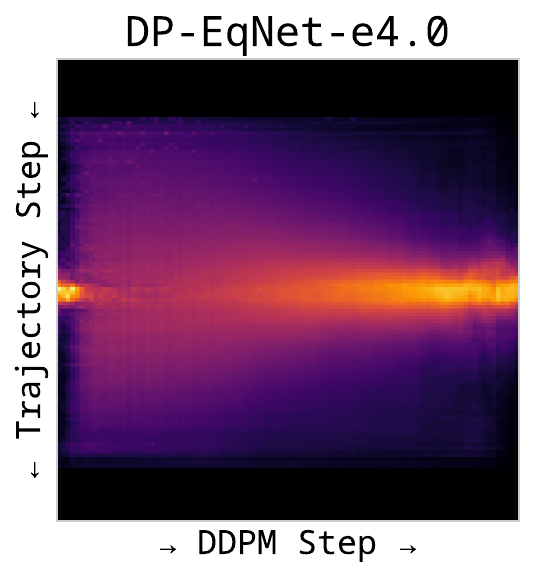}
    \end{subfigure}%
\begin{subfigure}[T]{.33\linewidth}
    \centering
    \includegraphics[width=\linewidth]{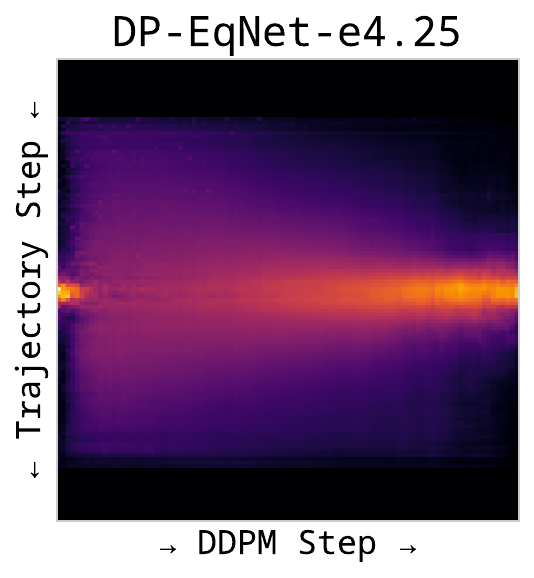}
    \end{subfigure}%
\begin{subfigure}[T]{.33\linewidth}
    \centering
    \includegraphics[width=\linewidth]{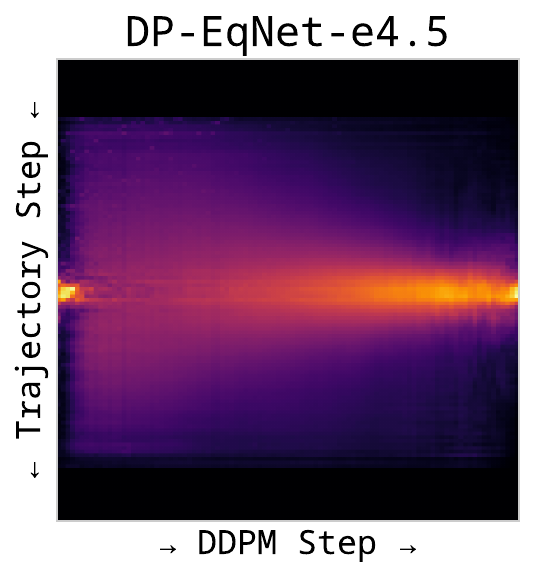}
    \end{subfigure}%
\end{figure}

\end{document}